\pdfoutput=1

\documentclass[11pt]{article}

\usepackage[]{EMNLP2023}

\usepackage{times}
\usepackage{latexsym}

\usepackage[T1]{fontenc}

\usepackage[utf8]{inputenc}

\usepackage{microtype}

\usepackage{inconsolata}
\usepackage{dsfont}
\usepackage{multirow}
\usepackage{graphicx}
\usepackage{arydshln}
\usepackage{tabularray}
\usepackage{subfig}
\usepackage{stfloats}
\usepackage{listings}
\usepackage{multirow}
\usepackage{lscape}
\usepackage{multicol}

\usepackage{supertabular,booktabs}

\newcommand{\RNum}[1]{\uppercase\expandafter{\romannumeral #1\relax}}
\usepackage{amsmath}

\definecolor{f1color}{HTML}{03c988}
\definecolor{pf1color}{HTML}{ed2b2a}
\definecolor{accolor}{HTML}{f97b22}
\definecolor{paccolor}{HTML}{62cdff}
\newcommand{\OB}{$\star$}

\title{Synthetic Data Generation with Large Language Models for Text Classification: Potential and Limitations} 

\author{
Zhuoyan Li\textsuperscript{1}, Hangxiao Zhu\textsuperscript{2}, Zhuoran Lu\textsuperscript{1}, Ming Yin\textsuperscript{1} \\
\textsuperscript{1}Purdue University  \\
\textsuperscript{2}Washington University in St. Louis \\ \texttt{\{li4178, lu800, mingyin\}@purdue.edu}, \texttt{hangxiao@wustl.edu} 
}

\begin{document}
\maketitle
\begin{abstract}
The collection and curation of high-quality training data is crucial for developing text classification models with superior performance, but it is often associated with significant costs and time investment. 
Researchers have recently explored using large language models (LLMs) to generate synthetic datasets as an alternative approach. However, the effectiveness of the LLM-generated synthetic data in supporting model training is inconsistent across different classification tasks. 
To better understand factors that moderate the effectiveness of the LLM-generated synthetic data, in this study, we look into how the performance of models trained on these synthetic data may vary with the {\em subjectivity} of classification. 
Our results indicate that subjectivity, at both the task level and instance level, is negatively associated with the performance of the model trained on synthetic data.  
We conclude by discussing the implications of our work on the potential and limitations of leveraging LLM for synthetic data generation\footnote{The collected human annotations are available at \\huggingface.co/datasets/xfleezy/human\_annotation\_emnlp23.}. 
\end{abstract}

\section{Introduction}
Today, machine-learning-powered text classification models have been widely applied in diverse applications such as detecting biased or toxic language on online platforms~\cite{wiegand2019detection} and filtering spam emails~\cite{jindal2007review}. However, the performance of these models largely depends on the quality of the training data. This poses a substantial challenge in practice, especially when models need to be built for a novel task domain or to incorporate new classification categories, as the training data collection and curation process is often costly, time-consuming, and complex. 

Meanwhile, with the recent advancements in large language models (LLMs), researchers have started to explore the potential of utilizing LLMs for generating synthetic data tailored to specific tasks and augmenting the training data in low-resourced data settings~\cite{kumar2020data,yoo2021gpt3mix,hartvigsen2022toxigen,sahu2022data}.
Most recently, a few studies also investigate into the feasibility of generating a synthetic dataset from scratch using LLMs to support zero-shot learning~\cite{ye2022zerogen,wang2021towards,tang2023does,gao2023self}. 
While LLM-based data augmentation is often found to outperform other data augmentation methods in boosting the model performance, mixed results are reported regarding whether the LLM-generated synthetic data can effectively support model training to enable a level of model performance that is comparable to models trained on the data collected in the real world and carefully annotated. This leaves uncertainty for researchers and practitioners in deciding whether to rely on LLMs for synthetic data generation or to proceed with the traditional data collection and curation pipeline when they need to construct a text classification model for a new task. Naturally, one may wonder {\em what factors might moderate the effectiveness of LLM-generated synthetic data in facilitating successful model training}. 

We conjecture that one such factor could be the {\em subjectivity} of classification tasks. Indeed, language is inherently subjective and interpretive~\cite{benveniste1971subjectivity,wiebe2004learning}. Previous research has showed that people often perceive the same text in different ways because of their personal biases and perspectives~\cite{sap2021annotators,li2022towards,gordon2022jury}. Thus, achieving high model performance for classification tasks with high subjectivity seems to impose a greater demand on the training data in reflecting the richness and nuances present in human language, and the extent to which LLM-generated synthetic data can acompolish this objective is unclear.

Thus, in this paper, we formally evaluate the effectiveness of LLM (i.e., the cutting-edge GPT-3.5-Turbo model) in generating synthetic data to support model training for different text classification tasks. We adopt two approaches for synthetic data generation---a {\em zero-shot} setting in which the LLM is directly prompted to generate text instances with different labels of interests, and a {\em few-shot} setting in which a few real-world data instances are provided as examples to guide the LLM in generating the synthetic data. We conduct two evaluation studies, each corresponding to one dimension of subjectivity---the first study examines the effectiveness of the synthetic data on 10 types of classification tasks and explores how it varies with the {\em task-level subjectivity} (i.e., whether this type of classification task is subjective); the second study concerns that given a specific classification task, how the performance of a model trained on synthetic data changes with the {\em instance-level subjectivity} (i.e., whether people tend to disagree with each other on the label of this task instance).  Our findings suggest that across the 10 types of classification tasks that we have considered in this study, models trained on the LLM-generated synthetic data generally perform worse than those trained on the real-world data, yet guiding LLM's synthetic data generation process with a small amount of real-world data (i.e., as done in the few-shot data generation setting) can improve the effectiveness of the data generated. 
Moreover, we find that the performance of models trained on the LLM-generated synthetic data is very close to those trained on the real-world data for tasks with low subjectivity (e.g., news topic classification, spam email detection), while the performance decrease is much bigger on tasks with high subjectivity (e.g., humor or sarcasm detection). Finally, even within the same type of classification task, models trained on the LLM-generated synthetic data tend to exhibit a higher level of performance on those task instances with lower subjectivity, for which human annotators exhibit a higher level of agreement in their annotation.

Together, our study provides important experimental evidence regarding the potential and limitations of using LLMs to generate synthetic data for text classification tasks. We conclude by discussing the implications, limitations, and future work of our study.

\section{Related Work}
\noindent \textbf{Generative AI in synthetic data generation.} Recent advancements in generative AI have motivated numerous studies to explore the potential of leveraging generative models to create synthetic data for training machine learning models, especially for computer vision (CV) and natural language processing (NLP) tasks. In the realm of CV, several works have utilized GAN-based models~\cite{karras2019style} or diffusion models~\cite{nichol2021glide} to generate synthetic data for image recognition~\cite{besnier2020dataset,he2022synthetic} or object segmentation~\cite{zhang2021datasetgan}. Similarly, in the NLP field, researchers have also probed into the capacity of language models in generating synthetic data for various text classification tasks~\cite{kumar2020data,chung2023increasing, sahu2022data, yoo2021gpt3mix, ye2022zerogen,wang2021towards,hartvigsen2022toxigen,meng2022generating,gao2022self,aggarwal2022entity, chen2022weakly}, with mixed results reported regarding the effectiveness of the synthetic data generated.  
In this study, we aim to obtain a better understanding of {\em when} the synthetic data generated by language models can lead to effective model training, and we focus on exploring the role of task subjectivity in moderating the effectiveness of the synthetic data.

\noindent \textbf{Large language models.}
Based on the Transformer architecture~\cite{vaswani2017attention}, large language models (LLMs) have facilitated remarkable progress in the field of natural language processing. The utilization of bidirectional contexts in the BERT model~\cite{devlin2018bert} has resulted in superior performance across a wide range of tasks. Building on this, OpenAI's GPT series, comprising of models like GPT-2~\cite{radford2019language}, the colossal GPT-3~\cite{brown2020language} with an impressive 175 billion parameters and the most recent GPT-4~\cite{openai2023gpt4}, pushed the boundaries of possibilities of LLMs. These models exhibit remarkable proficiency in generating high-quality human-like text~\cite{clark2021all,dou2021gpt,10.1145/3544548.3581318}, showcasing capabilities in rudimentary reasoning~\cite{wei2021finetuned}, translation~\cite{brown2020language}, scientific synthetic data generation~\cite{10.1145/3544548.3580688}, and code generation~\cite{10.1145/3544548.3580940}.  
In this study, we focus on leveraging the cutting-edge GPT-3.5-Turbo model\footnote{We used GPT-3.5-Turbo as the foundational model to generate synthetic data because at the time of this study, an official API for the more advanced GPT-4 model was not yet available from OpenAI.} to explore its capabilities and limitations in synthesizing data for text classification tasks with different subjectivity levels.

\section{Methodolgy}
In this section, we outline the procedure we have followed when leveraging the large language model  to generate the synthetic training data for text classification. We consider two data generation settings in this study, i.e., the \textit{zero-shot} setting and the \textit{few-shot} setting.

\subsection{Zero-shot Synthetic Data Generation}
\label{zero-shot}
Under the {\em zero-shot} synthetic data generation setting, given a text classification task, we assume that the real-world data in the form of ``text-label pairs'' do not exist. 
Thus, in order to obtain synthetic training data for the text classification task, two sequential prompts are constructed and supplied to the pretrained large language model (i.e., the GPT-3.5-Turbo model). First, a customized ``context prompt'' relevant to the targeted domain of interest is used to set the context. For example, in the case of the IMDB movie review classification task~\cite{maas-EtAl:2011:ACL-HLT2011}, the customized context prompt used is ``Imagine you are a movie reviewer on the IMDB platform''. This prompt aims to encourage the LLM to generate synthetic data that resemble the real texts produced in the targeted domain. 
After the context is set, a second prompt, i.e., the ``data generation prompt'', is provided to the LLM, instructing the model to generate texts with a specific style, label (with respect to the classification task of interest), and word limit. 
For example, for the IMDB movie review classification task, the style of the text is a movie review, and the label is a targeted sentiment conveyed by the review (i.e., ``positive'' or ``negative''). 
To further enhance the diversity of the generated data, after the generation of every $n$ data points (i.e., texts of targeted styles, labels, and word limits)\footnote{To increase data diversity while maintaining a reasonable data generation speed, $n$ is set to 10 for generating short texts (i.e., texts with a maximum length of 30 words), and 1 for generating longer paragraphs. 
},  
we provide a ``diversity prompt'' to the LLM---``Can you provide something more diverse compared to the previously generated data?''---aiming to increase the diversity of the synthetic data generated.


\subsection{Few-shot Synthetic Data Generation}
\label{few-shot}
Under the {\em few-shot} synthetic data generation setting, we assume that a small amount of 
real-world data 
are available for the text classification task. These data points can then serve as the examples for the large language model in the data generation process, which can potentially provide LLM with insights of the patterns exhibited in the real-world data.  We again start the data generation process by using a context prompt to set the context. However, different from that in the zero-shot setting, here, each time before we instruct the LLM to generate a piece of text, we first provide the model with a few randomly sampled real-world data instances (including both the text and the label) as the examples. 
To keep the LLM from merely rephrasing the provided examples, an additional prompt is used to impose a constraint on the LLM in generating the synthetic data (i.e., ``You should imitate the example I have provided, but you cannot simply modify or rewrite the example I have given.''). 

For more details about prompts used for generating data for each type of text classification task, please refer to the App.~\ref{prompts}.
\section{Evaluation \RNum{1}: Comparison Across Different Types of Tasks}
In our first evaluation study, we investigate into how well the synthetic data generated by LLM under both zero-shot and few-shot settings can support effective model training for different types of text classification tasks. 
We are especially interested in comparing the model performance between those trained on the real-world data and on the LLM-generated synthetic data, and in understanding how the performance of those models trained on the LLM-generated synthetic data varies with the subjectivity of the text classification task.
\subsection{Datasets and Tasks}
We experiment  with 10 representative datasets covering a variety of text classification tasks: AG's news \citep{Zhang2015CharacterlevelCN}, IMDB reviews \citep{maas-EtAl:2011:ACL-HLT2011}, 
SMS  spam~\citep{Almeida2011SpamFiltering}, Financial phrase bank~\citep{Malo2014GoodDO},
Reddit emotion~\citep{demszky2020goemotions}, 
Relation classification~\citep{gao-etal-2019-fewrel}, Tweet irony speech~\citep{van2018semeval}, 
Tweet emotions~\citep{mohammad2018semeval}, 
Sarcasm news (\citealp{misra2023Sarcasm}, \citealp{misra2021sculpting}), and 
Humor speech~\citep{annamoradnejad2020colbert}. 
See App.~\ref{dataset} for detailed descriptions of datasets and the corresponding text classification tasks.
These datasets are selected with the goal of spanning a wide range of task subjectivity in mind. For example, we conjecture that classifying the news topic category (e.g., as that in the AG's news dataset) is relatively objective, while determining whether texts are humorous (e.g., as that in the Humor speech dataset) is quite subjective~\cite{veatch1998theory}.

\subsection{Task-level Subjectivity Determination}
\label{subrank}

To formally determine the subjectivity levels of different text classification tasks, we first conduct a crowdsourced study to collect subjectivity judgements from the crowd.

\noindent \textbf{Study procedure.}
We adopt a comparative approach to collect crowdsourced subjectivity judgements in this study. 
Specifically, 
we recruited crowd workers from Amazon Mechanical Turk (MTurk), and each worker was asked to complete a sequence of 10 subjectivity judgement tasks. 
In each task, we randomly sampled a pair of text classification tasks from the 10 tasks that we considered in this evaluation, and we presented to the worker the task description, label description, and task examples for each task in the pair. 
Then, the worker was asked to determine which text classification task in the pair was more objective, with ``objectivity'' of a task defined as ``the classification of a piece of text is based on clear, identifiable features in the text (e.g., keywords or phrases), and can be done without being affected by any personal interpretation of the text resulted from personal biases, emotions or beliefs.''  The study was restricted to U.S. workers. Each worker was allowed to participate only once and received a \$1.2 payment. An attention check question was included in the study to validate the worker's engagement, and only the data from workers who successfully passed the attention check were considered valid.

\noindent \textbf{Ranking task subjectivity.}
After excluding responses from inattentive workers, a total of 540 pairwise subjectivity comparisons for the 10 tasks were obtained from 54 workers. For each pair of tasks, we aggregated relative subjectivity judgments made on this pair to determine which task was perceived as more subjective (i.e., less objective). To produce a ranking of the  subjectivity of the 10 tasks, we constructed a directed graph based on the pairwise subjectivity comparisons---each task was a node in this graph, and directed edges were added between each pair of tasks, pointing from the one that was deemed as more subjective (on the aggregate level) to the one deemed as less subjective. 
The topological sort algorithm~\cite{cormen2022introduction} was then applied to this directed graph to obtain a linear ordering of the nodes. If a cycle was detected within the graph, the corresponding tasks were considered to have the same level of subjectivity
and were merged into a single meta-node before re-runing the algorithm. 
Our final task subjectivity ranking results are shown in  Table~\ref{tab:results}.

\subsection{Model Training}
Given a text classification task, following the procedures outlined in Sections~\ref{zero-shot} and~\ref{few-shot}, 3,000 synthetic data points were generated for each candidate label under both zero-shot and few-shot settings. 
We then trained classification models using the real-world training data provided by the original dataset, the synthetic data generated under the zero-shot settings, and the synthetic data generated under the few-shot settings\footnote{Under the few-shot setting, we randomly sampled $10\%$ of the data points from the real-world training data provided in the original dataset as the example pool to guide the LLM's synthetic data generation process, but only the sythetic data generated were used to train the models.}, respectively. 
Specifically, we utilized the pre-trained BERT~\cite{devlin2018bert} and RoBERTa~\cite{liu2019roberta} models from Huggingface's transformers library~\cite{wolf-etal-2020-transformers} as the encoders, and used the representation embeddings from the last layer of these models as the input to our classification models. The classification model itself comprised a hidden layer of 768 units and an output layer, and it was fine-tuned with a learning rate of $5e-5$ and a batch size of 64. For datasets that provided official partitions for training and test sets, we directly evaluated the classification model's performance on the test sets. Otherwise, we randomly divided the dataset into training (70\%), validation (5\%), and test (25\%) sets\footnote{To ensure a fair comparison, we maintained an equal size for both the real-world and synthetic training data by downsampling the dataset with a larger size. 

}.  
Models' performance was evaluated via Macro-F1 and Accuracy scores, and they were computed by comparing the model's predictions with the gold labels provided in the test sets. 
To ensure the robustness of our results, all experiments were repeated three times, and the average performance across these repetitions was reported. 

\begin{table*}[t]
  \centering
  \resizebox{\textwidth}{!}{
  \begin{tblr}{
    row{2} = {c},
    row{3} = {c},
    column{2} = {l},
    cell{1}{1} = {r=3}{},
    cell{1}{2} = {r=3}{},
    cell{1}{3} = {c=6}{c},
    cell{1}{9} = {c=6}{c},
    cell{2}{3} = {c=2}{},
    cell{2}{5} = {c=2}{},
    cell{2}{7} = {c=2}{},
    cell{2}{9} = {c=2}{},
    cell{2}{11} = {c=2}{},
    cell{2}{13} = {c=2}{},
    cell{4}{3} = {c},
    cell{4}{4} = {c},
    cell{4}{5} = {c},
    cell{4}{6} = {c},
    cell{4}{7} = {c},
    cell{4}{8} = {c},
    cell{4}{9} = {c},
    cell{4}{10} = {c},
    cell{4}{11} = {c},
    cell{4}{12} = {c},
    cell{4}{13} = {c},
    cell{4}{14} = {c},
    cell{5}{3} = {c},
    cell{5}{4} = {c},
    cell{5}{5} = {c},
    cell{5}{6} = {c},
    cell{5}{7} = {c},
    cell{5}{8} = {c},
    cell{5}{9} = {c},
    cell{5}{10} = {c},
    cell{5}{11} = {c},
    cell{5}{12} = {c},
    cell{5}{13} = {c},
    cell{5}{14} = {c},
    cell{6}{3} = {c},
    cell{6}{4} = {c},
    cell{6}{5} = {c},
    cell{6}{6} = {c},
    cell{6}{7} = {c},
    cell{6}{8} = {c},
    cell{6}{9} = {c},
    cell{6}{10} = {c},
    cell{6}{11} = {c},
    cell{6}{12} = {c},
    cell{6}{13} = {c},
    cell{6}{14} = {c},
    cell{7}{3} = {c},
    cell{7}{4} = {c},
    cell{7}{5} = {c},
    cell{7}{6} = {c},
    cell{7}{7} = {c},
    cell{7}{8} = {c},
    cell{7}{9} = {c},
    cell{7}{10} = {c},
    cell{7}{11} = {c},
    cell{7}{12} = {c},
    cell{7}{13} = {c},
    cell{7}{14} = {c},
    cell{8}{3} = {c},
    cell{8}{4} = {c},
    cell{8}{5} = {c},
    cell{8}{6} = {c},
    cell{8}{7} = {c},
    cell{8}{8} = {c},
    cell{8}{9} = {c},
    cell{8}{10} = {c},
    cell{8}{11} = {c},
    cell{8}{12} = {c},
    cell{8}{13} = {c},
    cell{8}{14} = {c},
    cell{9}{3} = {c},
    cell{9}{4} = {c},
    cell{9}{5} = {c},
    cell{9}{6} = {c},
    cell{9}{7} = {c},
    cell{9}{8} = {c},
    cell{9}{9} = {c},
    cell{9}{10} = {c},
    cell{9}{11} = {c},
    cell{9}{12} = {c},
    cell{9}{13} = {c},
    cell{9}{14} = {c},
    cell{10}{3} = {c},
    cell{10}{4} = {c},
    cell{10}{5} = {c},
    cell{10}{6} = {c},
    cell{10}{7} = {c},
    cell{10}{8} = {c},
    cell{10}{9} = {c},
    cell{10}{10} = {c},
    cell{10}{11} = {c},
    cell{10}{12} = {c},
    cell{10}{13} = {c},
    cell{10}{14} = {c},
    cell{11}{3} = {c},
    cell{11}{4} = {c},
    cell{11}{5} = {c},
    cell{11}{6} = {c},
    cell{11}{7} = {c},
    cell{11}{8} = {c},
    cell{11}{9} = {c},
    cell{11}{10} = {c},
    cell{11}{11} = {c},
    cell{11}{12} = {c},
    cell{11}{13} = {c},
    cell{11}{14} = {c},
    cell{12}{3} = {c},
    cell{12}{4} = {c},
    cell{12}{5} = {c},
    cell{12}{6} = {c},
    cell{12}{7} = {c},
    cell{12}{8} = {c},
    cell{12}{9} = {c},
    cell{12}{10} = {c},
    cell{12}{11} = {c},
    cell{12}{12} = {c},
    cell{12}{13} = {c},
    cell{12}{14} = {c},
    cell{13}{3} = {c},
    cell{13}{4} = {c},
    cell{13}{5} = {c},
    cell{13}{6} = {c},
    cell{13}{7} = {c},
    cell{13}{8} = {c},
    cell{13}{9} = {c},
    cell{13}{10} = {c},
    cell{13}{11} = {c},
    cell{13}{12} = {c},
    cell{13}{13} = {c},
    cell{13}{14} = {c},
    vline{2-4,9} = {1-3}{},
    vline{4,6,10,12} = {2}{dashed},
    vline{8} = {2}{},
    vline{4,6,8,10,12,14} = {3-13}{dotted},
    vline{5,7,11,13} = {2-13}{dashed},
    vline{9} = {3}{},
    vline{2-3,9} = {4-13}{},
    hline{1,4,14} = {-}{},
    hline{2-3} = {3-14}{},
  }
  \textbf{Dataset}                  & \textbf{Subjectivity} & \textbf{BERT}           &                         &                                       &                                       &                                       &                                       & \textbf{RoBERTa}        &                         &                                       &                                       &                                       &                                       \\
                                    &                             & \textbf{Real-world data}       &                         & \textbf{Zero-shot setting}                    &                                       & \textbf{Few-shot setting}       &                                       & \textbf{Real-world data}       &                         & \textbf{Zero-shot setting}                    &                                       & \textbf{Few-shot setting}       &                                       \\
                                    &                             & \textbf{Macro-F1 } & \textbf{Accuracy Score} & \textbf{Macro-F1 }               & \textbf{Accuracy Score}               & \textbf{Macro-F1 }               & \textbf{Accuracy Score}               & \textbf{Macro-F1} & \textbf{Accuracy Score} & \textbf{Macro-F1}               & \textbf{Accuracy Score}               & \textbf{Macro-F1}               & \textbf{Accuracy Score}               \\
  AG                         & \OB                         & 95.3\%                  & 95.3\%                  & 89.3\% \textcolor{pf1color}{(-6.0\%)}  & 89.3\% \textcolor{pf1color}{(-6.0\%)}  & 91.5\% \textcolor{pf1color}{(-3.8\%)}  & 91.6\% \textcolor{pf1color}{(-3.7\%)}  & 94.6\%                  & 94.6\%                  & 88.6\% \textcolor{pf1color}{(-6.0\%)}  & 88.6\% \textcolor{pf1color}{(-6.0\%)}  & 92.9\% \textcolor{pf1color}{(-1.7\%)}  & 92.9\% \textcolor{pf1color}{(-1.7\%)}  \\
  Relation                    & \OB\OB                      & 98.6\%                  & 98.6\%                  & 92.4\% \textcolor{pf1color}{(-6.2\%)}  & 92.7\% \textcolor{pf1color}{(-5.9\%)}  & 96.4\% \textcolor{pf1color}{(-2.2\%)}  & 96.4\% \textcolor{pf1color}{(-2.2\%)}  & 97.0\%                  & 96.9\%                  & 91.4\% \textcolor{pf1color}{(-5.6\%)}  & 91.6\% \textcolor{pf1color}{(-5.3\%)}  & 94.1\% \textcolor{pf1color}{(-2.9\%)}  & 94.1\% \textcolor{pf1color}{(-2.8\%)}  \\
  IMDB                   & \OB\OB\OB                   & 87.6\%                  & 87.6\%                  & 81.2\% \textcolor{pf1color}{(-6.4\%)}  & 81.5\% \textcolor{pf1color}{(-6.1\%)}  & 81.1\% \textcolor{pf1color}{(-6.5\%)}  & 81.2\% \textcolor{pf1color}{(-6.4\%)}  & 89.0\%                  & 89.0\%                  & 81.2\% \textcolor{pf1color}{(-7.8\%)}  & 81.3\% \textcolor{pf1color}{(-7.7\%)}  & 82.4\% \textcolor{pf1color}{(-1.6\%)}  & 82.4\% \textcolor{pf1color}{(-1.6\%)}  \\
  SMS spam                          & \OB\OB\OB\OB                & 97.2\%                  & 98.8\%                  & 93.8\% \textcolor{pf1color}{(-3.4\%)}  & 95.1\% \textcolor{pf1color}{(-3.7\%)}  & 94.3\% \textcolor{pf1color}{(-2.9\%)} & 94.8\% \textcolor{pf1color}{(-4.0\%)} & 97.3\%                  & 98.8\%                  & 93.5\% \textcolor{pf1color}{(-3.8\%)}  & 95.9\% \textcolor{pf1color}{(-2.9\%)}  & 94.0\% \textcolor{pf1color}{(-3.3\%)}  & 95.7\% \textcolor{pf1color}{(-3.1\%)}  \\

  Reddit emotion                      & \OB\OB\OB\OB\OB             & 93.7\%                  & 94.6\%                  & 72.7\% \textcolor{pf1color}{(-21.0\%)} & 74.4\% \textcolor{pf1color}{(-20.2\%)} & 81.9\% \textcolor{pf1color}{(-11.8\%)} & 82.0\% \textcolor{pf1color}{(-12.6\%)} & 91.3\%                  & 92.1\%                  & 77.9\% \textcolor{pf1color}{(-13.4\%)} & 78.1\% \textcolor{pf1color}{(-14.0\%)} & 87.5\% \textcolor{pf1color}{(-3.8\%)}  & 87.7\% \textcolor{pf1color}{(-4.4\%)}  \\
  Tweet irony   & \OB\OB\OB\OB\OB             & 72.2\%                  & 73.9\%                  & 63.4\% \textcolor{pf1color}{(-8.8\%)}  & 63.6\% \textcolor{pf1color}{(-10.3\%)} & 81.5\% \textcolor{f1color}{(+9.3\%)} & 81.9\% \textcolor{f1color}{(+8.0\%)} & 74.0\%                  & 75.5\%                  & 57.8\% \textcolor{pf1color}{(-16.2\%)} & 59.1\% \textcolor{pf1color}{(-16.4\%)} & 83.3\% \textcolor{f1color}{(+9.3\%)} & 83.7\% \textcolor{f1color}{(+8.2\%)} \\
  Tweet emotions & \OB\OB\OB\OB\OB             & 77.7\%                  & 81.1\%                  & 58.1\% \textcolor{pf1color}{(-19.6\%)} & 64.5\% \textcolor{pf1color}{(-16.6\%)} & 64.6\% \textcolor{pf1color}{(-13.1\%)} & 69.1\% \textcolor{pf1color}{(-12.0\%)} & 75.8\%                  & 78.9\%                  & 64.6\% \textcolor{pf1color}{(-11.2\%)} & 71.5\% \textcolor{pf1color}{(-7.4\%)}  & 66.3\% \textcolor{pf1color}{(-9.5\%)}  & 72.7\% \textcolor{pf1color}{(-6.2\%)}  \\
  Sarcasm              & \OB\OB\OB\OB\OB             & 89.9\%                  & 90.3\%                  & 51.1\% \textcolor{pf1color}{(-38.8\%)} & 51.2\% \textcolor{pf1color}{(-39.1\%)} & 63.6\% \textcolor{pf1color}{(-26.3\%)} & 64.8\% \textcolor{pf1color}{(-25.5\%)} & 91.8\%                  & 92.0\%                  & 54.3\% \textcolor{pf1color}{(-37.5\%)} & 54.3\% \textcolor{pf1color}{(-37.7\%)} & 61.5\% \textcolor{pf1color}{(-30.3\%)} & 63.6\% \textcolor{pf1color}{(-28.4\%)} \\
   Financial              & \OB\OB\OB\OB\OB             & 83.2\%                  & 84.6\%                  & 48.2\% \textcolor{pf1color}{(-35.0\%)} & 60.7\% \textcolor{pf1color}{(-23.9\%)} & 70.6\% \textcolor{pf1color}{(-12.6\%)} & 74.2\% \textcolor{pf1color}{(-10.4\%)} & 85.0\%                  & 86.6\%                  & 58.5\% \textcolor{pf1color}{(-26.5\%)} & 70.3\% \textcolor{pf1color}{(-16.3\%)} & 75.0\% \textcolor{pf1color}{(-10.0\%)} & 78.9\% \textcolor{pf1color}{(-7.7\%)}  \\
  Humor speech          & \OB\OB\OB\OB\OB             & 97.0\%                  & 97.0\%                  & 56.0\% \textcolor{pf1color}{(-41.0\%)} & 61.7\% \textcolor{pf1color}{(-35.3\%)} & 86.9\% \textcolor{pf1color}{(-10.1\%)} & 87.0\% \textcolor{pf1color}{(-10.0\%)} & 96.7\%                  & 96.7\%                  & 54.9\% \textcolor{pf1color}{(-41.8\%)} & 60.9\% \textcolor{pf1color}{(-35.8\%)} & 84.0\% \textcolor{pf1color}{(-12.7\%)} & 84.0\% \textcolor{pf1color}{(-12.7\%)} 
  \end{tblr}
  }
  \caption{Comparing the performance of classification models trained on the LLM-generated synthetic data under the zero-shot or few-shot settings, with those trained with the original real-world data, in terms of Macro-F1 (\%) and Accuracy Score (\%). In the ``Subjectivity'' column, more "$\star$" symbols indicate a higher level of task subjectivity.  }
  \label{tab:results}
\end{table*}

\subsection{Evaluation Results}
Table~\ref{tab:results} summarizes the comparative performance  of classification models trained with different data. Below, we highlight a few key observations we get from this comparison. 

\noindent \textbf{Models trained on the real-world data consistently outperform those trained on the synthetic data.} Our results indicate that models trained on the original real-world data consistently outperform their counterparts trained on the synthetic data generated under either zero-shot or few-shot settings, almost for every task. In particular, with the RoBERTa model, we observe that the average improvements of the model trained on the real-world data 
over 
the models trained on zero-shot synthetic data and few-shot synthetic data
are $16.9\%$ and $6.7\%$ in terms of Macro-F1, and $14.9\%$ and  $6.1\%$ in terms of accuracy. 
Similar trends are observed with the BERT model as well.  

\noindent \textbf{Guiding LLM with real-world data examples can boost the effectiveness of the synthetic data.}
We also observe that models trained on those synthetic data generated under the few-shot settings almost always outperform those trained on the synthetic data generated under the zero-shot settings. For instance, for the BERT model, we see an average increase of $10.6\%$ and $8.8\%$ in Macro-F1 and accuracy scores, respectively, across the 10 tasks in the few-shot setting, as compared to the zero-shot setting. Similarly, with the RoBERTa model, there is an average increase of $10.3\%$ in Macro-F1 and $8.9\%$ in accuracy scores across the 10 tasks when the real-world data are used as examples for LLM to mimic in the synthetic data generation process. For more analysis of the few-shot synthetic data, please see App.~\ref{e1a1} and \ref{e1a3}.

\noindent \textbf{Synthetic data support more effective model training for tasks that are less subjective.} Finally, we notice that for classification tasks with relatively low levels of subjectivity (e.g., those in the AG's news, Relation classification, IMDB reviews, and SMS spam datasets), the performance difference between models trained on the synthetic data and those trained on the real-world data is remarkably small. 
However, for tasks with high subjectivity, the performance decrease resulted from the usage of the synthetic data is more significant---for instance, across the cluster of 6 tasks with the highest level of subjectivity in our evaluation, there is an average decrease of 27.4\% and 24.2\% in Macro-F1 and accuracy, respectively, comparing the BERT models trained on the zero-shot synthetic data with those trained on the real-world data. In other words, for text classification tasks that are highly objective, there is great potential in training high-performing models simply based on synthetic data generated by LLMs, but the same method falls short in generating synthetic data that can effectively support model training for highly subjective classifications.

\subsection{Exploratory Analysis: Data Diversity}
\label{sec:exploratory}
\begin{figure}[t]
  \centering
  \subfloat[Remote Clique]{\includegraphics[width=0.24\textwidth]{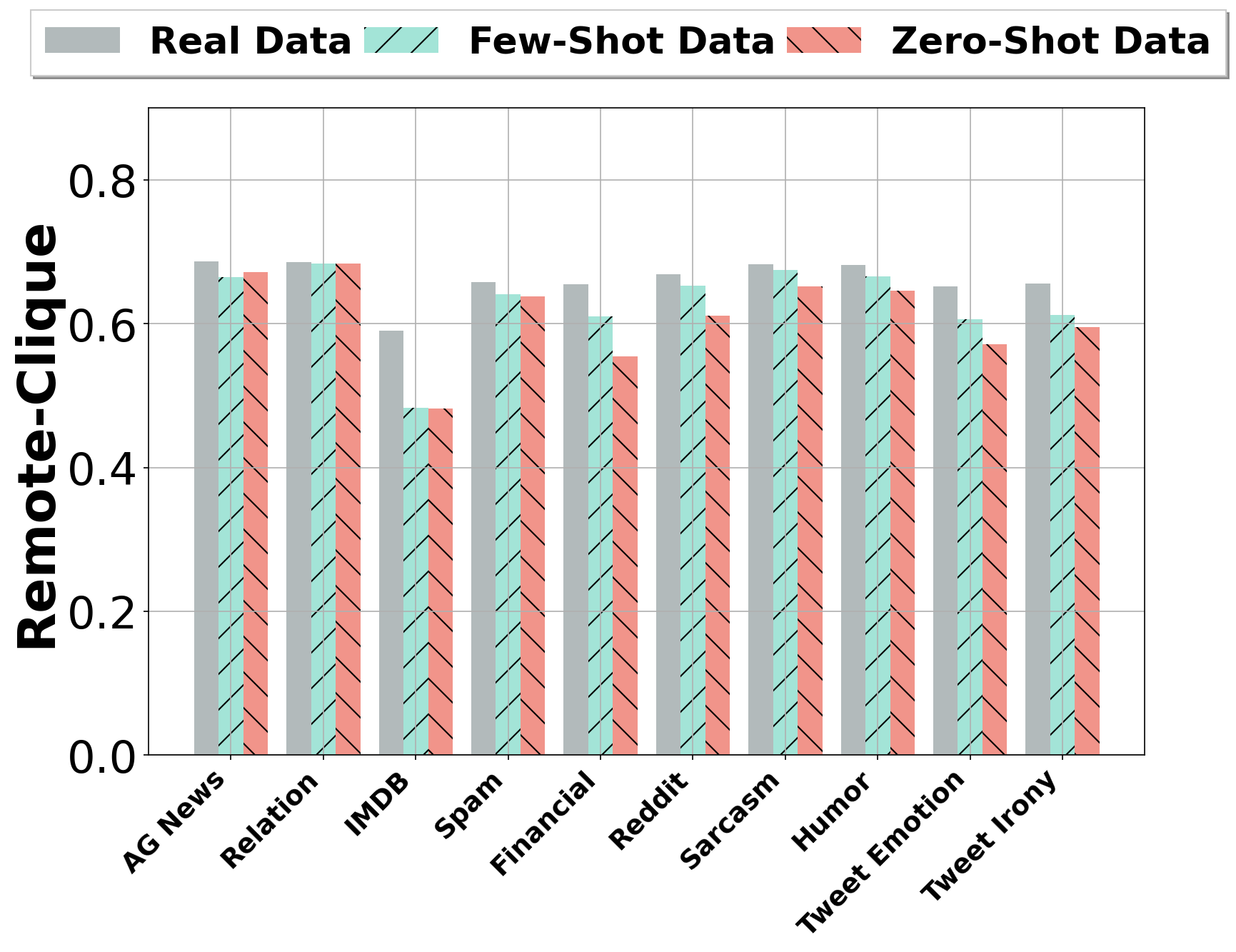}\label{fig:remote_clique}}
  \hfill
  \subfloat[Chamfer Distance]{\includegraphics[width=0.24\textwidth]{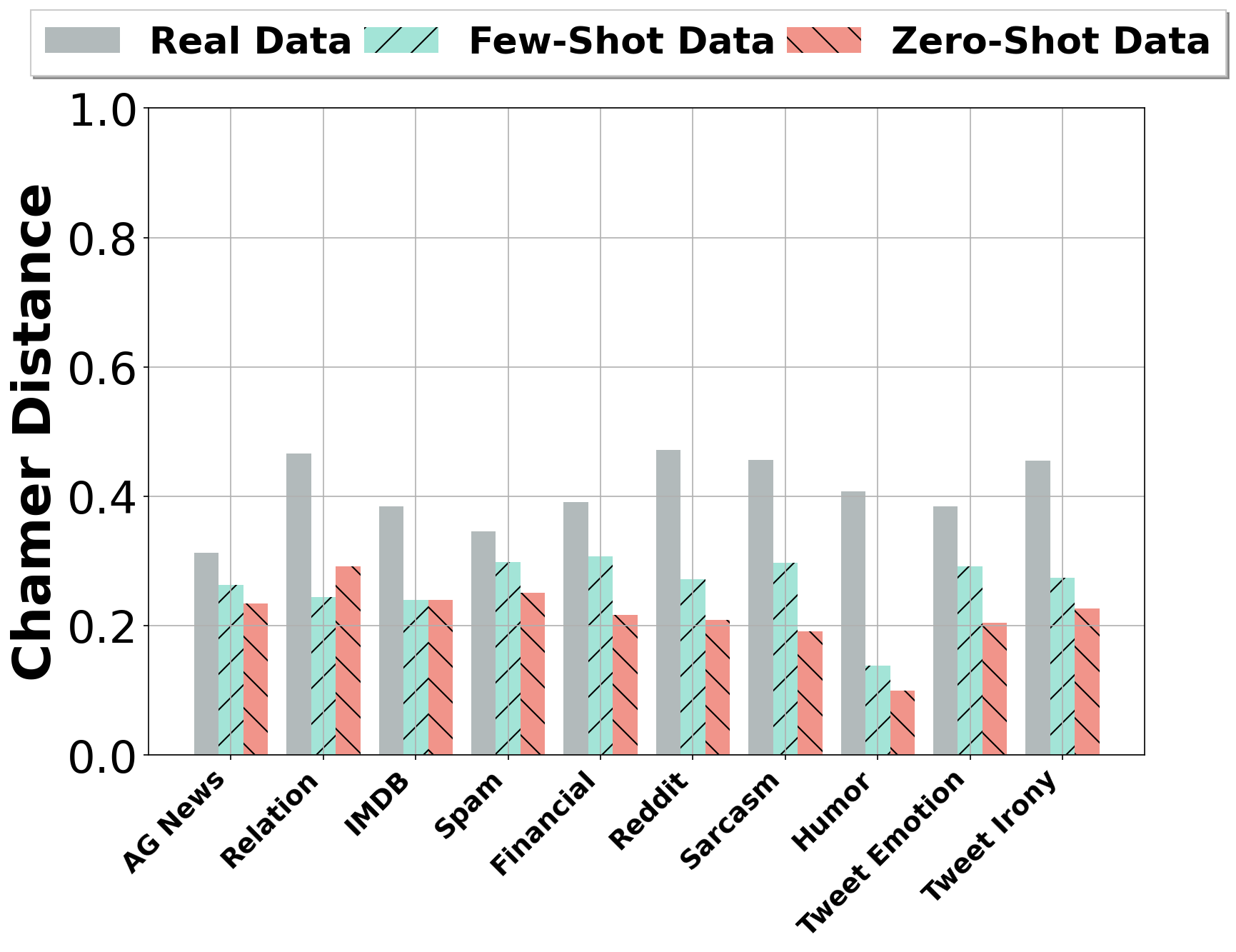}\label{fig:chamfer}}
  \caption{Comparing the diversity of the real-world data and the synthetic data.}
  \label{fig: diversity}
\end{figure}

To explore the potential reasons underlying the model performance difference, we conducted an exploratory analysis on the diversity of the training data. Following \citeauthor{10.1145/3411764.3445782}~\shortcite{10.1145/3411764.3445782}, we used the {\em Remote Clique Score} (i.e., the average mean distance of a data instance to other instances) and the {\em Chamfer Distance Score} (i.e., the average minimum distance of a data instance to other instances) to quantify the diversity of a set of data. 
For both metrics, higher values indicate greater data diversity. 
As shown in Figure~\ref{fig: diversity}, we find that in general, the real-world data appear to be more diverse than the synthetic data generated under the few-shot settings, which in turn seem to be more diverse than the zero-shot synthetic data. This might partially explain why models trained on the real-world data and the few-shot synthetic data tend to outperform those trained on the zero-shot synthetic data. 

In addition, we also notice that compared to that on the low subjectivity tasks (i.e., AG, Relation, IMDB, Spam), the differences in data diversity between the real-world data and the synthetic data seem to be more salient on the high subjectivity tasks (i.e., the other 6 tasks), especially in terms of the Chamfer Distance Score. In fact, a t-test shows that the decrease of the Chamfer Distance Score in the zero-shot synthetic data compared to the real data is significantly larger for the high subjectivity tasks than for the low subjectivity tasks ($p<0.01$). 
This suggests that for tasks with high subjectivity, such as interpreting humor or sarcasm in language, LLMs may not be able to 
generate data instances that can cover the full spectrum of real-life scenarios, which may limit the performance of models trained on the synthetic data. 

\section{Evaluation \RNum{2}: Comparison Across Different Task Instances}

\begin{table*}
\centering
\resizebox{\linewidth}{!}{
\begin{tblr}{
  cells = {c},
  hline{1-2,5} = {-}{},
}
\textbf{Dataset} & \textbf{AG} & \textbf{Relation} & \textbf{IMDB}  & \textbf{SMS Spam}  & \textbf{Reddit Emotion}  &\textbf{Humor Speech} &\textbf{Tweet Irony} & \textbf{Sarcasm }   & \textbf{Tweet Emotions}  & \textbf{Finanical} \\
Average Agreement $\overline{a}$     & 0.80 (4.2)  & 0.78 (4.5) & 0.76 (7.3)  & 0.73 (8.5) & 0.69 (6.6)  & 0.68 (7.1) & 0.68 (6.7)  & 0.64 (7.7) &  0.64 (4.6) & 0.57 (7.6) \\
Krippendorff's $\alpha$   & 0.51  & 0.43 & 0.19  & 0.27 & 0.30  & 0.06 & 0.03  & 0.01 &  0.17 & -0.03 \\
Subjectivity Level &\OB  &\OB\OB  & \OB\OB\OB &  \OB\OB\OB\OB & \OB\OB\OB\OB\OB  & \OB\OB\OB\OB\OB  &\OB\OB\OB\OB\OB  &\OB\OB\OB\OB\OB  & \OB\OB\OB\OB\OB &\OB\OB\OB\OB\OB  
\end{tblr}
}
\caption{The average instance-level annotation agreement for different types of tasks, 
alongside the corresponding task-level subjectivity. Numbers in parentheses in the first row
represent the average number of annotations received per task instance. Higher values for both the average agreement $\overline{a}$ and Krippendorff's $\alpha$ indicate a higher degree inter-annotator agreement.
}
\label{tab:hit2}
\end{table*}

In the previous section, we have discovered that the subjectivity of a task can adversely affect the performance of classification models trained on the LLM-generated synthetic data. However, even for the same type of task, the classification for each individual task instance may exhibits different levels of subjectivity as well. Naturally, one may wonder whether models trained on the LLM-generated synthetic data may show different performance on task instances of different subjectivity. We aim to explore the answers to this question in this section.

\subsection{Instance-level Subjectivity Determination}
Given a text classification task and a specific text instance, we consider the degree of {\em agreement among annotators} on the label of this text as a proxy for the subjectivity of this instance---a lower level of agreement means that annotators hold more divergent views, hence the task may have a higher level of subjectivity. Thus, to formally quantify the subjectivity of different instances for different tasks, we again conduct a crowdsourced study to collect instance-level annotations.

\noindent \textbf{Study procedure.} We again considered the 10 types of text classification tasks as that in the first evaluation study. For each type of task, we randomly sampled 50 text instances per category from the test set to compose our ``evaluation dataset'' for that task.
We then recruited U.S. workers from MTurk to complete annotation tasks for those instances in our evaluation dataset. Specifically,  each worker was randomly assigned to one type of text classification tasks. After going through a brief 
instruction of the assigned task, 
the worker was asked to complete 20 classification tasks of the assigned type to get a payment of \$1.2, where the texts presented in these 20 tasks were randomly sampled from the evaluation dataset for the assigned type of task.  
Again, we included two attention check questions in our study to filter out inattentive workers.  We ensured that each task instance received at least three annotations from unique MTurk workers. 

\noindent \textbf{Computing instance subjectivity.}
Based on annotations we obtained from attentive workers, 
we quantify the subjectivity level of each task instance using the fraction of annotators who agree with the majority label for the task instance, that is: 
\begin{equation}
    a_i = \frac{{\max}_{y \in \mathcal{Y}} \sum_{k=1}^{K_i} \mathds{1}{(r_{i}^{k} = y)}}{K_i}
\end{equation}
where $\mathcal{Y}=\{1,\cdot\cdot\cdot, Y\}$ is the set of all possible labels, $K_i$ is the total number of annotators who labeled instance $i$, and $r_{i}^{k}$ is the $k$-th annotator's annotation on instance $i$. Intuitively, a lower value of $a_i$ suggests that consensus is less likely to be reached among annotators on  instance $i$, thus instance $i$ may have a higher level of subjectivity.  In Table~\ref{tab:hit2}, we report the average values of
$a_i$
(i.e., $\overline{a}$) for instances in the evaluation datasets of different types of tasks, along with the average inter-annotator agreement on each task instance (as measured by the Krippendorff's $\alpha$) as well as the task-level subjectivity level for different types of tasks. We can see that 
$\overline{a}$ closely aligns with the Krippendorff's $\alpha$, and tasks with higher levels of subjectivity also exhibit a higher value of $\overline{a}$ in general, indicating that $a_i$ can potentially serve as a reasonable proxy for the subjectivity of each task instance.

\begin{figure*}[htbp]
  \centering
  \subfloat[AG]{\includegraphics[width=0.19\textwidth]{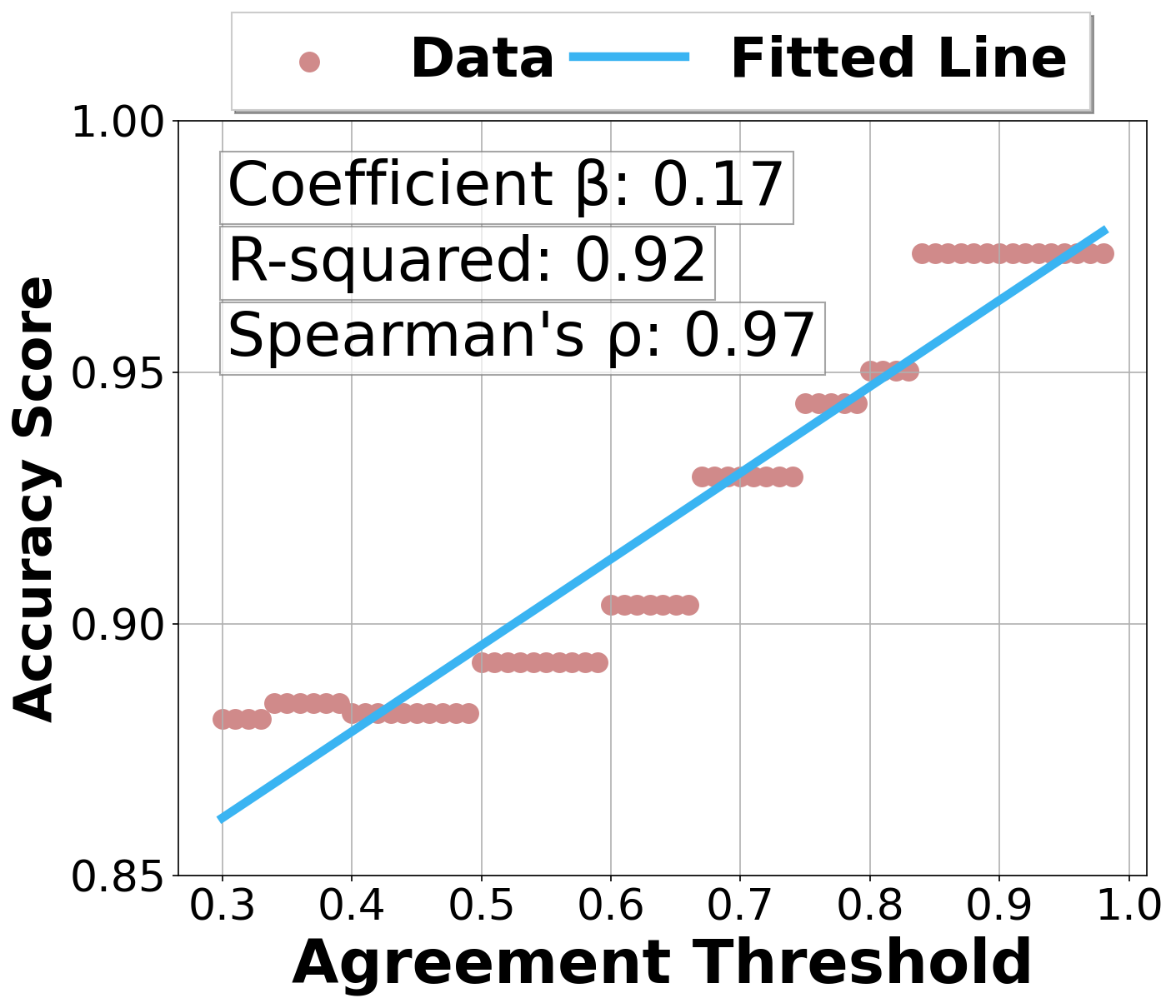}\label{fig:ag_line}}
  \hfill
  \subfloat[Relation]{\includegraphics[width=0.19\textwidth]{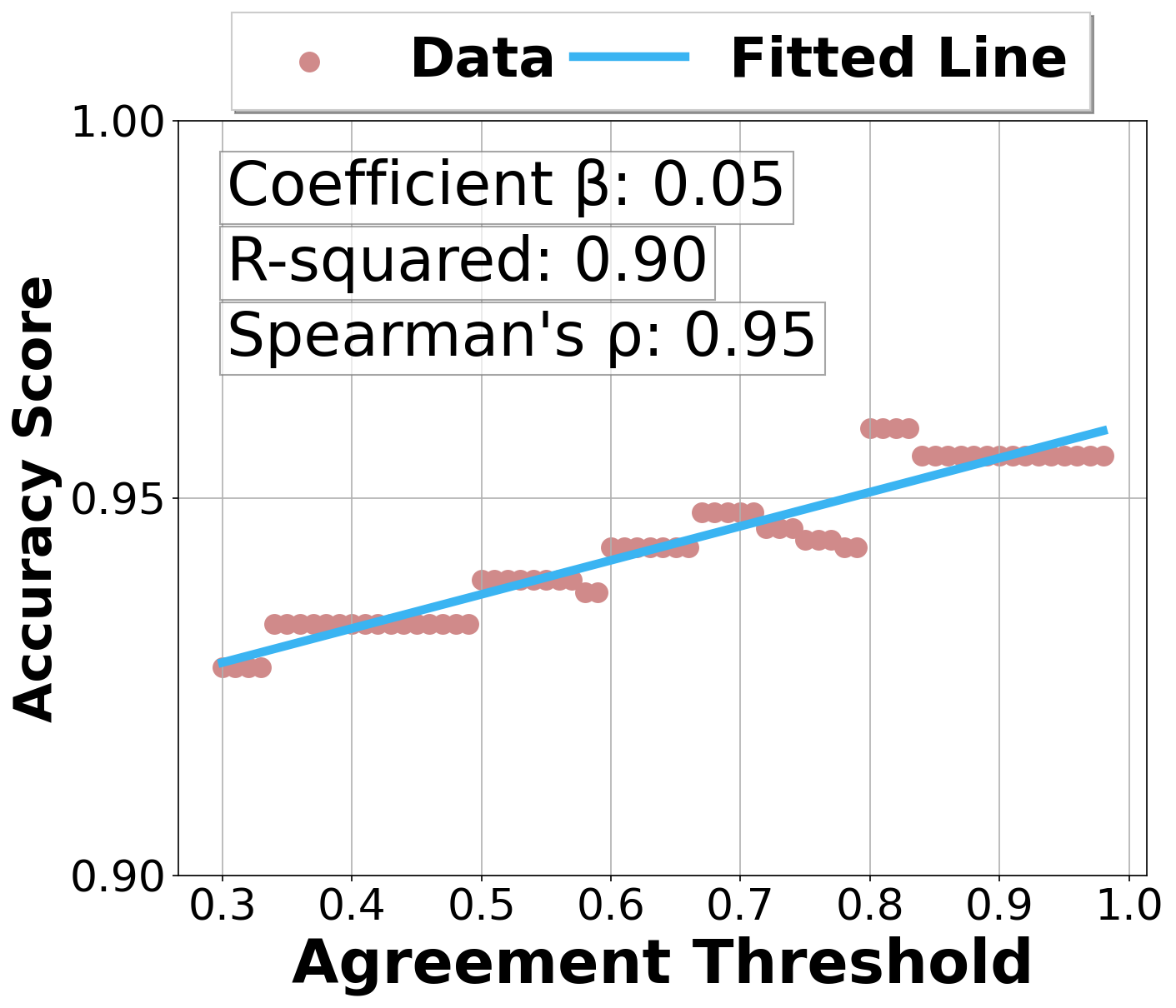}\label{fig:rel_line}}
  \hfill
  \subfloat[IMDB Reviews]{\includegraphics[width=0.19\textwidth]{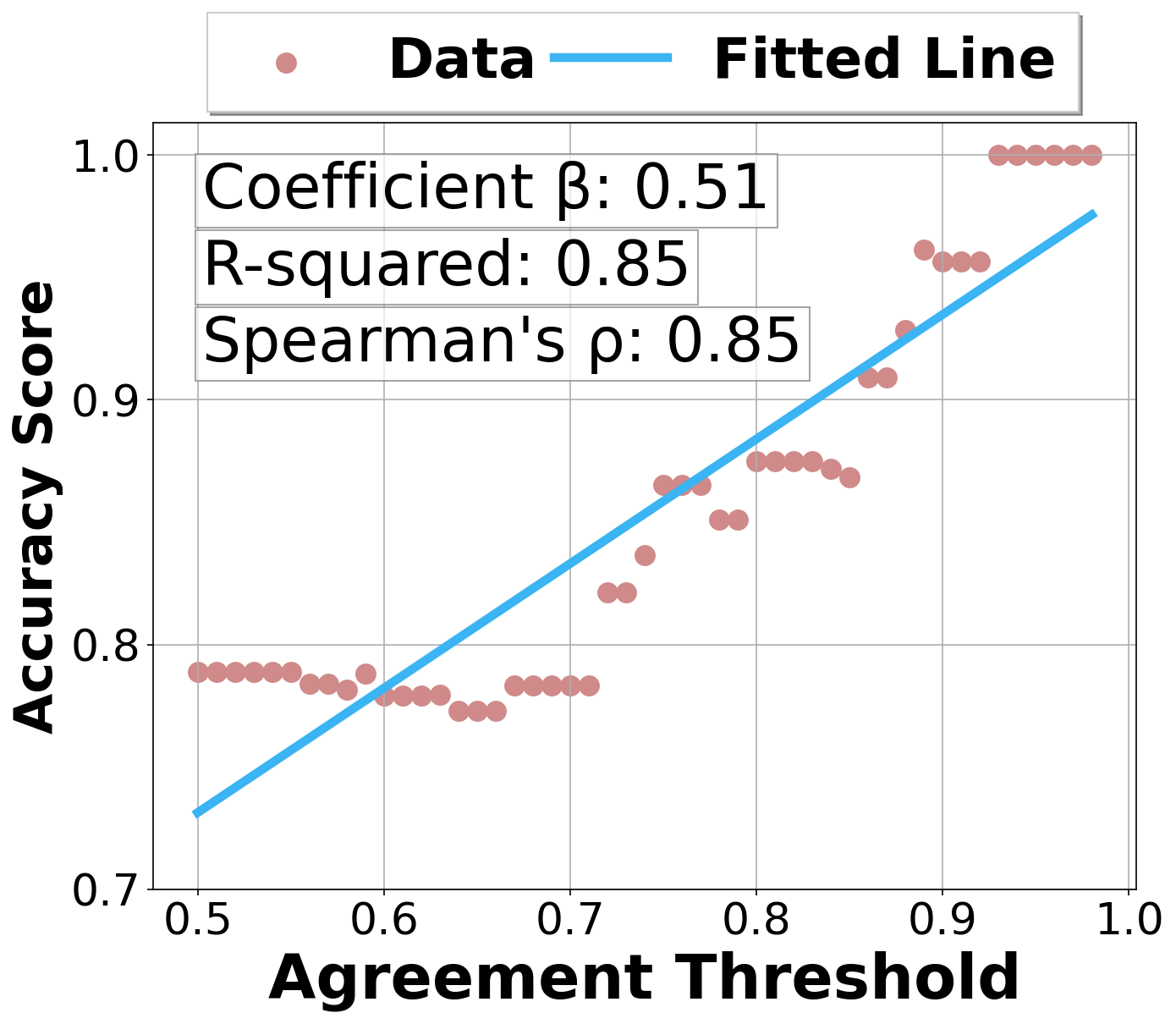}\label{fig:imdb_line}}
  \hfill
  \subfloat[SMS Spam]{\includegraphics[width=0.19\textwidth]{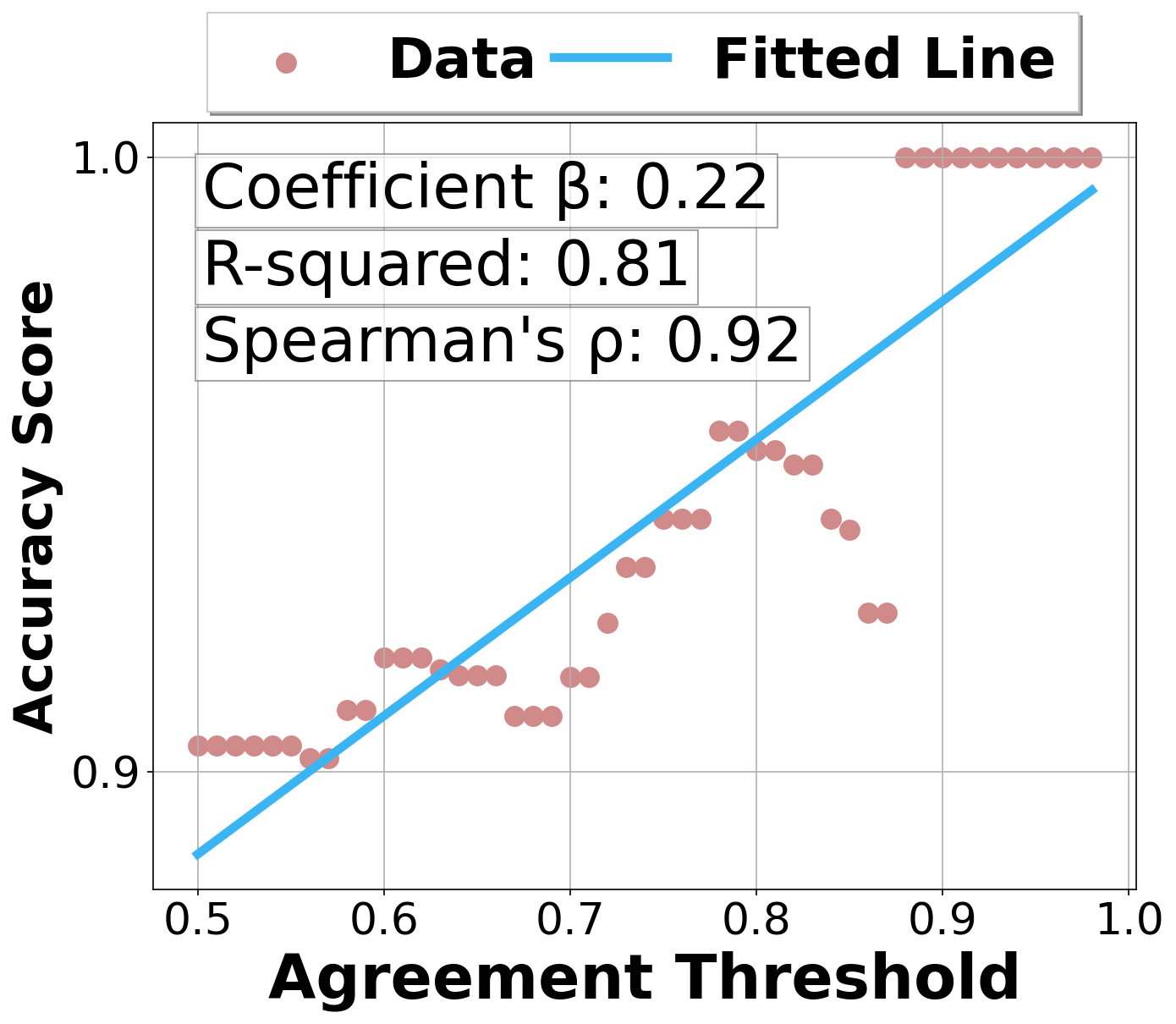}\label{fig:spam_line}}
  \hfill
  \subfloat[Reddit Emotion]{\includegraphics[width=0.19\textwidth]{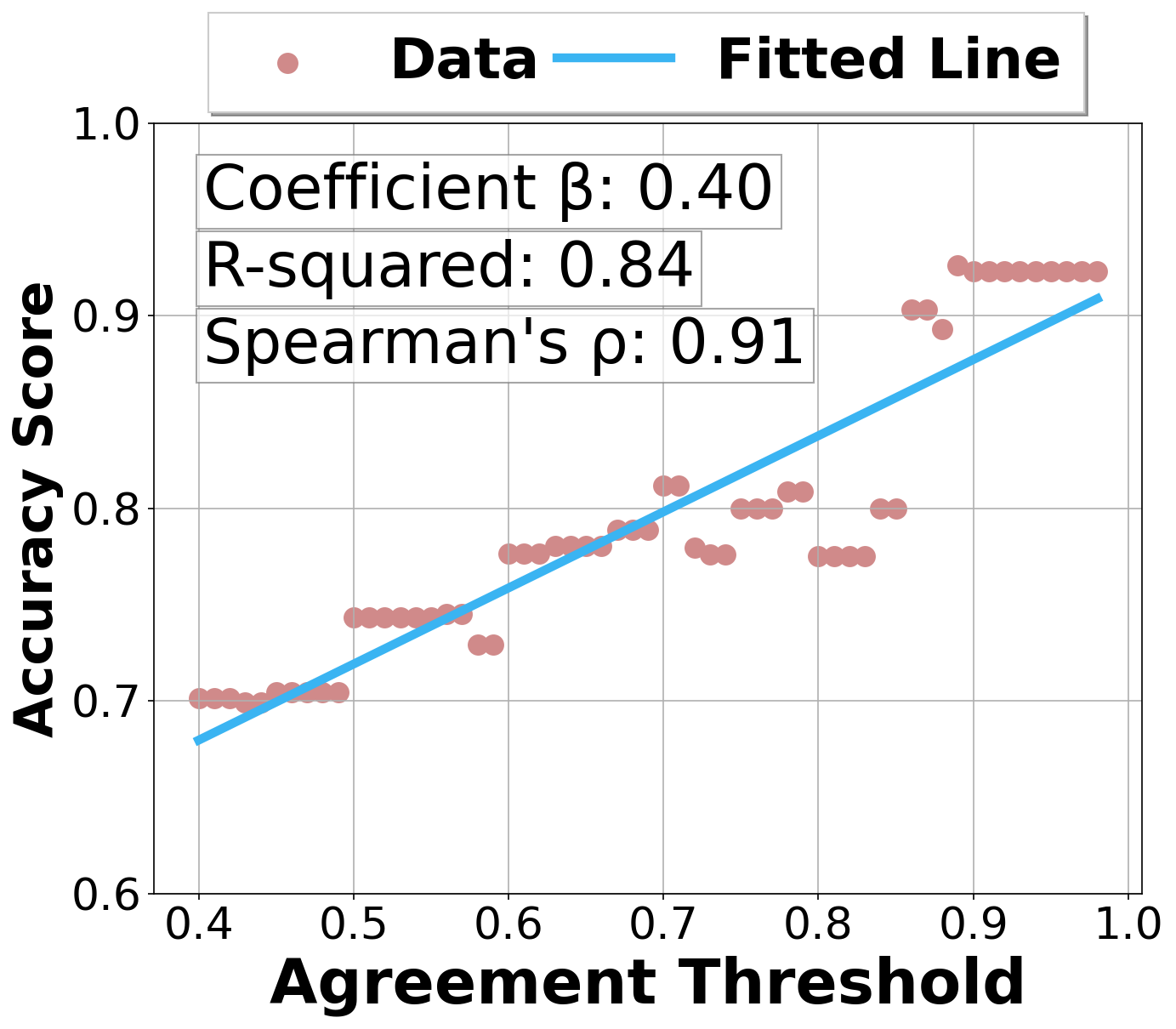}\label{fig:goemo_line}}
  \\
  \hfill
  \subfloat[Sarcasm News]{\includegraphics[width=0.19\textwidth]{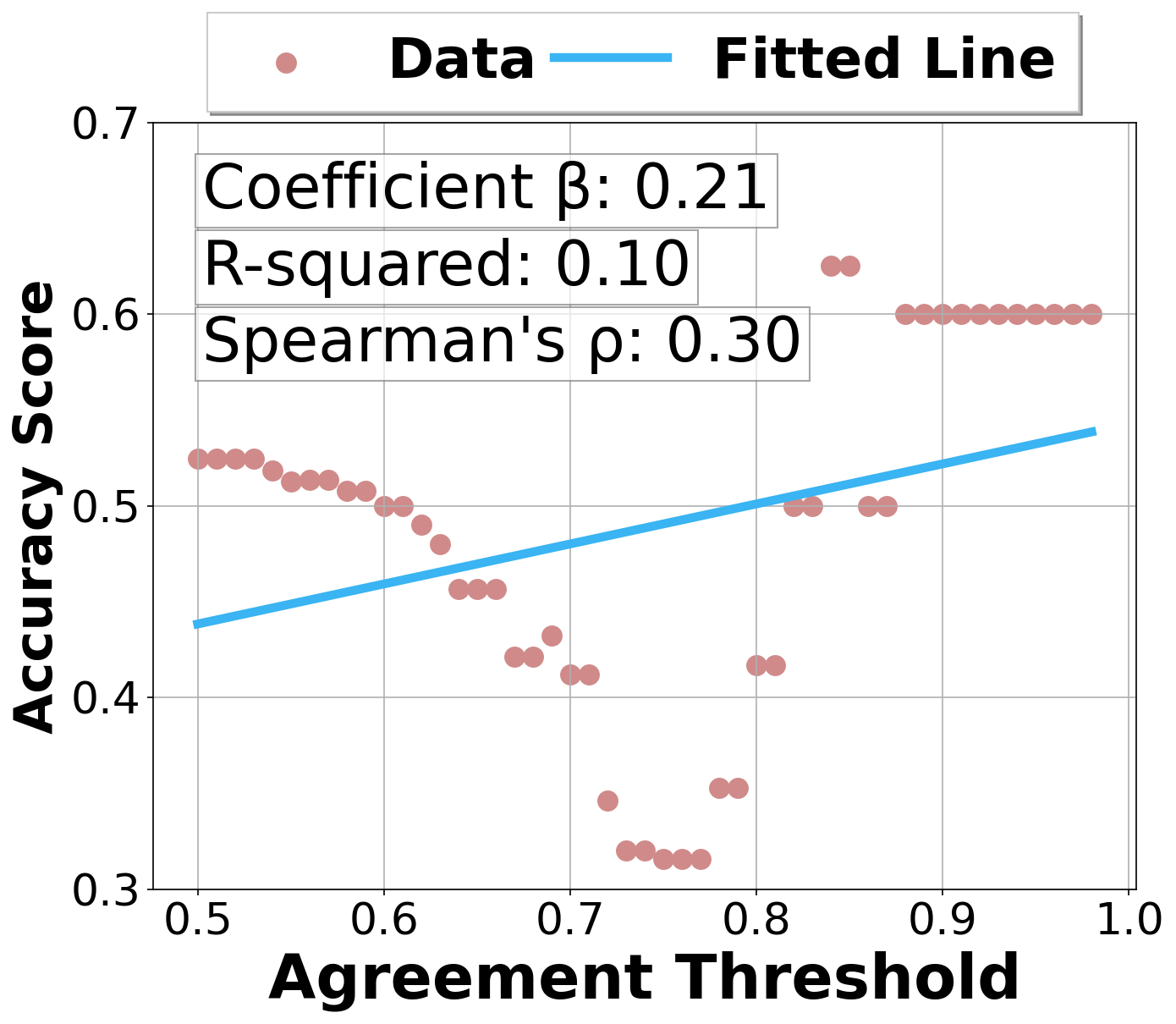}\label{fig:sarcasm_line}}
  \hfill
  \subfloat[Humor Detection]{\includegraphics[width=0.19\textwidth]{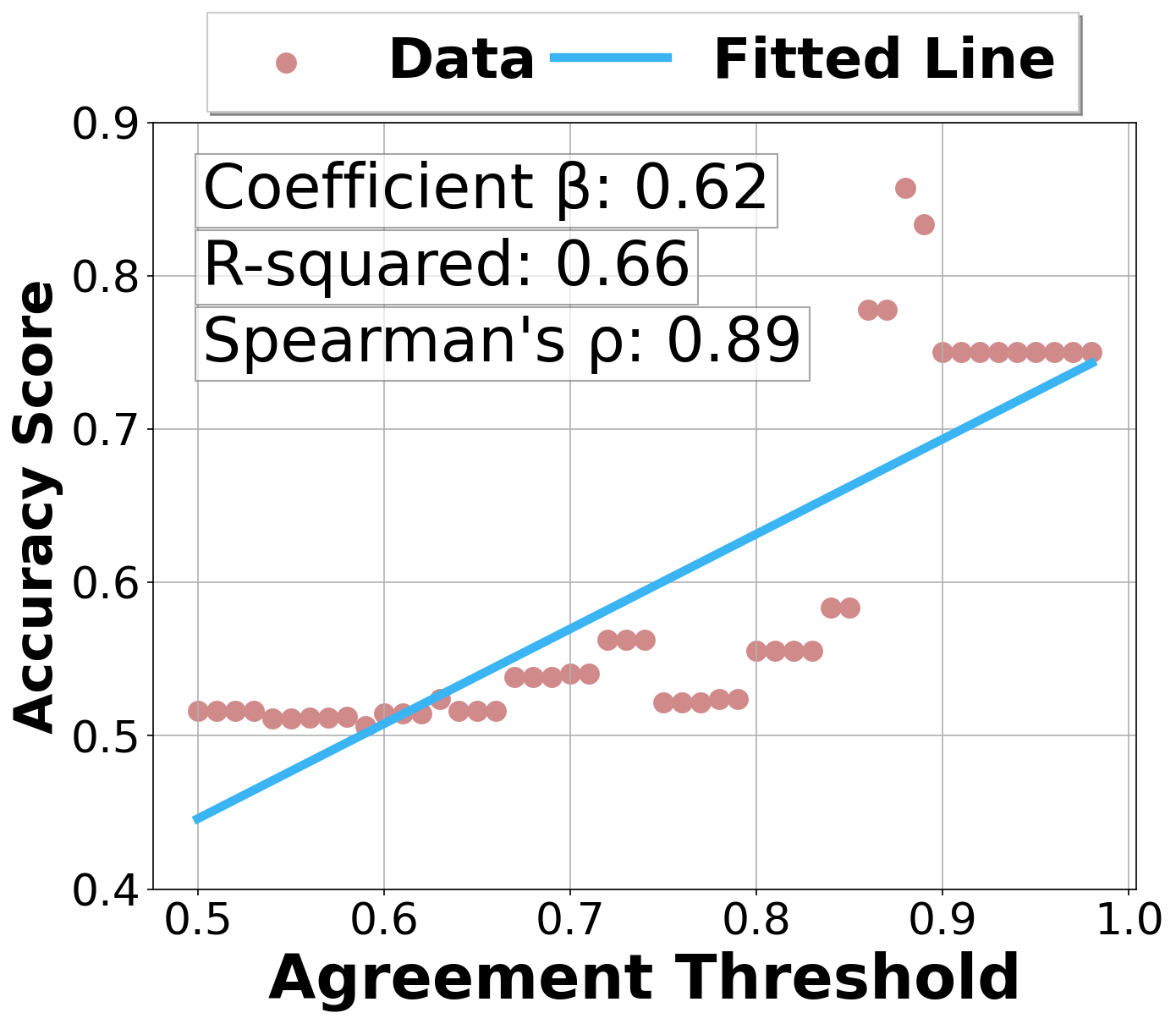}\label{fig:humor_line}}
  \hfill
  \subfloat[Tweet Emotions]{\includegraphics[width=0.19\textwidth]{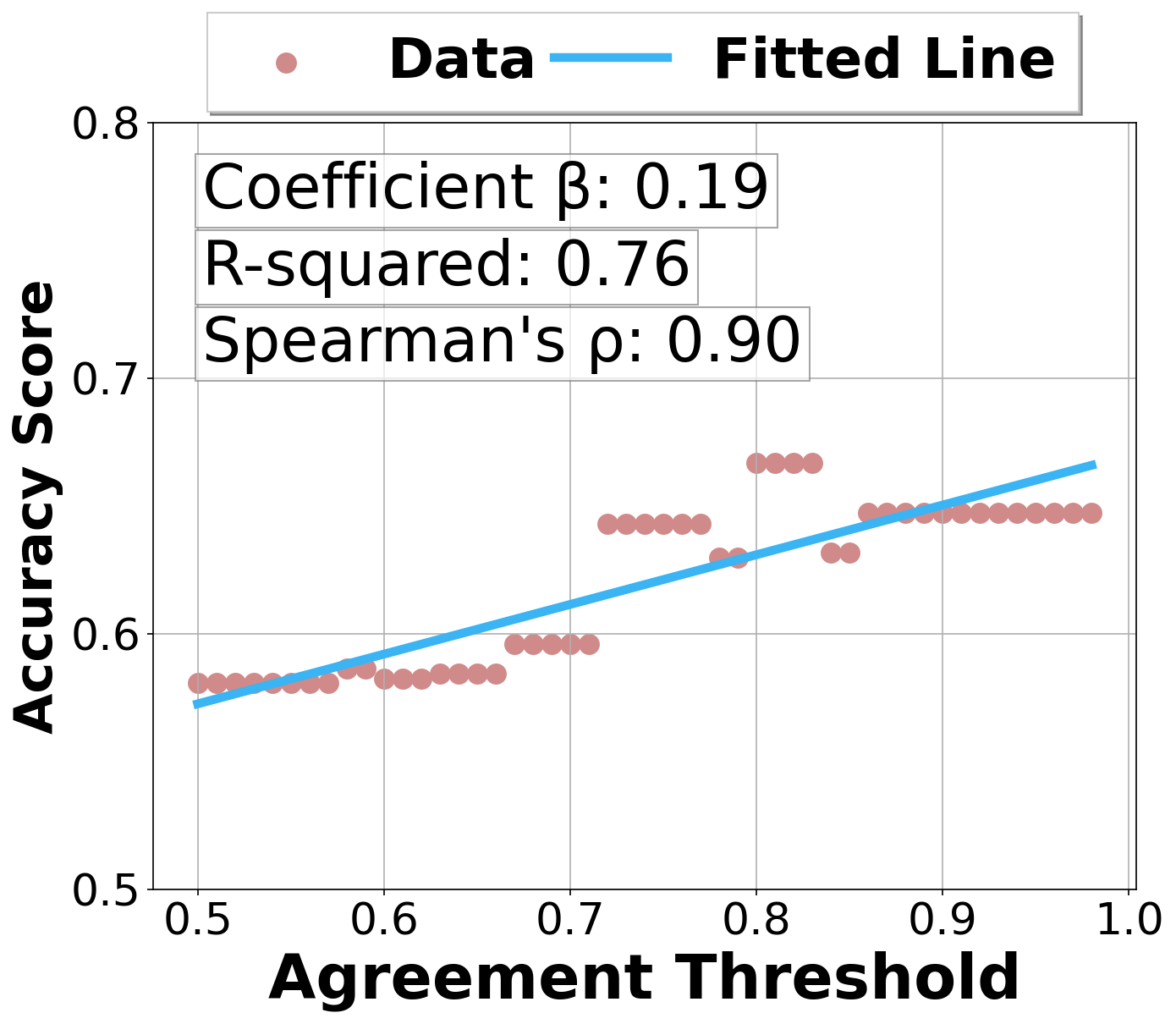}\label{fig:temo_line}}
  \hfill
  \subfloat[Tweet Irony Speech]{\includegraphics[width=0.19\textwidth]{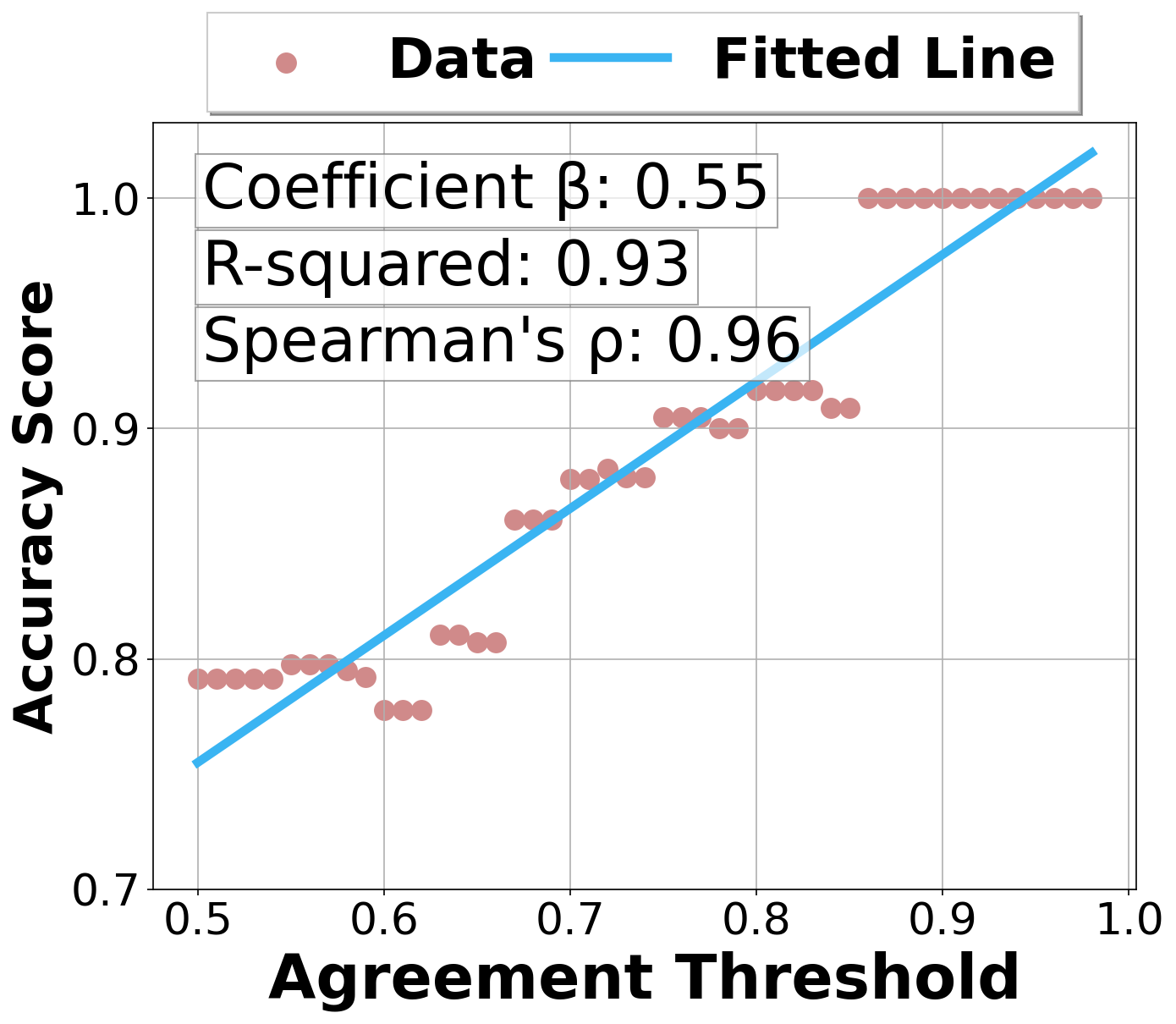}\label{fig:tirony_line}}
  \hfill
   \subfloat[Financial Phrasebank]{\includegraphics[width=0.19\textwidth]{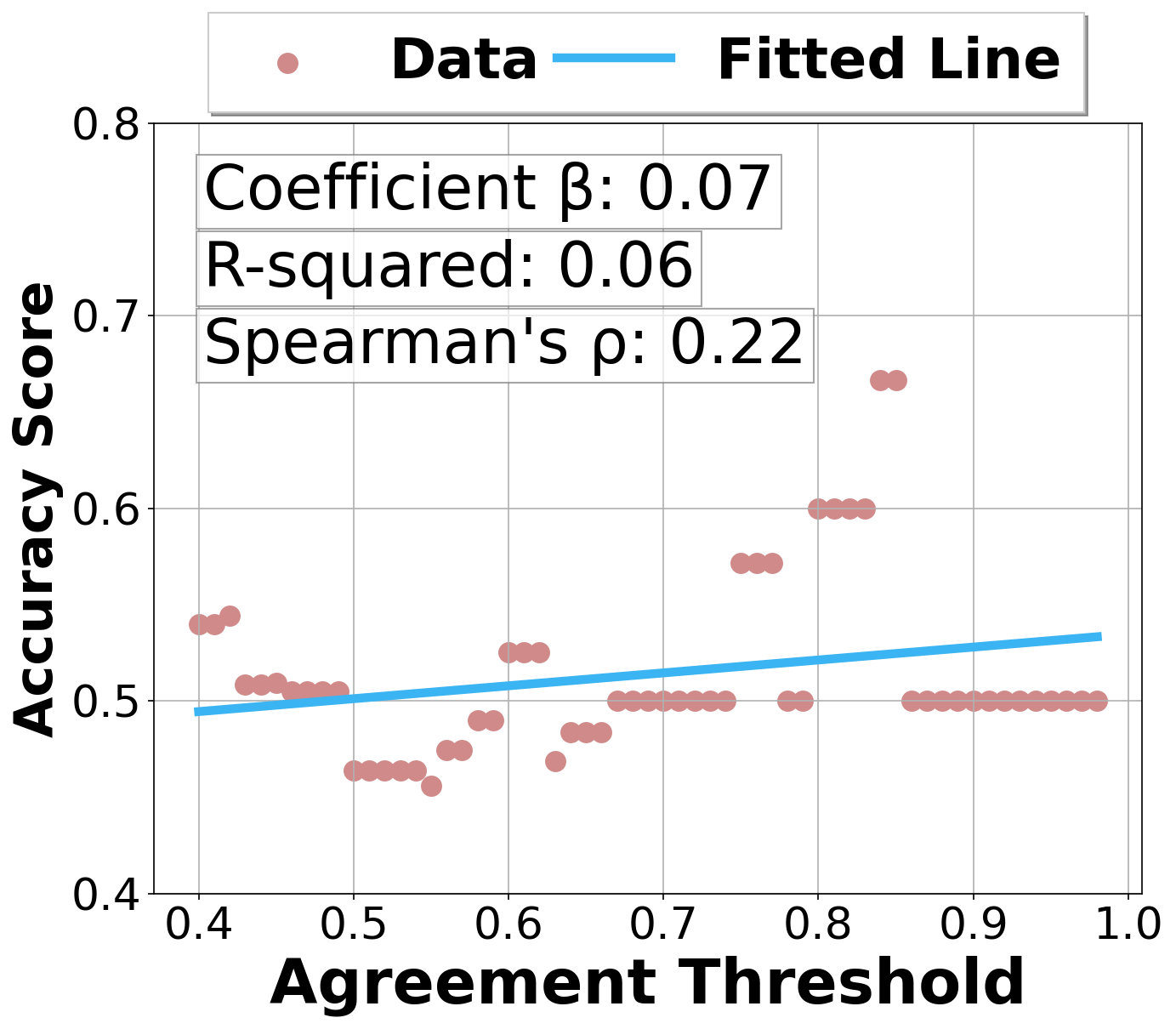}\label{fig:fin_line}}
  
  \caption{Changes in the accuracy of the BERT model trained on zero-shot synthetic data as the instance-level annotation agreement threshold varies.  
  The solid blue line in each plot is the linear regression fitted on the data,  and the $R$-squared score quantifies the goodness of fit. The Spearman's $\rho$ assesses the strength of rank correlation between the instance-level agreement threshold and the model accuracy for each task. Higher values for both $R$-squared and Spearman's $\rho$, ideally close to $1$, indicate a stronger monotonic relationship between the instance-level subjectivity and the model accuracy. }
  \label{fig:instance_corr}
  \vspace{-10pt}
\end{figure*}

\subsection{Evaluation Results}
\label{sec:eval2}

We now look into whether models trained on the LLM-generated synthetic data exhibit different performance on instances with different levels of subjectivity, and we focus on the models trained on zero-shot synthetic data in this evaluation. Specifically, given a classification task, we trained a BERT model using the zero-shot synthetic data and computed its accuracy on the subset of task instances in the evaluation dataset whose instance-level annotation agreement (i.e., $a_i$) exceeds a threshold $\gamma$, and we repeated this computation for many times as we varied the value of $\gamma$.

Figure~\ref{fig:instance_corr} illustrates how the model accuracy varies with the instance-level annotation agreement threshold $\gamma$ for different types of tasks. For most tasks (except for the tasks in the Scarcasm News and Finanical Phrasebank datasets), we observe a strong {\em monotonically increasing} relationship between $\gamma$ and the model accuracy, with correlations between them (i.e., $\beta$) being positive and  values of the Spearman's rank correlation coefficient $\rho$ often exceeding 0.85. Since increasing the instance-level annotation agreement threshold $\gamma$ effectively filters out task instances with high subjectivity, this observation suggests that models trained on synthetic data indeed tend to have varying performance on different instances---even within the same type of tasks, these models still perform better on those task instances with low subjectivity.

As a comparison, we also investigate into whether models trained on the real-world data exhibit similar behaviors. The detailed results are reported in App.~\ref{e2a}. On the high level, while we also observe the trend that these models' performance appears to increase as the instance-level task subjectivity decreases, such relationship is usually weaker than that illustrated in the models trained on the synthetic data (e.g., $\beta$ and $\rho$ are smaller).

\vspace{-8pt}

\section{Conclusions and Discussions}
\vspace{-4pt}
In this paper, we present an initial exploration into factors that moderate the effectiveness of LLM-generated synthetic data for facilitating the training of text classification models.  
Our results show that the performance of the models trained on synthetic data decreases both for classification tasks with higher levels of subjectivity and on task instances with higher subjectivity. In this section, we provide some potential explanations for the observations of our study, and discuss the implications, limitations, and future directions of our work.
\vspace{-2pt}
\subsection{Why subjectivity adversely impacts the effectiveness of the synthetic data?}
We provide a few explanations for why task subjectivity is found to be negatively associated with the performance of models trained on the LLM-generated synthetic data.
First, highly subjective tasks often require a deep understanding of nuanced human emotions and contextual subtleties, as well as the ability to discern and accurately interpret different perspectives. As such, LLMs may encounter limitations in generating data that can capture the extensive range and complexity of real-life use of language. Indeed, as shown in our exploratory analysis in Section~\ref{sec:exploratory}, the diversity of the LLM-generated synthetic data appears to be particularly limited on tasks with high subjectivity, when compared to the real-world data. This implies that one potential way to improve the effectiveness of synthetic data on high subjectivity tasks is to increase the data diversity and ensure the synthetic data can better reflect real-world data distributions.

Second, specific to the relationship between the instance-level subjectivity and model performance, we note that the ``gold label'' of a task instance is usually decided by a majority vote within a group of annotators. This means that the gold label may not represent the perspective of each individual~\cite{goyal2022your}, and they are sometimes ``biased'' themselves depending on the annotator decomposition~\cite{li2022towards}. Thus, it may be challenging for LLMs to generate synthetic data to recover such potentially biased ``majority view,'' especially if the LLMs are trained to maintain neutrality. Alternatively, one may ask for subjective task instances that humans can hardly reach any consensus on, whether the ``gold label'' is really the only ``correct'' label? If not, a rethinking of how to develop and evaluate models for these task instances is urgently needed.

\subsection{Explaining a few exceptions}

In Table~\ref{tab:results}, we surprisingly find that on the Tweet irony detection tasks, models trained on the few-shot synthetic data even outperform models trained on the real-world data. One plausible explanation is that the nature of generating irony texts for social media involves a creative writing task with few language formality constraints, and recent research suggests that LLMs have
the potential to exhibit comparable creativity with human writers in such task~\cite{franceschelli2023creativity}. Another exception we find is in Section~\ref{sec:eval2}---for the Financial Phrasebank and Scarcasm datasets, unlike other tasks, the effectiveness of the models trained on the synthetic data do not vary much with the instance-level task subjectivity. 
We conjecture that this can be caused by some 
task-specific properties.
On the Financial Phasebank dataset, accurate sentiment analysis 
requires the understanding of specialized terminology related to finance. Similarly, the Sarcasm detection task aims at identifying sarcasm in news headlines from selected sources and requires the comprehension on political topics. Thus, on these tasks, LLMs might not be fully equipped with the necessary domain knowledge to create effective synthetic data under the zero-shot setting. In fact, as shown in Figure~\ref{fig:instance_corr}, models trained on the zero-shot synthetic data have very low performance on these two datasets, regardless of the subjectivity levels of task instances. 

\vspace{-4pt}
\subsection{Limitations and future work}

We acknowledge that task subjectivity may not be the only factor that moderates the effectiveness of the LLM-generated synthetic data. Future studies can look into the potential moderating role of other factors, such as language formality and the requirement for domain-specific knowledge.  Our reliance on crowd workers in determining task subjectivity may introduce some variability due to their lack of linguistic expertise. Our evaluation is also based on the GPT-3.5-Turbo model only. It is important to note that the conclusions we get here may not generalize to other LLMs (e.g., the more advanced GPT-4), considering the continuous improvements of LLMs in generating human-like texts.

Our findings suggest that incorporating real-world data examples into the synthetic data generation process can increase the data diversity and boost the performance of the resulting models. Thus, future work can explore strategies that leverage human intelligence, such as feedback or direct intervention in the generation process, to further enrich the diversity of synthetic data~\cite{chung2023increasing} and to identify the most ``informative'' type of data instance to generate.  Finally, the significant correlation between the subjectivity of tasks or instances and the performance of models trained on synthetic data also suggests the potential to utilize the performance of such models as a proxy for approximating task or instance subjectivity, or to estimate the reliability of gold labels.

\bibliography{cite}

\begin{thebibliography}{56}
\expandafter\ifx\csname natexlab\endcsname\relax\def\natexlab#1{#1}\fi

\bibitem[{Aggarwal et~al.(2022)Aggarwal, Jin, and Ahmad}]{aggarwal2022entity}
Karan Aggarwal, Henry Jin, and Aitzaz Ahmad. 2022.
\newblock Entity-controlled synthetic text generation using contextual question
  and answering with pre-trained language models.

\bibitem[{all MiniLM-L6-v2(2023)}]{allminiLM}
all MiniLM-L6-v2. 2023.
\newblock sentence-transformers/all-minilm-l6-v2.
\newblock Accessed on Hugging Face Model Hub.
\newblock Available from:
  \url{https://huggingface.co/sentence-transformers/all-MiniLM-L6-v2}.

\bibitem[{Almeida et~al.(2011)Almeida, Hidalgo, and
  Yamakami}]{Almeida2011SpamFiltering}
Tiago~A. Almeida, Jose Maria~Gomez Hidalgo, and Akebo Yamakami. 2011.
\newblock Contributions to the study of sms spam filtering: New collection and
  results.
\newblock In \emph{Proceedings of the 2011 ACM Symposium on Document
  Engineering (DOCENG'11)}.

\bibitem[{Annamoradnejad and Zoghi(2020)}]{annamoradnejad2020colbert}
Issa Annamoradnejad and Gohar Zoghi. 2020.
\newblock Colbert: Using bert sentence embedding for humor detection.
\newblock \emph{arXiv preprint arXiv:2004.12765}.

\bibitem[{Benveniste(1971)}]{benveniste1971subjectivity}
Emile Benveniste. 1971.
\newblock Subjectivity in language.
\newblock \emph{Problems in general linguistics}, 1:223--30.

\bibitem[{Besnier et~al.(2020)Besnier, Jain, Bursuc, Cord, and
  P{\'e}rez}]{besnier2020dataset}
Victor Besnier, Himalaya Jain, Andrei Bursuc, Matthieu Cord, and Patrick
  P{\'e}rez. 2020.
\newblock This dataset does not exist: training models from generated images.
\newblock In \emph{ICASSP 2020-2020 IEEE International Conference on Acoustics,
  Speech and Signal Processing (ICASSP)}, pages 1--5. IEEE.

\bibitem[{Brown et~al.(2020)Brown, Mann, Ryder, Subbiah, Kaplan, Dhariwal,
  Neelakantan, Shyam, Sastry, Askell et~al.}]{brown2020language}
Tom Brown, Benjamin Mann, Nick Ryder, Melanie Subbiah, Jared~D Kaplan, Prafulla
  Dhariwal, Arvind Neelakantan, Pranav Shyam, Girish Sastry, Amanda Askell,
  et~al. 2020.
\newblock Language models are few-shot learners.
\newblock \emph{Advances in neural information processing systems},
  33:1877--1901.

\bibitem[{Chen et~al.(2022)Chen, Papangelis, Tao, Rosenbaum, Kim, Liu, Yu, and
  Hakkani-Tur}]{chen2022weakly}
Maximillian Chen, Alexandros Papangelis, Chenyang Tao, Andy Rosenbaum, Seokhwan
  Kim, Yang Liu, Zhou Yu, and Dilek Hakkani-Tur. 2022.
\newblock Weakly supervised data augmentation through prompting for dialogue
  understanding.
\newblock \emph{arXiv preprint arXiv:2210.14169}.

\bibitem[{Chung et~al.(2023)Chung, Kamar, and Amershi}]{chung2023increasing}
John Joon~Young Chung, Ece Kamar, and Saleema Amershi. 2023.
\newblock Increasing diversity while maintaining accuracy: Text data generation
  with large language models and human interventions.
\newblock \emph{arXiv preprint arXiv:2306.04140}.

\bibitem[{Clark et~al.(2021)Clark, August, Serrano, Haduong, Gururangan, and
  Smith}]{clark2021all}
Elizabeth Clark, Tal August, Sofia Serrano, Nikita Haduong, Suchin Gururangan,
  and Noah~A Smith. 2021.
\newblock All that's' human'is not gold: Evaluating human evaluation of
  generated text.
\newblock \emph{arXiv preprint arXiv:2107.00061}.

\bibitem[{Cormen et~al.(2022)Cormen, Leiserson, Rivest, and
  Stein}]{cormen2022introduction}
Thomas~H Cormen, Charles~E Leiserson, Ronald~L Rivest, and Clifford Stein.
  2022.
\newblock \emph{Introduction to algorithms}.
\newblock MIT press.

\bibitem[{Demszky et~al.(2020)Demszky, Movshovitz-Attias, Ko, Cowen, Nemade,
  and Ravi}]{demszky2020goemotions}
Dorottya Demszky, Dana Movshovitz-Attias, Jeongwoo Ko, Alan Cowen, Gaurav
  Nemade, and Sujith Ravi. 2020.
\newblock {GoEmotions: A Dataset of Fine-Grained Emotions}.
\newblock In \emph{58th Annual Meeting of the Association for Computational
  Linguistics (ACL)}.

\bibitem[{Devlin et~al.(2018)Devlin, Chang, Lee, and
  Toutanova}]{devlin2018bert}
Jacob Devlin, Ming-Wei Chang, Kenton Lee, and Kristina Toutanova. 2018.
\newblock Bert: Pre-training of deep bidirectional transformers for language
  understanding.
\newblock \emph{arXiv preprint arXiv:1810.04805}.

\bibitem[{Dou et~al.(2021)Dou, Forbes, Koncel-Kedziorski, Smith, and
  Choi}]{dou2021gpt}
Yao Dou, Maxwell Forbes, Rik Koncel-Kedziorski, Noah~A Smith, and Yejin Choi.
  2021.
\newblock Is gpt-3 text indistinguishable from human text? scarecrow: A
  framework for scrutinizing machine text.
\newblock \emph{arXiv preprint arXiv:2107.01294}.

\bibitem[{Ekman et~al.(1999)}]{ekman1999basic}
Paul Ekman et~al. 1999.
\newblock Basic emotions.
\newblock \emph{Handbook of cognition and emotion}, 98(45-60):16.

\bibitem[{Franceschelli and Musolesi(2023)}]{franceschelli2023creativity}
Giorgio Franceschelli and Mirco Musolesi. 2023.
\newblock On the creativity of large language models.
\newblock \emph{arXiv preprint arXiv:2304.00008}.

\bibitem[{Gao et~al.(2022)Gao, Pi, Yong, Xu, Ye, Wu, ZHANG, Liang, Li, and
  Kong}]{gao2022self}
Jiahui Gao, Renjie Pi, LIN Yong, Hang Xu, Jiacheng Ye, Zhiyong Wu, WEIZHONG
  ZHANG, Xiaodan Liang, Zhenguo Li, and Lingpeng Kong. 2022.
\newblock Self-guided noise-free data generation for efficient zero-shot
  learning.
\newblock In \emph{The Eleventh International Conference on Learning
  Representations}.

\bibitem[{Gao et~al.(2023)Gao, Pi, Yong, Xu, Ye, Wu, ZHANG, Liang, Li, and
  Kong}]{gao2023self}
Jiahui Gao, Renjie Pi, LIN Yong, Hang Xu, Jiacheng Ye, Zhiyong Wu, WEIZHONG
  ZHANG, Xiaodan Liang, Zhenguo Li, and Lingpeng Kong. 2023.
\newblock Self-guided noise-free data generation for efficient zero-shot
  learning.
\newblock In \emph{The Eleventh International Conference on Learning
  Representations}.

\bibitem[{Gao et~al.(2019)Gao, Han, Zhu, Liu, Li, Sun, and
  Zhou}]{gao-etal-2019-fewrel}
Tianyu Gao, Xu~Han, Hao Zhu, Zhiyuan Liu, Peng Li, Maosong Sun, and Jie Zhou.
  2019.
\newblock \href {https://doi.org/10.18653/v1/D19-1649} {{F}ew{R}el 2.0: Towards
  more challenging few-shot relation classification}.
\newblock In \emph{Proceedings of the 2019 Conference on Empirical Methods in
  Natural Language Processing and the 9th International Joint Conference on
  Natural Language Processing (EMNLP-IJCNLP)}, pages 6251--6256, Hong Kong,
  China. Association for Computational Linguistics.

\bibitem[{Gordon et~al.(2022)Gordon, Lam, Park, Patel, Hancock, Hashimoto, and
  Bernstein}]{gordon2022jury}
Mitchell~L Gordon, Michelle~S Lam, Joon~Sung Park, Kayur Patel, Jeff Hancock,
  Tatsunori Hashimoto, and Michael~S Bernstein. 2022.
\newblock Jury learning: Integrating dissenting voices into machine learning
  models.
\newblock In \emph{Proceedings of the 2022 CHI Conference on Human Factors in
  Computing Systems}, pages 1--19.

\bibitem[{Goyal et~al.(2022)Goyal, Kivlichan, Rosen, and
  Vasserman}]{goyal2022your}
Nitesh Goyal, Ian~D Kivlichan, Rachel Rosen, and Lucy Vasserman. 2022.
\newblock Is your toxicity my toxicity? exploring the impact of rater identity
  on toxicity annotation.
\newblock \emph{Proceedings of the ACM on Human-Computer Interaction},
  6(CSCW2):1--28.

\bibitem[{H\"{a}m\"{a}l\"{a}inen et~al.(2023)H\"{a}m\"{a}l\"{a}inen, Tavast,
  and Kunnari}]{10.1145/3544548.3580688}
Perttu H\"{a}m\"{a}l\"{a}inen, Mikke Tavast, and Anton Kunnari. 2023.
\newblock \href {https://doi.org/10.1145/3544548.3580688} {Evaluating large
  language models in generating synthetic hci research data: A case study}.
\newblock In \emph{Proceedings of the 2023 CHI Conference on Human Factors in
  Computing Systems}, CHI '23, New York, NY, USA. Association for Computing
  Machinery.

\bibitem[{Hartvigsen et~al.(2022)Hartvigsen, Gabriel, Palangi, Sap, Ray, and
  Kamar}]{hartvigsen2022toxigen}
Thomas Hartvigsen, Saadia Gabriel, Hamid Palangi, Maarten Sap, Dipankar Ray,
  and Ece Kamar. 2022.
\newblock Toxigen: A large-scale machine-generated dataset for adversarial and
  implicit hate speech detection.
\newblock \emph{arXiv preprint arXiv:2203.09509}.

\bibitem[{He et~al.(2022)He, Sun, Yu, Xue, Zhang, Torr, Bai, and
  Qi}]{he2022synthetic}
Ruifei He, Shuyang Sun, Xin Yu, Chuhui Xue, Wenqing Zhang, Philip Torr, Song
  Bai, and Xiaojuan Qi. 2022.
\newblock Is synthetic data from generative models ready for image recognition?
\newblock \emph{arXiv preprint arXiv:2210.07574}.

\bibitem[{Jindal and Liu(2007)}]{jindal2007review}
Nitin Jindal and Bing Liu. 2007.
\newblock Review spam detection.
\newblock In \emph{Proceedings of the 16th international conference on World
  Wide Web}, pages 1189--1190.

\bibitem[{Karras et~al.(2019)Karras, Laine, and Aila}]{karras2019style}
Tero Karras, Samuli Laine, and Timo Aila. 2019.
\newblock A style-based generator architecture for generative adversarial
  networks.
\newblock In \emph{Proceedings of the IEEE/CVF conference on computer vision
  and pattern recognition}, pages 4401--4410.

\bibitem[{Kumar et~al.(2020)Kumar, Choudhary, and Cho}]{kumar2020data}
Varun Kumar, Ashutosh Choudhary, and Eunah Cho. 2020.
\newblock Data augmentation using pre-trained transformer models.
\newblock \emph{arXiv preprint arXiv:2003.02245}.

\bibitem[{Li et~al.(2022)Li, Lu, and Yin}]{li2022towards}
Zhuoyan Li, Zhuoran Lu, and Ming Yin. 2022.
\newblock Towards better detection of biased language with scarce, noisy, and
  biased annotations.
\newblock In \emph{Proceedings of the 2022 AAAI/ACM Conference on AI, Ethics,
  and Society}, pages 411--423.

\bibitem[{Liu et~al.(2019)Liu, Ott, Goyal, Du, Joshi, Chen, Levy, Lewis,
  Zettlemoyer, and Stoyanov}]{liu2019roberta}
Yinhan Liu, Myle Ott, Naman Goyal, Jingfei Du, Mandar Joshi, Danqi Chen, Omer
  Levy, Mike Lewis, Luke Zettlemoyer, and Veselin Stoyanov. 2019.
\newblock Roberta: A robustly optimized bert pretraining approach.
\newblock \emph{arXiv preprint arXiv:1907.11692}.

\bibitem[{Maas et~al.(2011)Maas, Daly, Pham, Huang, Ng, and
  Potts}]{maas-EtAl:2011:ACL-HLT2011}
Andrew~L. Maas, Raymond~E. Daly, Peter~T. Pham, Dan Huang, Andrew~Y. Ng, and
  Christopher Potts. 2011.
\newblock \href {http://www.aclweb.org/anthology/P11-1015} {Learning word
  vectors for sentiment analysis}.
\newblock In \emph{Proceedings of the 49th Annual Meeting of the Association
  for Computational Linguistics: Human Language Technologies}, pages 142--150,
  Portland, Oregon, USA. Association for Computational Linguistics.

\bibitem[{Malo et~al.(2014)Malo, Sinha, Korhonen, Wallenius, and
  Takala}]{Malo2014GoodDO}
P.~Malo, A.~Sinha, P.~Korhonen, J.~Wallenius, and P.~Takala. 2014.
\newblock Good debt or bad debt: Detecting semantic orientations in economic
  texts.
\newblock \emph{Journal of the Association for Information Science and
  Technology}, 65.

\bibitem[{Mcnutt et~al.(2023)Mcnutt, Wang, Deline, and
  Drucker}]{10.1145/3544548.3580940}
Andrew~M Mcnutt, Chenglong Wang, Robert~A Deline, and Steven~M. Drucker. 2023.
\newblock \href {https://doi.org/10.1145/3544548.3580940} {On the design of
  ai-powered code assistants for notebooks}.
\newblock In \emph{Proceedings of the 2023 CHI Conference on Human Factors in
  Computing Systems}, CHI '23, New York, NY, USA. Association for Computing
  Machinery.

\bibitem[{Meng et~al.(2022)Meng, Huang, Zhang, and Han}]{meng2022generating}
Yu~Meng, Jiaxin Huang, Yu~Zhang, and Jiawei Han. 2022.
\newblock Generating training data with language models: Towards zero-shot
  language understanding.
\newblock \emph{Advances in Neural Information Processing Systems},
  35:462--477.

\bibitem[{Misra and Arora(2023)}]{misra2023Sarcasm}
Rishabh Misra and Prahal Arora. 2023.
\newblock \href {https://doi.org/https://doi.org/10.1016/j.aiopen.2023.01.001}
  {Sarcasm detection using news headlines dataset}.
\newblock \emph{AI Open}, 4:13--18.

\bibitem[{Misra and Grover(2021)}]{misra2021sculpting}
Rishabh Misra and Jigyasa Grover. 2021.
\newblock \emph{Sculpting Data for ML: The first act of Machine Learning}.

\bibitem[{Mohammad et~al.(2018)Mohammad, Bravo-Marquez, Salameh, and
  Kiritchenko}]{mohammad2018semeval}
Saif Mohammad, Felipe Bravo-Marquez, Mohammad Salameh, and Svetlana
  Kiritchenko. 2018.
\newblock Semeval-2018 task 1: Affect in tweets.
\newblock In \emph{Proceedings of the 12th international workshop on semantic
  evaluation}, pages 1--17.

\bibitem[{Nichol et~al.(2021)Nichol, Dhariwal, Ramesh, Shyam, Mishkin, McGrew,
  Sutskever, and Chen}]{nichol2021glide}
Alex Nichol, Prafulla Dhariwal, Aditya Ramesh, Pranav Shyam, Pamela Mishkin,
  Bob McGrew, Ilya Sutskever, and Mark Chen. 2021.
\newblock Glide: Towards photorealistic image generation and editing with
  text-guided diffusion models.
\newblock \emph{arXiv preprint arXiv:2112.10741}.

\bibitem[{OpenAI(2023)}]{openai2023gpt4}
OpenAI. 2023.
\newblock \href {http://arxiv.org/abs/2303.08774} {Gpt-4 technical report}.

\bibitem[{Radford et~al.(2019)Radford, Wu, Child, Luan, Amodei, Sutskever
  et~al.}]{radford2019language}
Alec Radford, Jeffrey Wu, Rewon Child, David Luan, Dario Amodei, Ilya
  Sutskever, et~al. 2019.
\newblock Language models are unsupervised multitask learners.
\newblock \emph{OpenAI blog}, 1(8):9.

\bibitem[{Rhys~Cox et~al.(2021)Rhys~Cox, Wang, Abdul, von~der Weth, and
  Y.~Lim}]{10.1145/3411764.3445782}
Samuel Rhys~Cox, Yunlong Wang, Ashraf Abdul, Christian von~der Weth, and Brian
  Y.~Lim. 2021.
\newblock \href {https://doi.org/10.1145/3411764.3445782} {Directed diversity:
  Leveraging language embedding distances for collective creativity in crowd
  ideation}.
\newblock In \emph{Proceedings of the 2021 CHI Conference on Human Factors in
  Computing Systems}, CHI '21, New York, NY, USA. Association for Computing
  Machinery.

\bibitem[{Sahu et~al.(2022)Sahu, Rodriguez, Laradji, Atighehchian, Vazquez, and
  Bahdanau}]{sahu2022data}
Gaurav Sahu, Pau Rodriguez, Issam~H. Laradji, Parmida Atighehchian, David
  Vazquez, and Dzmitry Bahdanau. 2022.
\newblock \href {http://arxiv.org/abs/2204.01959} {Data augmentation for intent
  classification with off-the-shelf large language models}.

\bibitem[{Sap et~al.(2021)Sap, Swayamdipta, Vianna, Zhou, Choi, and
  Smith}]{sap2021annotators}
Maarten Sap, Swabha Swayamdipta, Laura Vianna, Xuhui Zhou, Yejin Choi, and
  Noah~A Smith. 2021.
\newblock Annotators with attitudes: How annotator beliefs and identities bias
  toxic language detection.
\newblock \emph{arXiv preprint arXiv:2111.07997}.

\bibitem[{Tang et~al.(2023)Tang, Han, Jiang, and Hu}]{tang2023does}
Ruixiang Tang, Xiaotian Han, Xiaoqian Jiang, and Xia Hu. 2023.
\newblock Does synthetic data generation of llms help clinical text mining?
\newblock \emph{arXiv preprint arXiv:2303.04360}.

\bibitem[{Van~Hee et~al.(2018)Van~Hee, Lefever, and Hoste}]{van2018semeval}
Cynthia Van~Hee, Els Lefever, and V{\'e}ronique Hoste. 2018.
\newblock Semeval-2018 task 3: Irony detection in english tweets.
\newblock In \emph{Proceedings of The 12th International Workshop on Semantic
  Evaluation}, pages 39--50.

\bibitem[{Vaswani et~al.(2017)Vaswani, Shazeer, Parmar, Uszkoreit, Jones,
  Gomez, Kaiser, and Polosukhin}]{vaswani2017attention}
Ashish Vaswani, Noam Shazeer, Niki Parmar, Jakob Uszkoreit, Llion Jones,
  Aidan~N Gomez, {\L}ukasz Kaiser, and Illia Polosukhin. 2017.
\newblock Attention is all you need.
\newblock \emph{Advances in neural information processing systems}, 30.

\bibitem[{Veatch(1998)}]{veatch1998theory}
Thomas~C Veatch. 1998.
\newblock A theory of humor.

\bibitem[{Wang et~al.(2021)Wang, Yu, Firat, and Cao}]{wang2021towards}
Zirui Wang, Adams~Wei Yu, Orhan Firat, and Yuan Cao. 2021.
\newblock Towards zero-label language learning.
\newblock \emph{arXiv preprint arXiv:2109.09193}.

\bibitem[{Wei et~al.(2021)Wei, Bosma, Zhao, Guu, Yu, Lester, Du, Dai, and
  Le}]{wei2021finetuned}
Jason Wei, Maarten Bosma, Vincent~Y Zhao, Kelvin Guu, Adams~Wei Yu, Brian
  Lester, Nan Du, Andrew~M Dai, and Quoc~V Le. 2021.
\newblock Finetuned language models are zero-shot learners.
\newblock \emph{arXiv preprint arXiv:2109.01652}.

\bibitem[{Wiebe et~al.(2004)Wiebe, Wilson, Bruce, Bell, and
  Martin}]{wiebe2004learning}
Janyce Wiebe, Theresa Wilson, Rebecca Bruce, Matthew Bell, and Melanie Martin.
  2004.
\newblock Learning subjective language.
\newblock \emph{Computational linguistics}, 30(3):277--308.

\bibitem[{Wiegand et~al.(2019)Wiegand, Ruppenhofer, and
  Kleinbauer}]{wiegand2019detection}
Michael Wiegand, Josef Ruppenhofer, and Thomas Kleinbauer. 2019.
\newblock Detection of abusive language: the problem of biased datasets.
\newblock In \emph{Proceedings of the 2019 conference of the North American
  Chapter of the Association for Computational Linguistics: human language
  technologies, volume 1 (long and short papers)}, pages 602--608.

\bibitem[{Wolf et~al.(2020)Wolf, Debut, Sanh, Chaumond, Delangue, Moi, Cistac,
  Rault, Louf, Funtowicz, Davison, Shleifer, von Platen, Ma, Jernite, Plu, Xu,
  Le~Scao, Gugger, Drame, Lhoest, and Rush}]{wolf-etal-2020-transformers}
Thomas Wolf, Lysandre Debut, Victor Sanh, Julien Chaumond, Clement Delangue,
  Anthony Moi, Pierric Cistac, Tim Rault, Remi Louf, Morgan Funtowicz, Joe
  Davison, Sam Shleifer, Patrick von Platen, Clara Ma, Yacine Jernite, Julien
  Plu, Canwen Xu, Teven Le~Scao, Sylvain Gugger, Mariama Drame, Quentin Lhoest,
  and Alexander Rush. 2020.
\newblock \href {https://doi.org/10.18653/v1/2020.emnlp-demos.6} {Transformers:
  State-of-the-art natural language processing}.
\newblock In \emph{Proceedings of the 2020 Conference on Empirical Methods in
  Natural Language Processing: System Demonstrations}, pages 38--45, Online.
  Association for Computational Linguistics.

\bibitem[{Ye et~al.(2022)Ye, Gao, Li, Xu, Feng, Wu, Yu, and
  Kong}]{ye2022zerogen}
Jiacheng Ye, Jiahui Gao, Qintong Li, Hang Xu, Jiangtao Feng, Zhiyong Wu, Tao
  Yu, and Lingpeng Kong. 2022.
\newblock Zerogen: Efficient zero-shot learning via dataset generation.
\newblock \emph{arXiv preprint arXiv:2202.07922}.

\bibitem[{Yoo et~al.(2021)Yoo, Park, Kang, Lee, and Park}]{yoo2021gpt3mix}
Kang~Min Yoo, Dongju Park, Jaewook Kang, Sang-Woo Lee, and Woomyeong Park.
  2021.
\newblock Gpt3mix: Leveraging large-scale language models for text
  augmentation.
\newblock \emph{arXiv preprint arXiv:2104.08826}.

\bibitem[{Zhang et~al.(2015)Zhang, Zhao, and LeCun}]{Zhang2015CharacterlevelCN}
Xiang Zhang, Junbo~Jake Zhao, and Yann LeCun. 2015.
\newblock Character-level convolutional networks for text classification.
\newblock In \emph{NIPS}.

\bibitem[{Zhang et~al.(2021)Zhang, Ling, Gao, Yin, Lafleche, Barriuso,
  Torralba, and Fidler}]{zhang2021datasetgan}
Yuxuan Zhang, Huan Ling, Jun Gao, Kangxue Yin, Jean-Francois Lafleche, Adela
  Barriuso, Antonio Torralba, and Sanja Fidler. 2021.
\newblock \href {http://arxiv.org/abs/2104.06490} {Datasetgan: Efficient
  labeled data factory with minimal human effort}.

\bibitem[{Zhou et~al.(2023)Zhou, Zhang, Luo, Parker, and
  De~Choudhury}]{10.1145/3544548.3581318}
Jiawei Zhou, Yixuan Zhang, Qianni Luo, Andrea~G Parker, and Munmun
  De~Choudhury. 2023.
\newblock \href {https://doi.org/10.1145/3544548.3581318} {Synthetic lies:
  Understanding ai-generated misinformation and evaluating algorithmic and
  human solutions}.
\newblock CHI '23, New York, NY, USA. Association for Computing Machinery.

\end{thebibliography}
\bibliographystyle{acl_natbib}

\clearpage

\appendix
\section{Appendices}
\label{sec:appendix}
\counterwithin{figure}{section}
\counterwithin{table}{section}

\subsection{Dataset and Task Descriptions}
\label{dataset}
\textbf{AG's News:} This task involves classifying news articles from the subset of AG's News Topic Classification dataset into one of thee categories: World, Sports and Sci/Tech. The AG's News Topic Classification dataset, collected from over 2,000 news sources by the academic news search engine, ComeToMyHead, consists of a training set of 120,000 instances and a test set of 7,600 instances.\\
\textbf{Relation Classification:} This task requires the identification of the relationships between two entities within a given sentence. In this study, we focus on four relations: `country', `league', `screenwriter', and `tributary'. The dataset comprises English text sourced from Wikipedia and supplemented with crowdsourced English annotations. Each relation has 700 instances. As the dataset does not provide an official division into train, validation, and test sets, we randomly allocated the dataset into train (70\%), validation (5\%), and test (25\%) sets. In our evaluation, this process was repeated three times, with the average performance reported. \\
\textbf{IMDB Reviews:} This task requires classifying the sentiment of movie reviews from the IMDB platform into one of two categories: positive (pos) or negative (neg). The dataset comprises 50,000 movie reviews evenly split, with 25,000 designated for training and 25,000 for testing. \\
\textbf{SMS Message Spam:} This task involves the classification of SMS messages from the SMS Spam Collection v.1 dataset into either `ham' (legitimate) or `spam' categories. The training dataset contains 5,574 English messages, each labeled according to its legitimacy. As the dataset does not provide an official division into train, validation, and test sets, we randomly divided the dataset into train (70\%), validation (5\%), and test (25\%) sets. In our evaluation, this process was repeated three times, with the average performance reported.\\
\textbf{Financial Phrasebank:} This task entails the classification of finance-related sentences into one of three categories---positive, negative, or neutral---based on the sentiment expressed by the sentence. The dataset comprises 4,840 English sentences sourced from financial news articles.  As the dataset does not provide an official division into train, validation, and test sets, we randomly allocated the dataset into train (70\%), validation (5\%), and test (25\%) sets. In our evaluation, this process was repeated three times, with the average performance reported.  \\
\textbf{Reddit Emotion:} The Reddit Emotion is the subset of the Go Emotions dataset. The  
Go Emotions dataset is comprised of 58,009 comments collected from Reddit, and each comment has been annotated with respect to 28 
emotion categories. In this task, we focus on three basic emotions~\cite{ekman1999basic}: joy, 
sadness, and surprise. \\
\textbf{Tweet Irony Speech:} The task involves classifying tweets into two categories: 
irony, non-irony. The dataset, which is composed of English-language tweets, has been manually annotated for these specific categories. The distribution of the data includes a training set of 2,862 instances and a test set of 784 instances. \\
\textbf{Tweet Emotion:} The task involves classifying tweets into four emotion 
categories: anger, joy, optimism, sadness.  Each tweet in this English-language dataset has been annotated by human reviewers with respect to these emotional categories. The dataset is partitioned into a training set of 3,257 instances and a test set of 1,421 instances. \\
\textbf{Sarcasm News Headlines:} This task requires distinguishing between sarcastic and non-sarcastic news headlines. The dataset comprises 26,709 headlines from two news sources: TheOnion, representing sarcasm, and HuffPost, representing non-sarcasm. As the dataset does not provide an official division into train, validation, and test sets, we randomly allocated the dataset into train (70\%), validation (5\%), and test (25\%) sets. In our evaluation, this process was repeated three times, with the average performance reported. \\
\textbf{Humor Speech Detection:} This task involves discerning humorous from non-humorous content for short texts. The dataset, specifically curated for humor detection, is composed of 200,000 instances, balanced between humorous and non-humorous data. It is divided into a training set of 160,000 instances and a test set of 40,000 instances.

\section{Evaluation \RNum{1}: Comparison Across Different Types of Tasks (Additional Results)}
\label{e1a}

\begin{figure*}[t]
  \centering
  \subfloat[AG's News]{\includegraphics[width=0.2\textwidth]{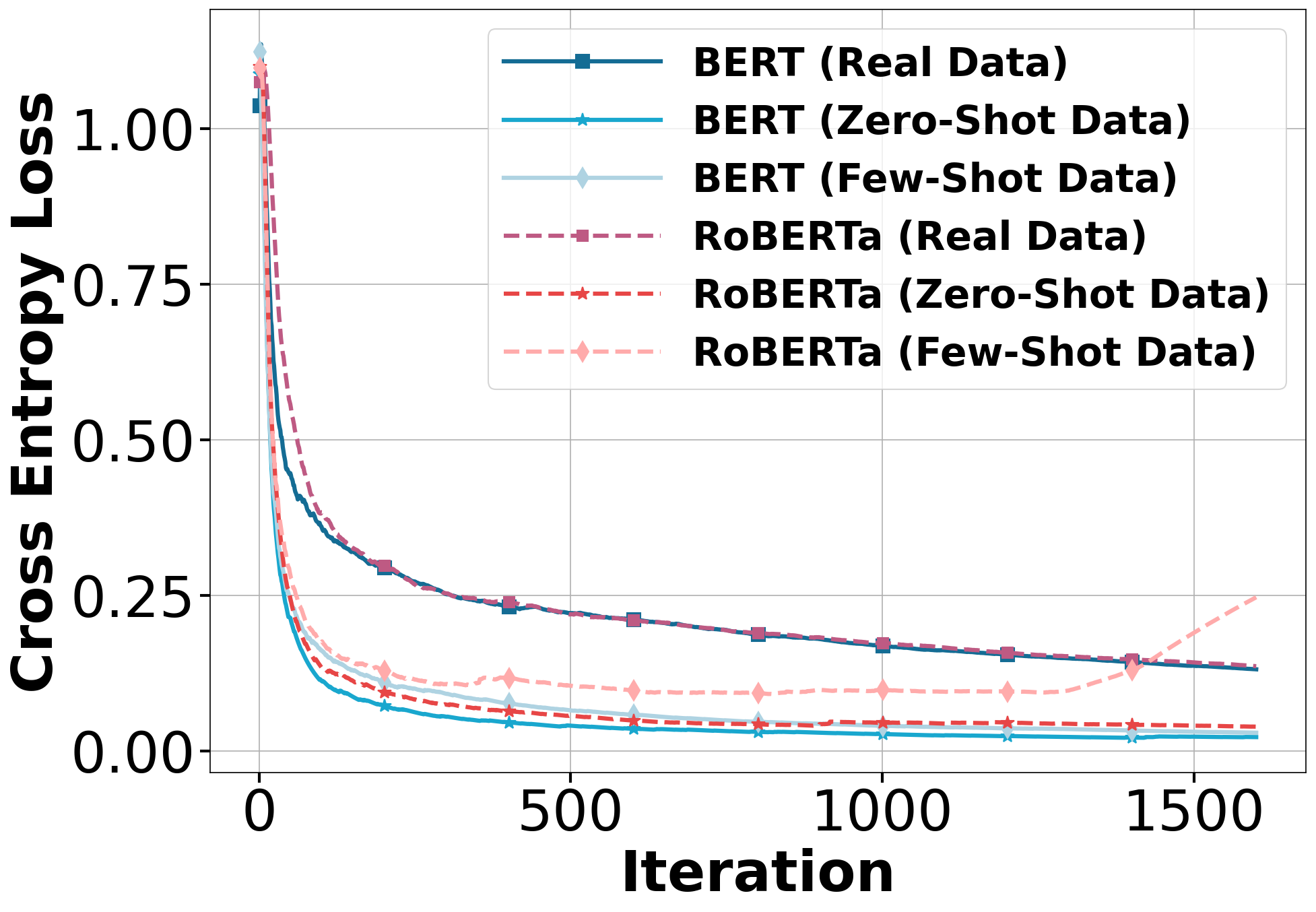}\label{fig:agnews}}
  \hfill
  \subfloat[Relation]{\includegraphics[width=0.2\textwidth]{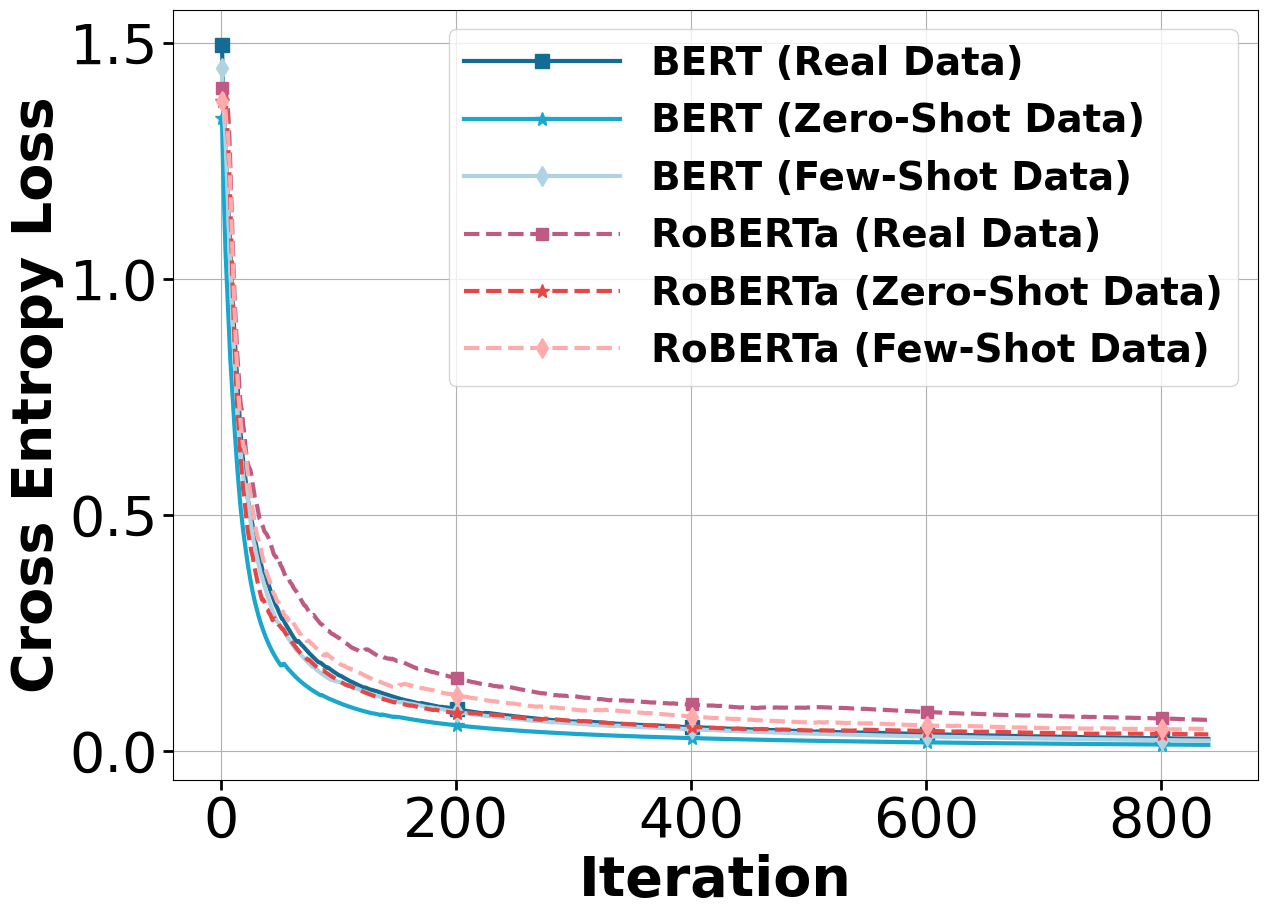}\label{fig:fewrel}}
  \hfill
  \subfloat[IMDB Reviews]{\includegraphics[width=0.2\textwidth]{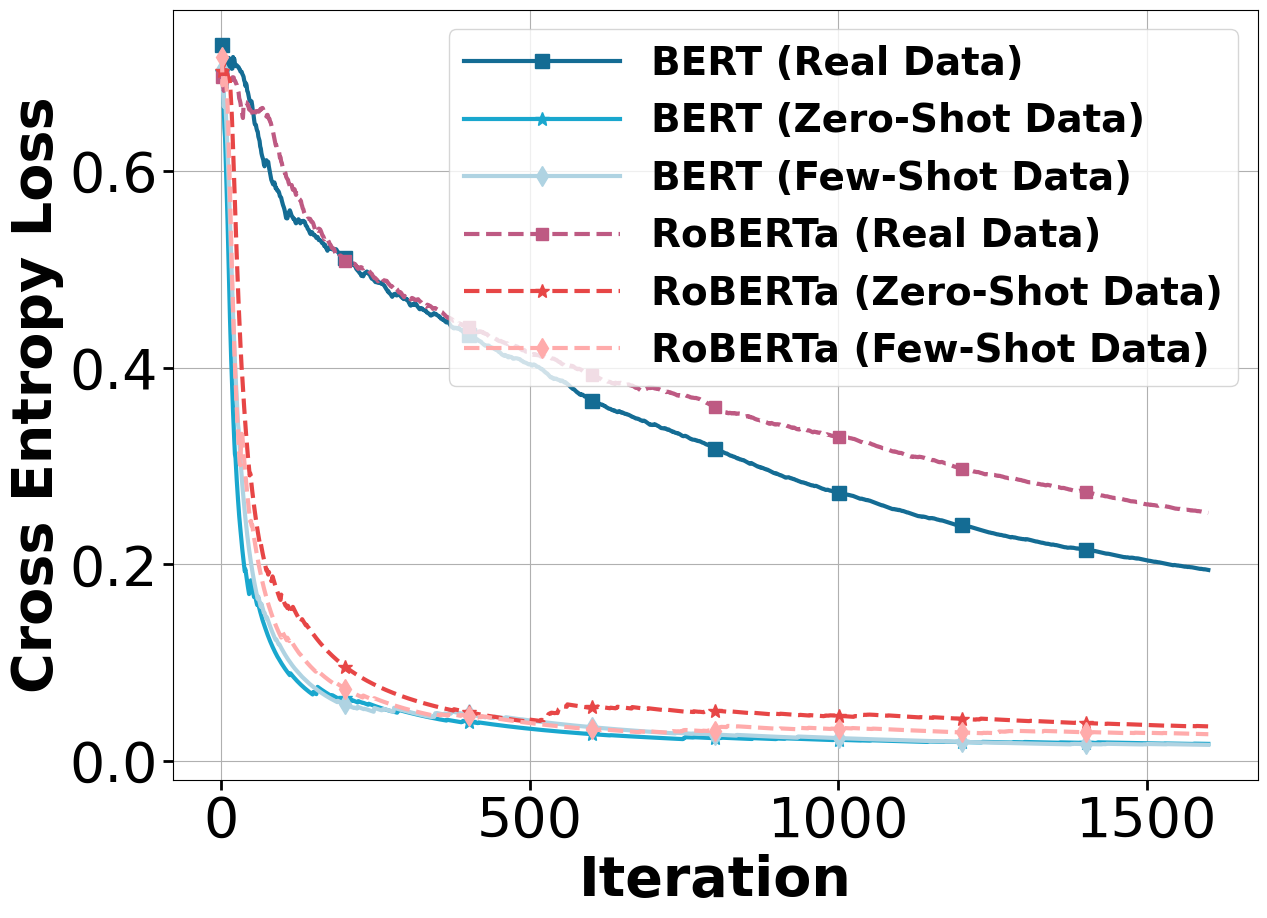}\label{fig:imdb}}
  \hfill
  \subfloat[SMS Spam]{\includegraphics[width=0.2\textwidth]{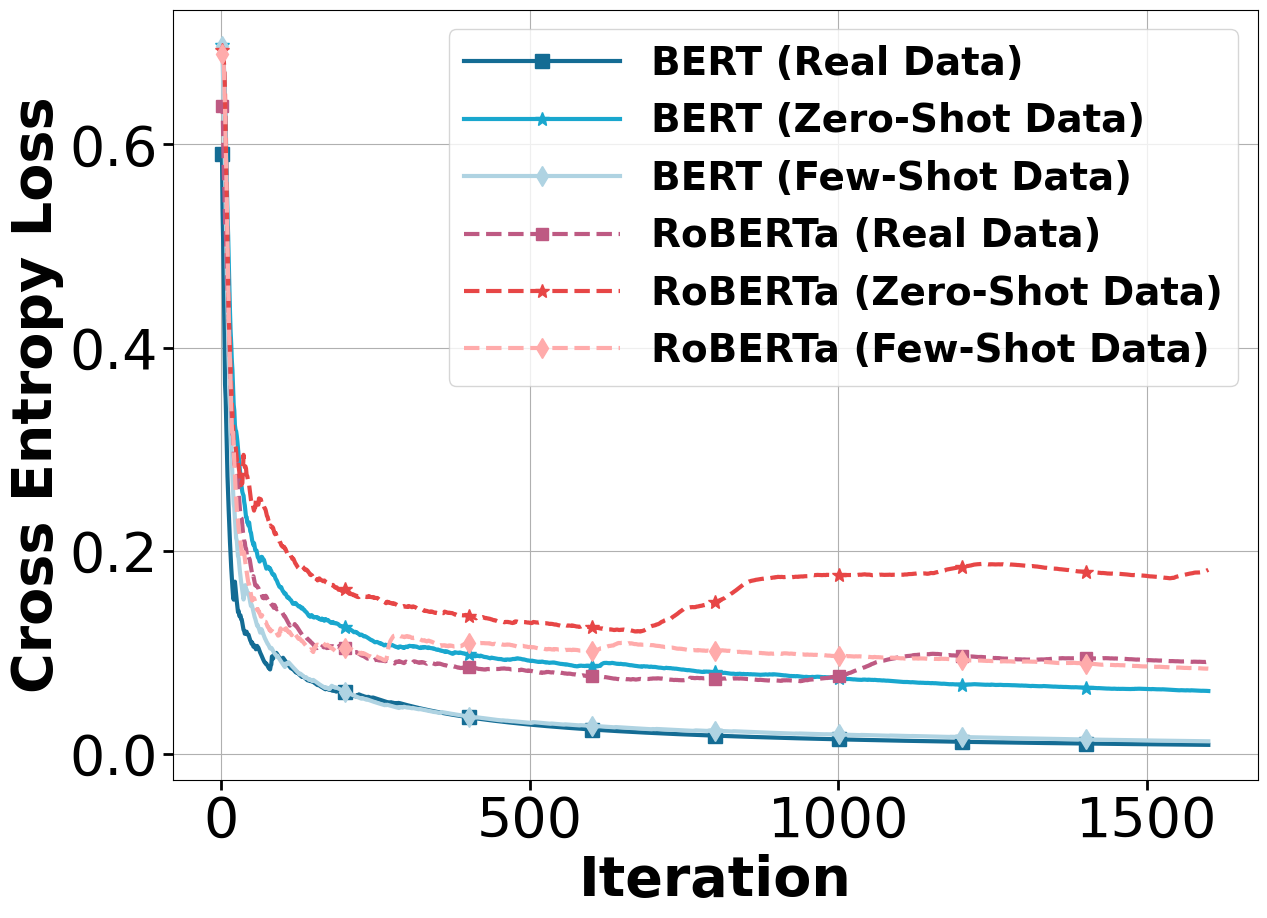}\label{fig:spam}}
  \hfill
  \subfloat[Financial Phrasebank]{\includegraphics[width=0.2\textwidth]{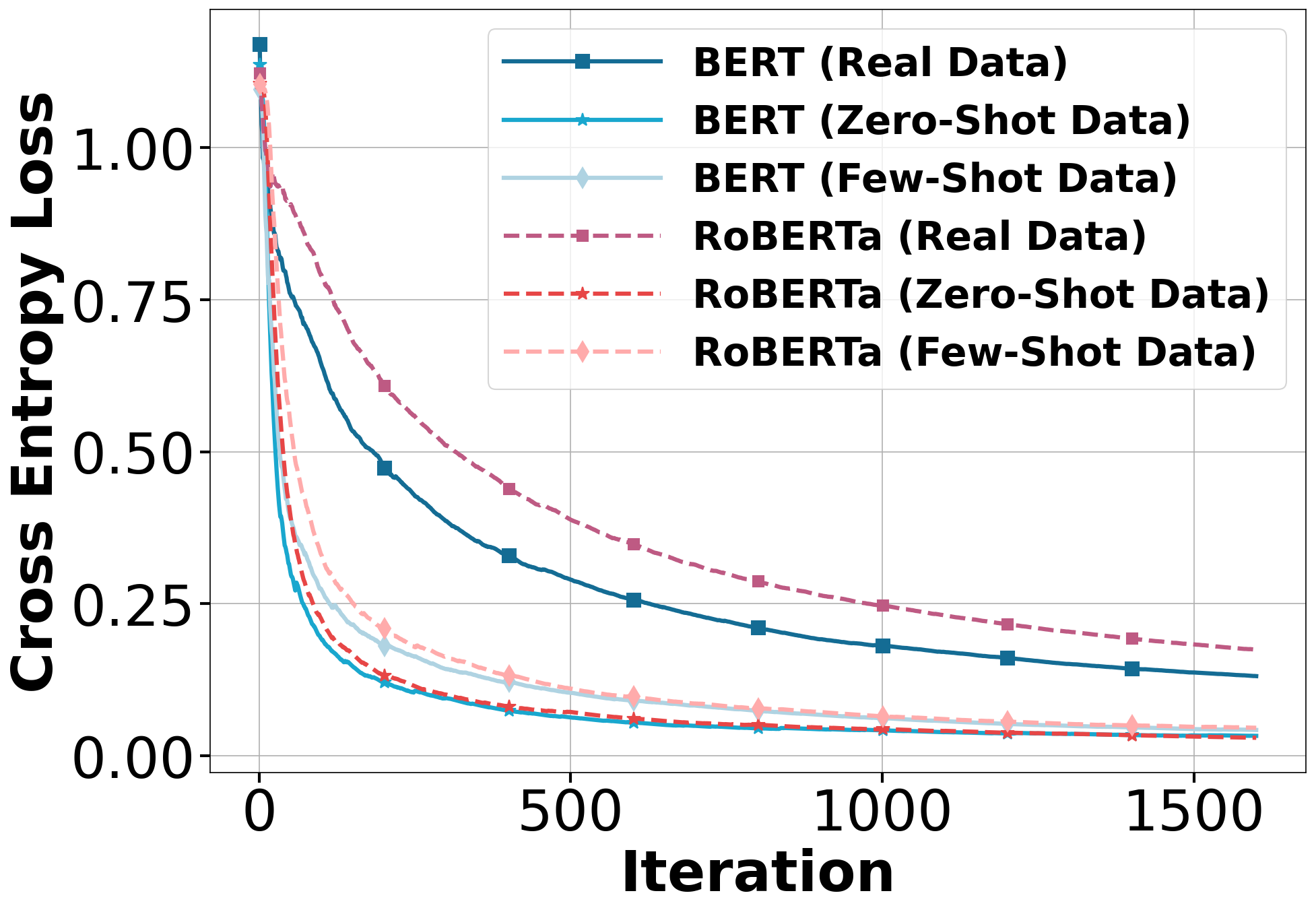}\label{fig:fin}}
  \\
  \subfloat[Reddit Emotion]{\includegraphics[width=0.2\textwidth]{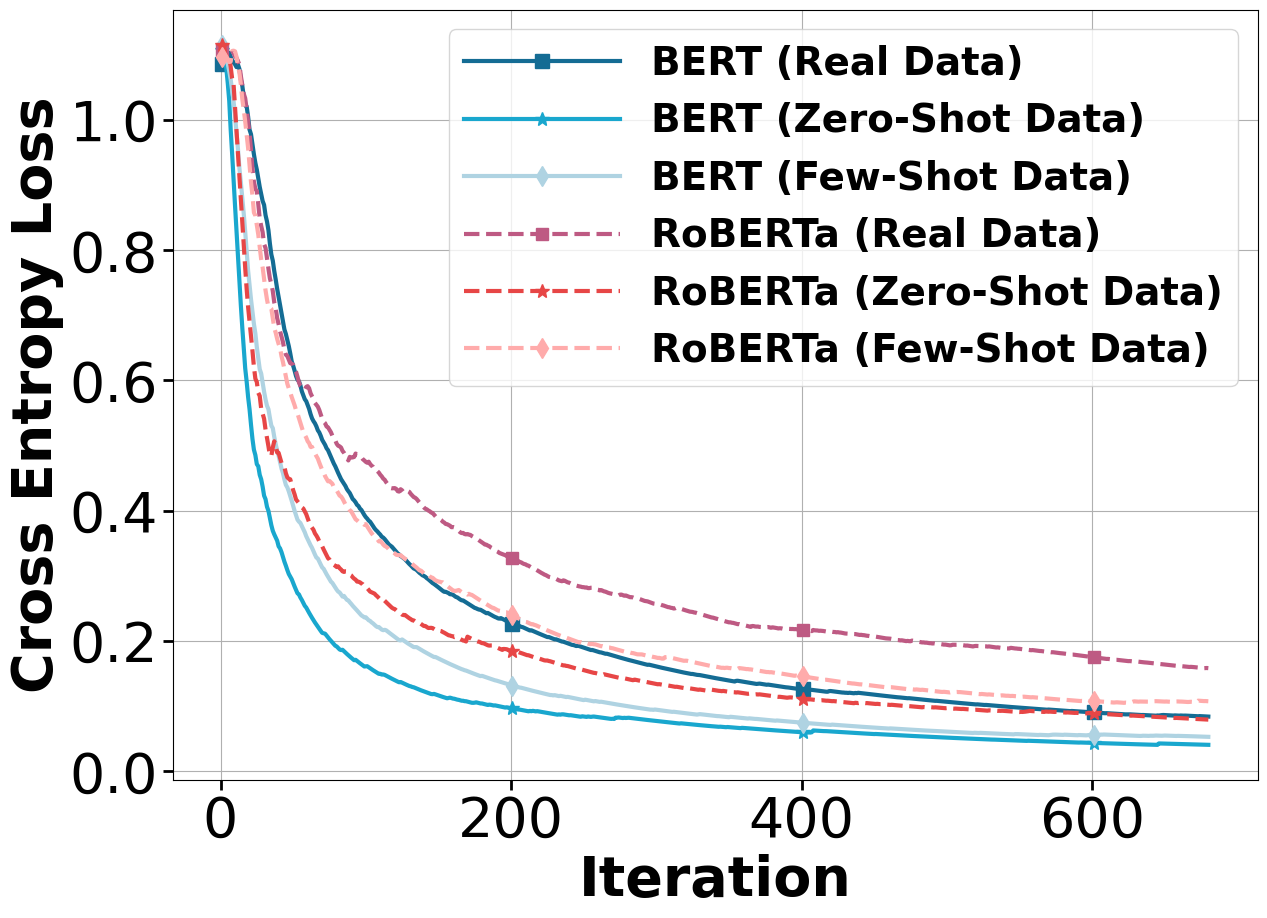}\label{fig:goemo}}
  \hfill
  \subfloat[Sarcasm News]{\includegraphics[width=0.2\textwidth]{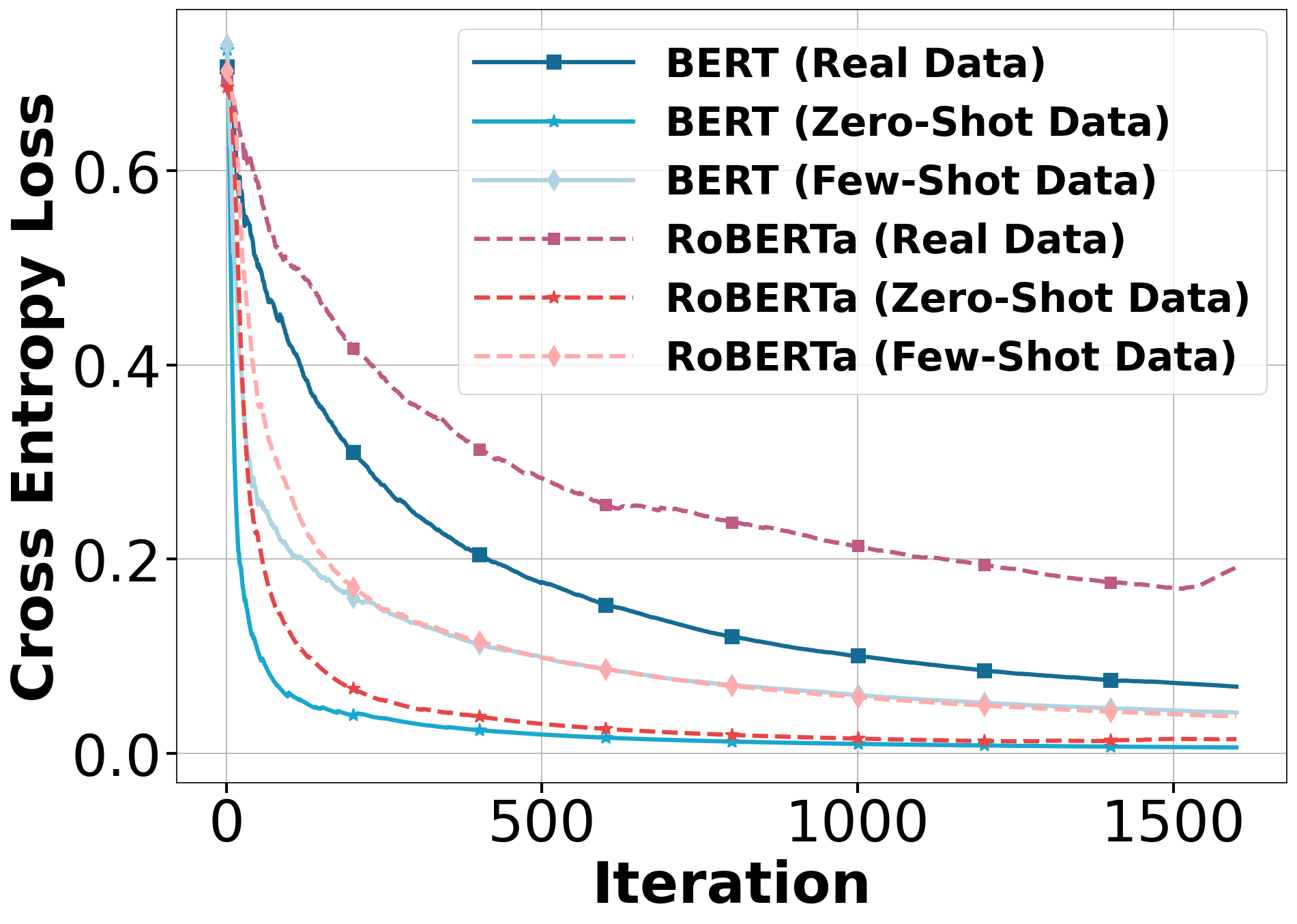}\label{fig:sarcasm}}
  \hfill
  \subfloat[Humor Detection]{\includegraphics[width=0.2\textwidth]{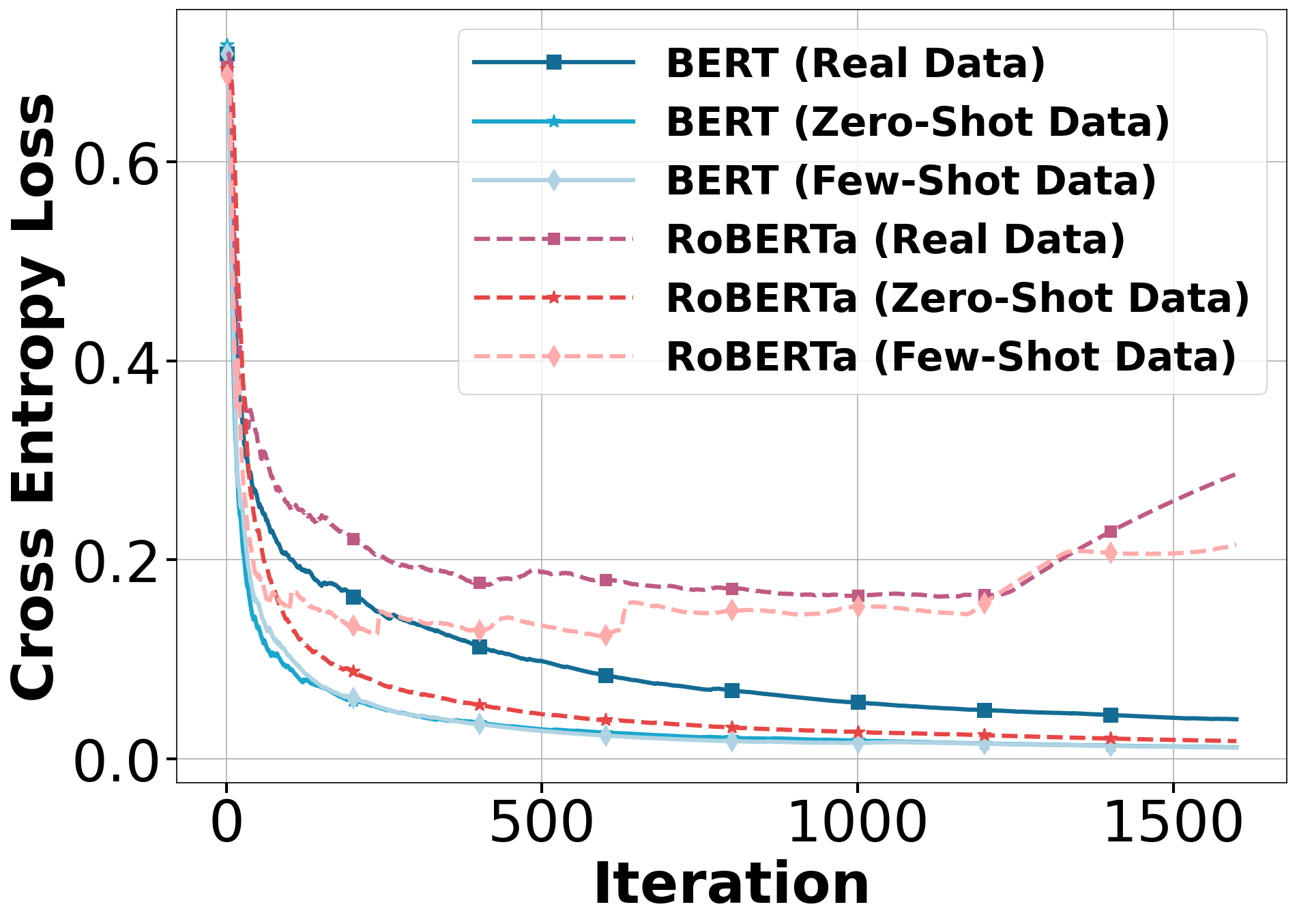}\label{fig:humor}}
  \hfill
  \subfloat[Tweet Emotions]{\includegraphics[width=0.2\textwidth]{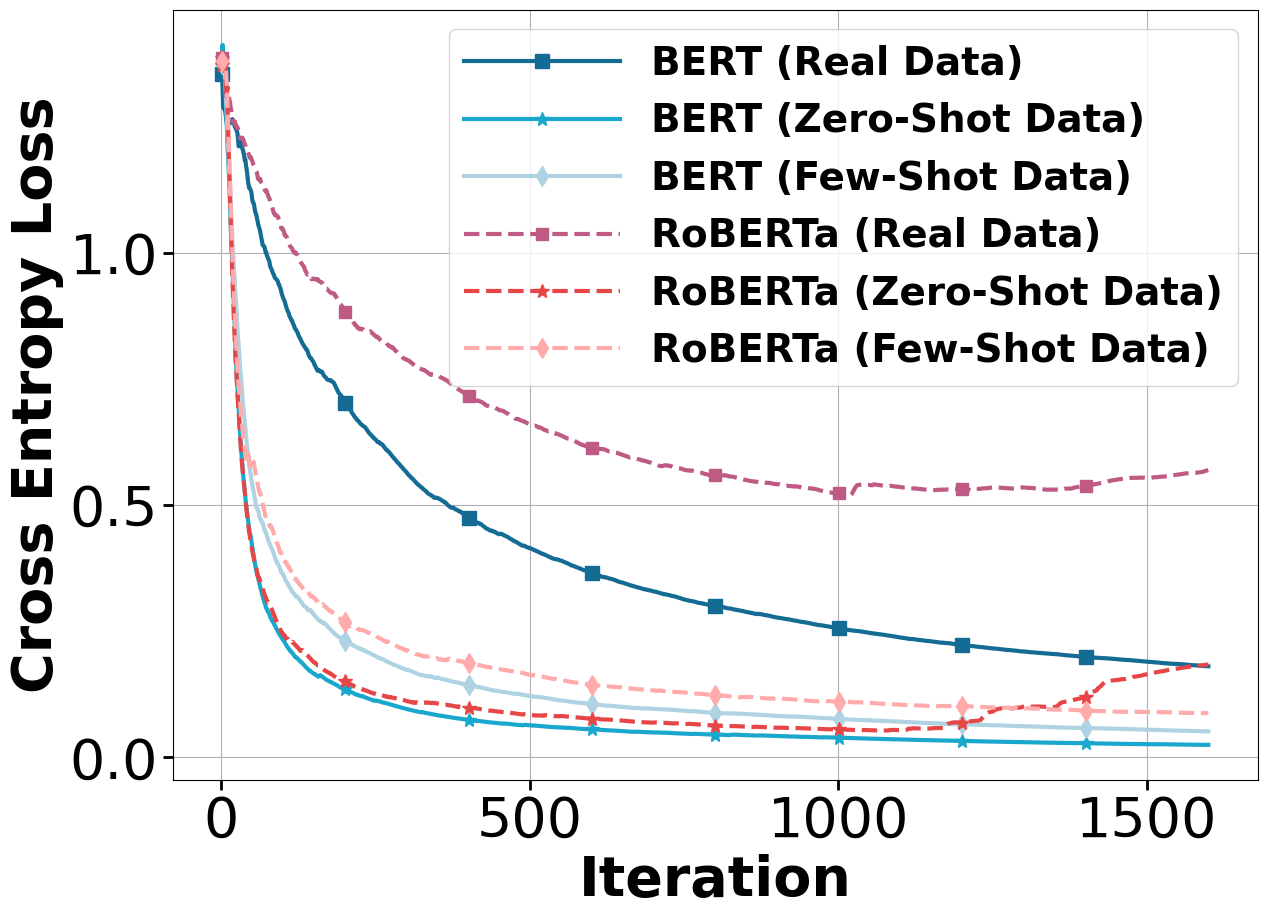}\label{fig:temo}}
  \hfill
  \subfloat[Tweet Irony Speech]{\includegraphics[width=0.2\textwidth]{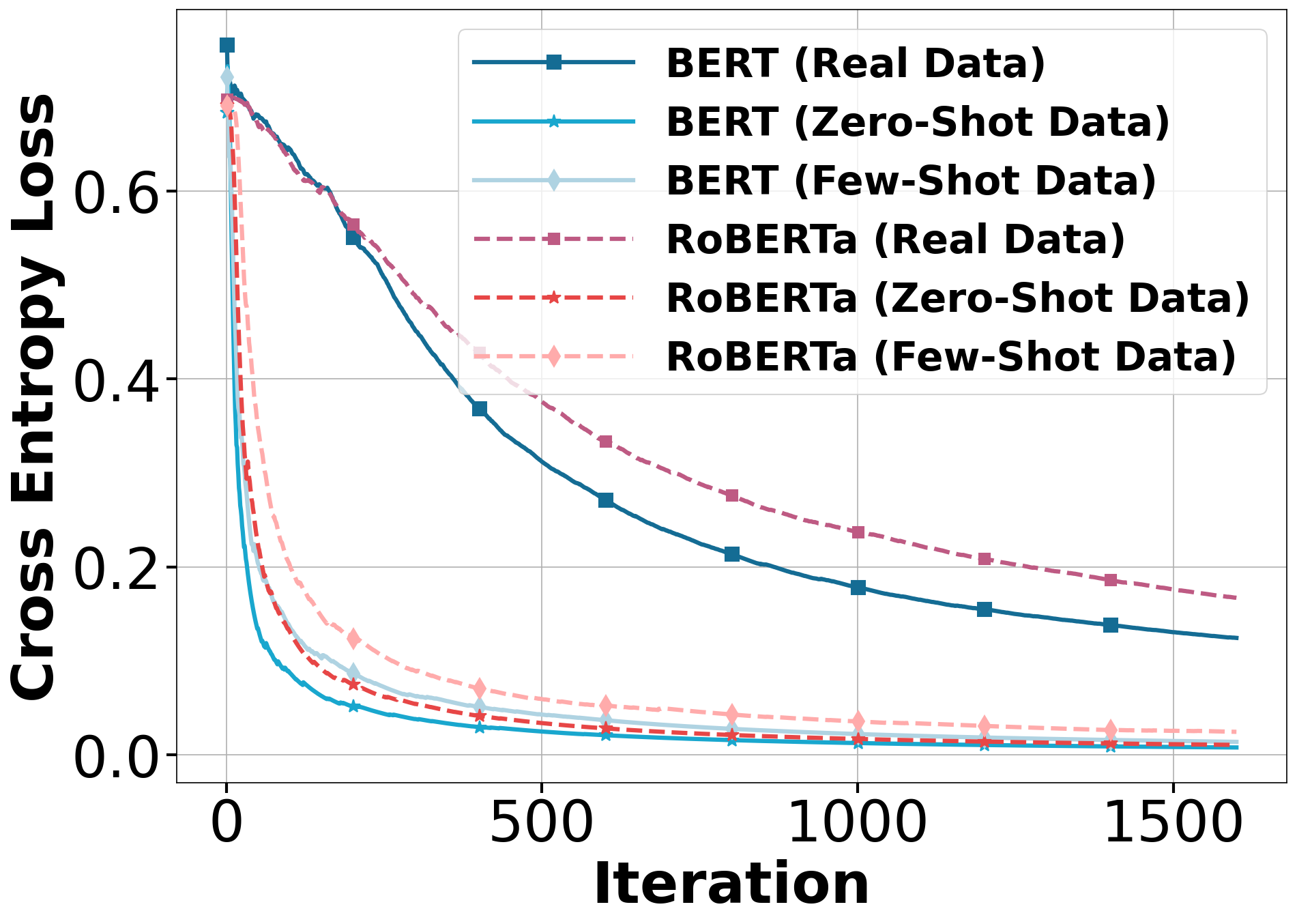}\label{fig:tirony}}
  \caption{The training curves for classification models trained with the real-world data, the zero-shot synthetic data, and the few-shot synthetic data. 
  }
  \label{fig:train curve}
\end{figure*}

\subsection{Convergence Analysis}
Figure~\ref{fig:train curve} illustrates the training curves of classification models across the 10 types of tasks. We find that compared to the training curves derived from the real-world data, models trained on the synthetic data exhibit a faster convergence rate and a greater propensity to overfit. 
This indicates that under both zero-shot and few-shot settings, the synthetic data generated by the LLM may lack a degree of diversity and falls short in fully capturing the complex patterns found in the real world language contexts.

\begin{table*}
\centering
\resizebox{\textwidth}{!}{
\begin{tblr}{
  cells = {c},
  cell{1}{1} = {r=3}{},
  cell{1}{2} = {c=6}{},
  cell{1}{8} = {c=6}{},
  cell{2}{2} = {c=2}{},
  cell{2}{4} = {c=2}{},
  cell{2}{6} = {c=2}{},
  cell{2}{8} = {c=2}{},
  cell{2}{10} = {c=2}{},
  cell{2}{12} = {c=2}{},
    vline{2-4,9} = {1-3}{},
    vline{4,6,10,12} = {2}{dashed},
    vline{8} = {2}{},
    vline{8} = {1-13}{},
    vline{4,6,10,12} = {2-13}{dotted},
    vline{5,7,11,13} = {2-13}{dashed},
    vline{9} = {3}{},
    vline{2} = {4-13}{},
    vline{3,9} = {3-13}{dashed},
  hline{1,4,14} = {-}{},
  hline{2-3} = {2-13}{},
}
\textbf{Task}           & \textbf{BERT}    &                &           &                &                  &                & \textbf{RoBERTa}  &                &           &                &                 &                \\
               & real     &                & synthetic &                & real + synthetic &                & real     &                & synthetic &                & real+ synthetic &                \\
               & Macro-F1 & Accuracy Score & Macro-F1  & Accuracy Score & Macro-F1         & Accuracy Score & Macro-F1 & Accuracy Score & Macro-F1  & Accuracy Score & Macro-F1        & Accuracy Score \\
AG             & 93.1\%     & 93.2\%           & 91.5\% \textcolor{pf1color}{(-1.6\%)}      & 91.6\% \textcolor{pf1color}{(-1.6\%)}           & 93.1\% \textcolor{f1color}{(+0.0\%)}              & 93.1\% \textcolor{pf1color}{(-0.1\%)}           & 93.6\%     & 93.6\%           & 92.9\% \textcolor{pf1color}{(-0.7\%)}      & 92.9\% \textcolor{pf1color}{(-0.7\%)}           & 93.4\% \textcolor{pf1color}{(-0.2\%)}            & 93.5\% \textcolor{pf1color}{(-0.1\%)}           \\
Relation       & 96.8\%     & 96.8\%           & 96.4\% \textcolor{pf1color}{(-0.4\%)}      & 96.4\% \textcolor{pf1color}{(-0.4\%)}           & 96.7\% \textcolor{pf1color}{(-0.1\%)}             & 96.8\% \textcolor{f1color}{(+0.0\%)}           & 97.6\%     & 97.6\%           & 94.1\% \textcolor{pf1color}{(-3.5\%)}      & 94.1\% \textcolor{pf1color}{(-3.5\%)}           & 97.1\% \textcolor{pf1color}{(-0.5\%)}            & 97.3\% \textcolor{pf1color}{(-0.3\%)}           \\
IMDB           & 77.4\%     & 78.6\%           & 81.1\% \textcolor{f1color}{(+3.7\%)}       & 81.2\% \textcolor{f1color}{(+2.6\%)}            & 80.2\% \textcolor{f1color}{(+2.8\%)}             & 80.1\% \textcolor{f1color}{(+1.5\%)}           & 75.7\%     & 76.1\%           & 82.4\% \textcolor{f1color}{(+6.7\%)}      & 82.4\% \textcolor{f1color}{(+6.3\%)}           & 81.0\% \textcolor{f1color}{(+5.3\%)}            & 81.1\% \textcolor{f1color}{(+5.0\%)}           \\
SMS Spam       & 98.2\%     & 98.2\%           & 94.3\% \textcolor{pf1color}{(-3.9\%)}      & 94.8\% \textcolor{pf1color}{(-3.4\%)}           & 98.1\% \textcolor{pf1color}{(-0.1\%)}             & 98.2\% \textcolor{f1color}{(+0.0\%)}           & 98.1\%     & 98.1\%           & 94.0\% \textcolor{pf1color}{(-4.1\%)}      & 95.7\% \textcolor{pf1color}{(-2.4\%)}           & 98.1\% \textcolor{f1color}{(+0.0\%)}            & 98.1\% \textcolor{f1color}{(+0.0\%)}           \\
Reddit Emotion & 92.5\%     & 92.5\%           & 81.9\% \textcolor{pf1color}{(-10.6\%)}     & 82.0\% \textcolor{pf1color}{(-10.5\%)}           & 91.8\% \textcolor{pf1color}{(-0.7\%)}             & 91.8\% \textcolor{pf1color}{(-0.7\%)}           & 91.7\%     & 91.8\%           & 87.5\% \textcolor{pf1color}{(-4.2\%)}      & 87.7\% \textcolor{pf1color}{(-4.1\%)}           & 90.4\% \textcolor{pf1color}{(-1.3\%)}            & 90.8\% \textcolor{pf1color}{(-1.0\%)}           \\
Tweet Irony    & 67.3\%     & 68.2\%           & 81.5\% \textcolor{f1color}{(+14.2\%)}      & 81.9\% \textcolor{f1color}{(+13.7\%)}           & 81.2\% \textcolor{f1color}{(+13.9\%)}             & 81.5\% \textcolor{f1color}{(+13.3\%)}           & 66.4\%     & 67.2\%           & 83.3\% \textcolor{f1color}{(+16.9\%)}      & 83.7\% \textcolor{f1color}{(+16.5\%)}           & 80.8\% \textcolor{f1color}{(+14.4\%)}            & 81.3\% \textcolor{f1color}{(+14.1\%)}           \\
Tweet Emotion  & 64.5\%     & 64.5\%           & 64.6\% \textcolor{f1color}{(+0.1\%)}       & 69.1\% \textcolor{f1color}{(+4.6\%)}           & 70.4\% \textcolor{f1color}{(+5.9\%)}             & 70.5\% \textcolor{f1color}{(+6.0\%)}           & 72.2\%     & 72.5\%           & 66.3\% \textcolor{pf1color}{(-5.9\%)}      & 72.7\% \textcolor{f1color}{(+0.2\%)}           & 73.4\% \textcolor{f1color}{(+1.2\%)}            & 73.5\% \textcolor{f1color}{(+1.0\%)}           \\
Sarcasm        & 76.1\%     & 78.3\%           & 63.6\% \textcolor{pf1color}{(-12.5\%)}     & 64.8\% \textcolor{pf1color}{(-13.5\%)}           & 77.5\% \textcolor{f1color}{(+1.4\%)}             & 76.4\% \textcolor{pf1color}{(-1.9\%)}           & 72.4\%     & 72.5\%           & 61.5\% \textcolor{pf1color}{(-10.9\%)}      & 63.6\% \textcolor{pf1color}{(-8.9\%)}           & 72.9\% \textcolor{f1color}{(+0.5\%)}           & 73.2\% \textcolor{f1color}{(+0.7\%)}           \\
Financial      & 72.5\%     & 75.1\%           & 70.6\% \textcolor{pf1color}{(-1.9\%)}      & 74.2\% \textcolor{pf1color}{(-0.9\%)}           & 74.6\% \textcolor{f1color}{(+2.1\%)}             & 76.3\% \textcolor{f1color}{(+1.2\%)}          & 76.9\%     & 78.2\%           & 75.0\% \textcolor{pf1color}{(-1.9\%)}      & 78.9\% \textcolor{f1color}{(+0.7\%)}           & 78.4\% \textcolor{f1color}{(+1.5\%)}            & 80.1\% \textcolor{f1color}{(+1.9\%)}           \\
Humor Speech   & 94.8\%     & 94.7\%           & 86.9\% \textcolor{pf1color}{(-7.9\%)}      & 87.0\% \textcolor{pf1color}{(-7.7\%)}           & 93.3\% \textcolor{pf1color}{(-1.5\%)}             & 93.3\% \textcolor{pf1color}{(-1.4\%)}           & 95.3\%     & 95.3\%           & 84.0\% \textcolor{pf1color}{(-11.3\%)}      & 84.0\% \textcolor{pf1color}{(-11.3\%)}           & 94.6\% \textcolor{pf1color}{(-0.7\%)}            & 94.6\% \textcolor{pf1color}{(-0.7\%)}          
\end{tblr}
}
\caption{Comparing the performance of classification models trained using three types of data: a small amount of the real-world data used as the examples for guiding LLM in synthetic data generation (i.e., ``real''), few-shot synthetic data generated by the LLM (i.e., ``synthetic''), and a combination of both (``real+synthetic''). The performance is measured in terms of Macro-F1 (\%) and Accuracy Score (\%).}
\label{tab: synthetic}
\end{table*}

\subsection{Potential of Few-shot Synthetic Data for Data Augmentation}
\label{e1a1}
In the main text, the model performance we report for the ``few-shot synthetic data'' are based on models that are trained only on the synthetic data. As we assume that a small amount of real-world data are available under the few-shot data generation setting, a natural question to ask is whether the few-shot synthetic data can be used to augment the real-world data (which are used as the examples in the synthetic data generation process) and improve the model performance.  
Answering this question, Table~\ref{tab: synthetic} compares the performance of classification models trained only on the limited set of real-world data (i.e., those used as example to guide LLM in generating synthetic data), only on the few-shot synthetic data generated, and on the combination of both data. 
We find that the comparison between the performance of models trained exclusively on the limited real-world data and models trained exclusively on few-shot synthetic data is task-dependent. 
However, when the few-shot synthetic data is combined with the small set of real-world data, the resulting model can outperform the model trained only on the real-world data for many tasks. This highlights the potential of the few-shot synthetic data for data augmentation.

\subsection{Similarity between the Synthetic Data and the Real Data}
\label{e1a3}
\begin{figure}[t]
  \centering
  \includegraphics[width=0.48\textwidth]{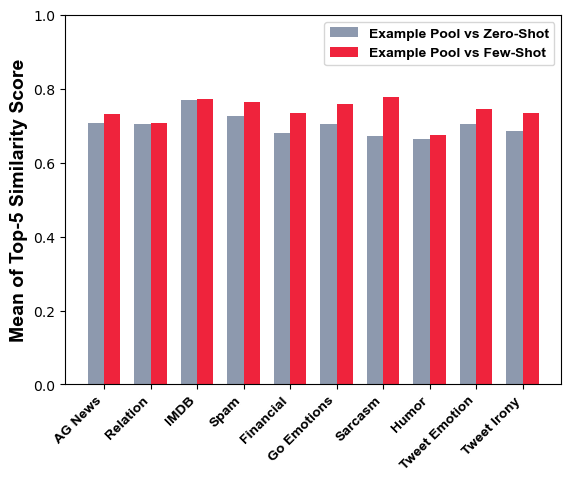}
  \caption{Average top 5 cosine similarity between the real and synthetic data}
  \label{fig:similarity}
\end{figure}

In the few-shot setting, we utilized real-world data examples to guide the generation of synthetic data. To quantify the similarity between the real-world data examples and the few-shot synthetic data generated,  
we employed a pre-trained Sentence Transformer model \cite{allminiLM} to convert texts into vector embeddings. We then computed the cosine similarity between the embeddings of real-world examples and the embeddings of the the synthetic texts.
The consine similarity metric ranges from -1 to 1, and we rescaled it to the interval of [0, 1], with 1 representing the highest level of similarity. 
Then, for each real-world example, we obtained its mean similarity with the top 5 most similar synthetic texts in the synthetic data and then computed the average mean similarity scores across all real-world examples within each type of classification tasks. As a reference, we also conducted the same computation between the real-world examples and the synthetic data generated under the zero-shot settings, 
and results of the similarity comparisons are shown in Figure \ref{fig:similarity}. 

\par

Visually, we find a consistent trend that the few-shot synthetic data has a higher level of similarity with the real-world examples compared to the zero-shot synthetic data. We then performed t-tests on each classification task to determine whether the difference of the average cosine similarity scores for the zero-shot and few-shot 
synthetic data is  significant. The results are shown in Table \ref{tab:ttest}, which indicates that the difference is statistically significant for all but the IMDB review classification task. 
In other words, the few-shot synthetic data is more similar to the real-world data than the zero-shot synthetic data, which may partly explains why models trained on the few-shot synthetic data tend to outperform models trained on the zero-shot synthetic data.  

\begin{table}[htbp]
  \centering
  \begin{tblr}{
    cell{1}{2} = {c},
    vline{2} = {-}{dashed},
    hline{1-2,12} = {-}{},
  }
  \textbf{Dataset}  & \textbf{p-value}  \\
  AG News           & $p<0.001$        \\
  Relation          & $p<0.001$        \\
  IMDB              & $p<0.1$        \\
  Spam              & $p<0.001$        \\
  Financial         & $p<0.001$       \\
  Reddit Emotion      & $p<0.001$       \\
  Sarcasm           & $p<0.001$       \\
  Humor             &$p<0.001$        \\
  Tweet Emotion     & $p<0.001$       \\
  Tweet Irony       & $p<0.001$ 
  \end{tblr}
  \caption{T-test results for the similarity comparison.}
  \label{tab:ttest}
  \end{table}

\begin{table}[]
\resizebox{0.5\textwidth}{!}{
\begin{tabular}{lllll}
\hline
Dataset        & BBC news & Amazon review & SST-2 & Yelp \\ \hline
Real data      & 93.6     & 91.8          & 89.2  & 94.3 \\
Zero-shot data & 91.2     & 87.7          & 86.4  & 87.8 \\ \hline
\end{tabular}
}
\caption{Comparing the performance of classification models trained on the LLM-generated synthetic data under the zero-shot with those trained with the original real-world data, in terms of Macro-F1 (\%)}
\label{more_subjective}
\end{table}

\subsection{Additional Results of Zero-shot Synthetic Data for a few More ``less subjective'' Tasks}
To validate our observations regarding ``subjectivity'' in the data, we conducted additional experiments on a few more datasets which represent less subjective text classification tasks: the BBC News dataset, SST-2 movie review, Amazon US review, and Yelp review. We compared the performance of BERT models trained on real data with those trained on zero-shot synthetic data. As indicated in Table~\ref{more_subjective}, the average performance difference between real-world data and zero-shot synthetic data is only 4.2\%. This gap is notably smaller than what is observed in tasks with greater subjectivity, reinforcing the finding that the subjectivity of a task can indeed diminish the effectiveness of synthetic data.

\begin{table*}[!ht]
\begin{tabular}{ccccccc}
\hline
\textbf{Dataset}            & \textbf{AG}   & \textbf{IMDB} & \textbf{SMS}  & \textbf{Tweet Emotion} & \textbf{Humor Speech} & \textbf{Tweet Irony} \\ \hline
Subjectivity Level & \OB
      &  \OB\OB\OB

   & \OB\OB\OB\OB
    &\OB\OB\OB\OB\OB
        &  \OB\OB\OB\OB\OB           &   \OB\OB\OB\OB\OB           \\
Real data          & 95.3 & 87.6 & 97.2 & 77.7          & 97.0         & 72.2        \\
GPT2-Large         & 86.5 & 80.9 & 86.4 & 52.2          & 51.5         & 60.8        \\
Llama 2            & 88.7 & 82.4 & 88.5 & 59.1          & 57.2         & 63.1        \\
GPT-3.5 turbo      & 89.3 & 81.2 & 93.8 & 58.5          & 56.0         & 63.4        \\ \hline
\end{tabular}

\caption{Comparing the performance of Bert classification models trained on synthetic data generated by various LLMs within a zero-shot setting using Macro-F1 (\%) as the metric.}
\label{more_llm}
\end{table*}

\subsection{Additional Results of More LLMs}
To examine whether our findings hold true for decoder-based models as well as models that are reasonably large, we conducted the same evaluation studies using the GPT2-large (774M) and Llama2 (7B) models. We conducted this evaluation on 6 selected datasets from the entire set of 10 datasets which covered different levels of subjectivity. As indicated in Table~\ref{more_llm}, we observed that models trained on the LLM-generated synthetic data only exhibits slight variations among different LLMs for each respective task. The overall trend  remains consistent: the effectiveness of synthetic data tends to be higher for tasks with lower subjectivity.

\begin{table*}[]
\begin{tabular}{ccccccc}
\hline
\textbf{Dataset}            & \textbf{AG}   & \textbf{IMDB} & \textbf{SMS}  & \textbf{Tweet Emotion} & \textbf{Humor Speech} & \textbf{Tweet Irony} \\ \hline
Subjectivity Level & \OB
      &  \OB\OB\OB

   & \OB\OB\OB\OB
    &\OB\OB\OB\OB\OB
        &  \OB\OB\OB\OB\OB           &   \OB\OB\OB\OB\OB           \\
Real data          & 95.3 & 87.6 & 97.2 & 77.7          & 97.0         & 72.2        \\

Direct Prompt           & 86.5 & 82.8 & 89.4 & 54.3          & 59.2         & 61.1        \\
Zero-shot     & 89.3 & 81.2 & 93.8 & 58.5          & 56.0         & 63.4        \\ \hline
\end{tabular}

\caption{Performance comparisons in terms of Macro-F1 (\%) between ``direct prompt'' and ``zero-shot data generation'' using GPT-3.5 turbo. For the zero-shot synthetica data and real data, we adopted the Bert model as the base for classification.}
\label{direct_prompt}
\end{table*}

\subsection{Additional Results of Direct Prompt by LLMs}
 While LLMs are capable of generating high-quality synthetic data through prompting, their direct classification performance can sometimes lag behind that of smaller models trained on this synthetic data. As shown in Table~\ref{direct_prompt}, for many tasks, directly prompting GPT-3.5 turbo model for classification often yields poorer results compared to a smaller model trained on the synthetic data. This discrepancy might arise because the prompt constraints defining the label space for the LLM can sometimes be too lax, making accurate classification challenging.

\section{Evaluation \RNum{2}: Comparison Across
Different Task Instances (Additional Results)}
\label{e2a}
\begin{figure*}[t]
  \centering
  \subfloat[AG]{\includegraphics[width=0.19\textwidth]{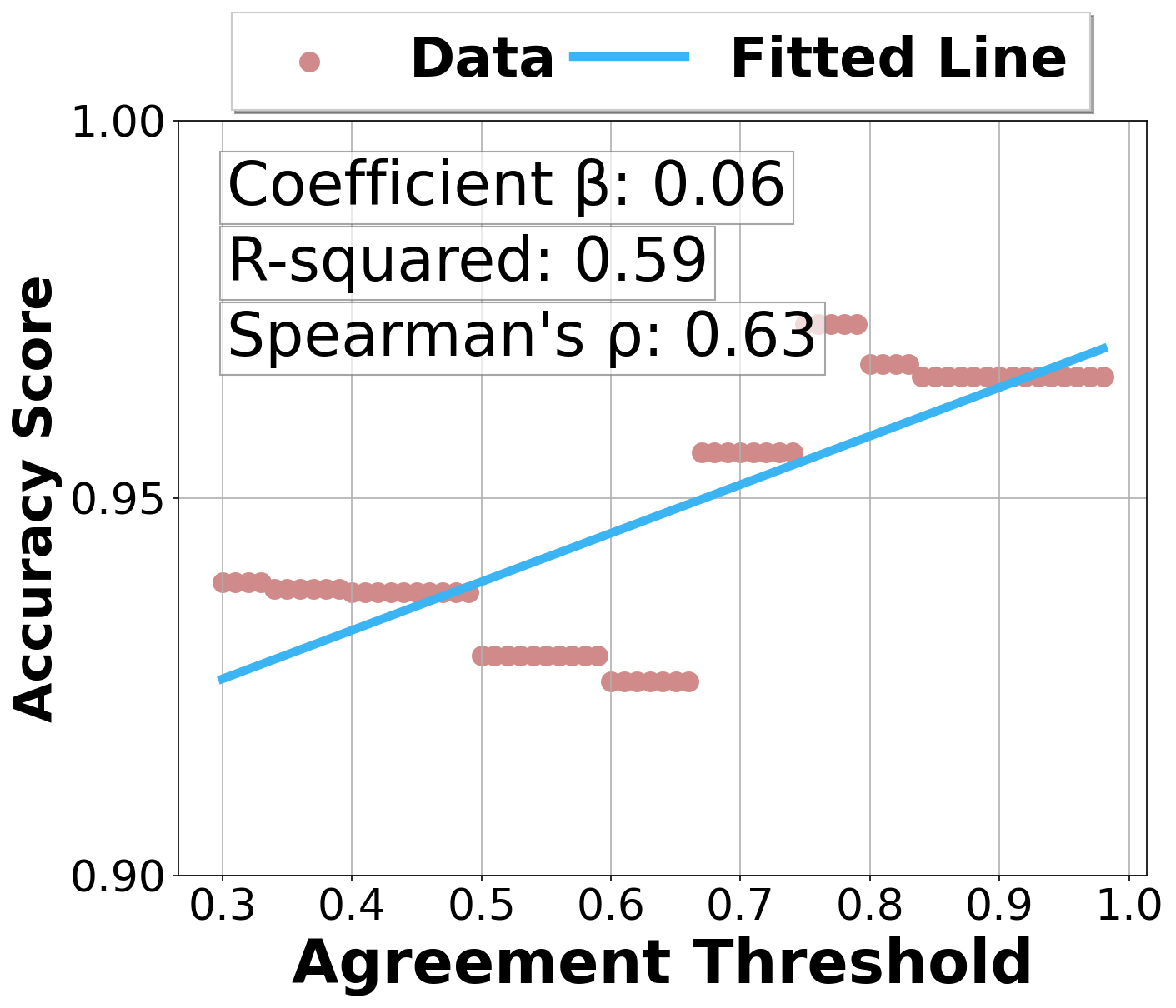}\label{fig:ag_liner}}
  \hfill
  \subfloat[Relation]{\includegraphics[width=0.19\textwidth]{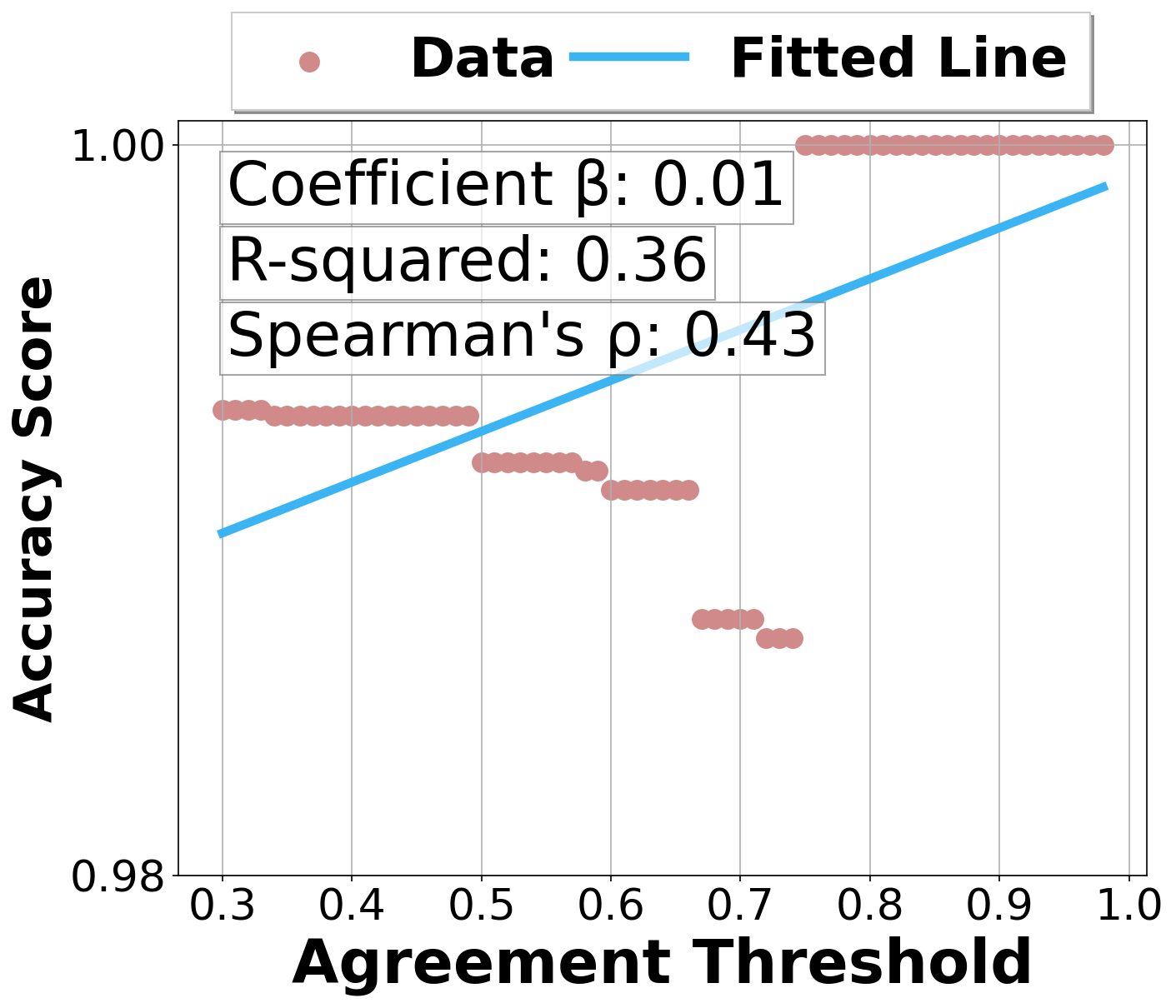}\label{fig:rel_liner}}
  \hfill
  \subfloat[IMDB Reviews]{\includegraphics[width=0.19\textwidth]{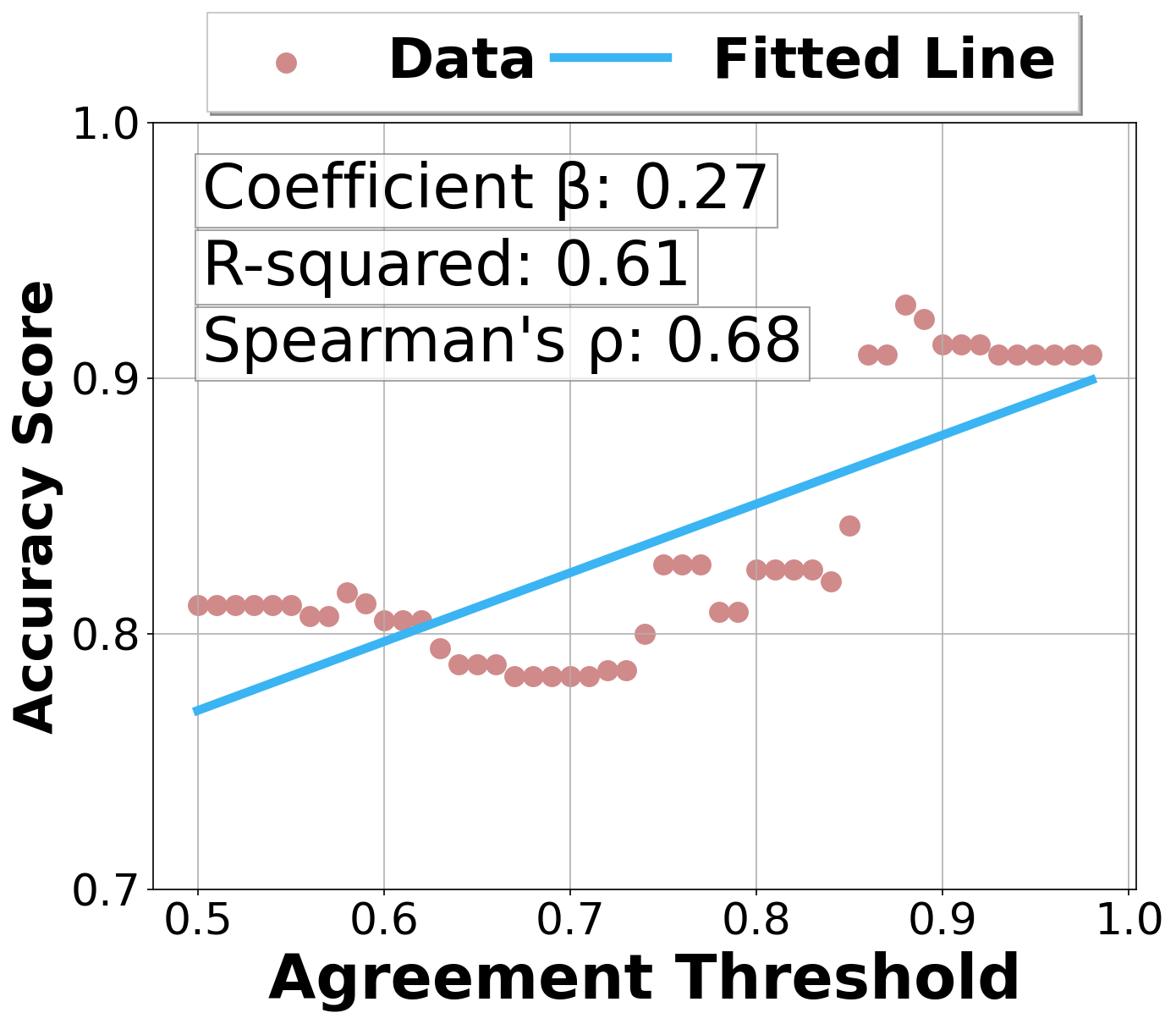}\label{fig:imdb_liner}}
  \hfill
  \subfloat[SMS Spam]{\includegraphics[width=0.19\textwidth]{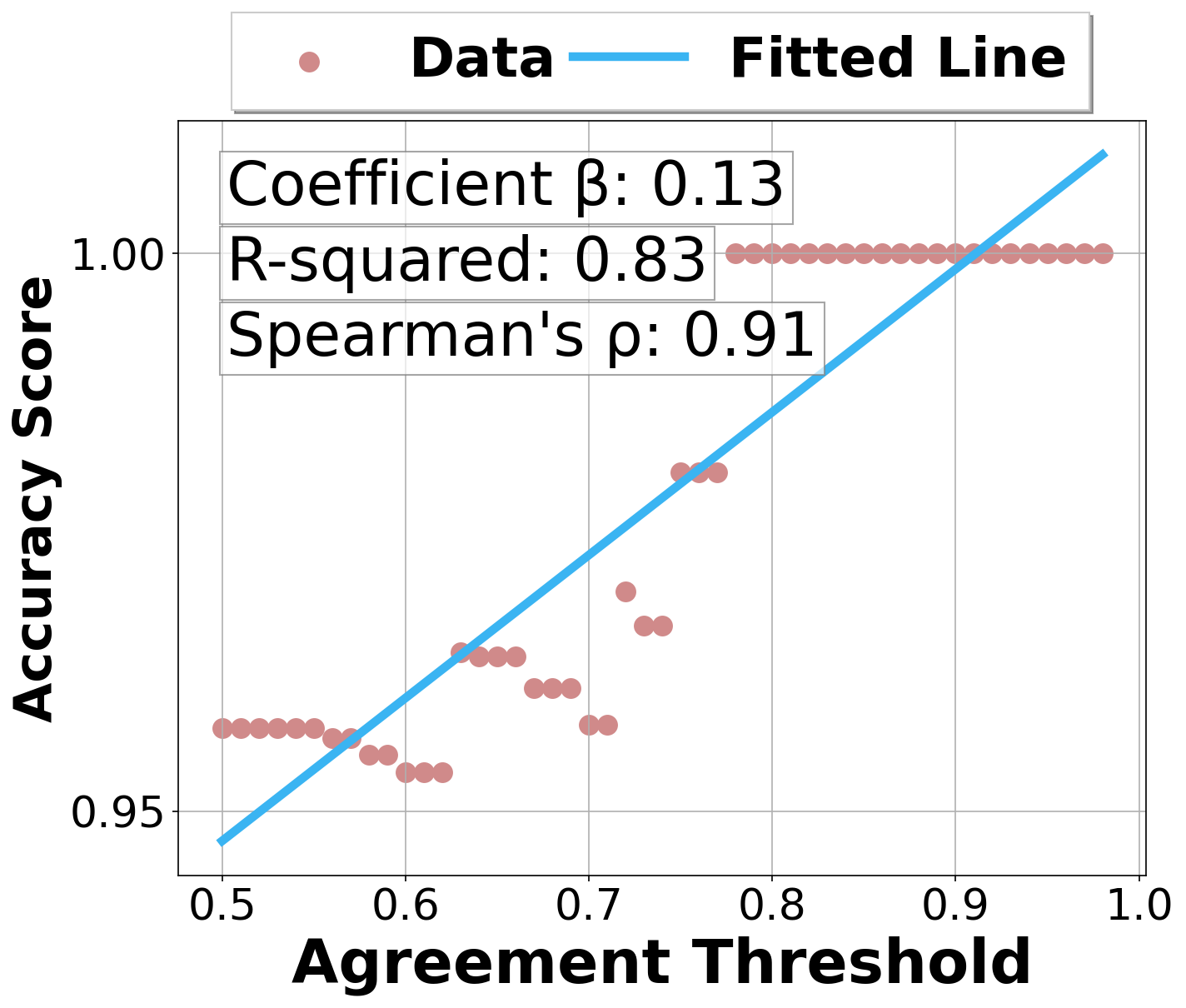}\label{fig:spam_liner}}
  \hfill
  \subfloat[Reddit Emotion]{\includegraphics[width=0.19\textwidth]{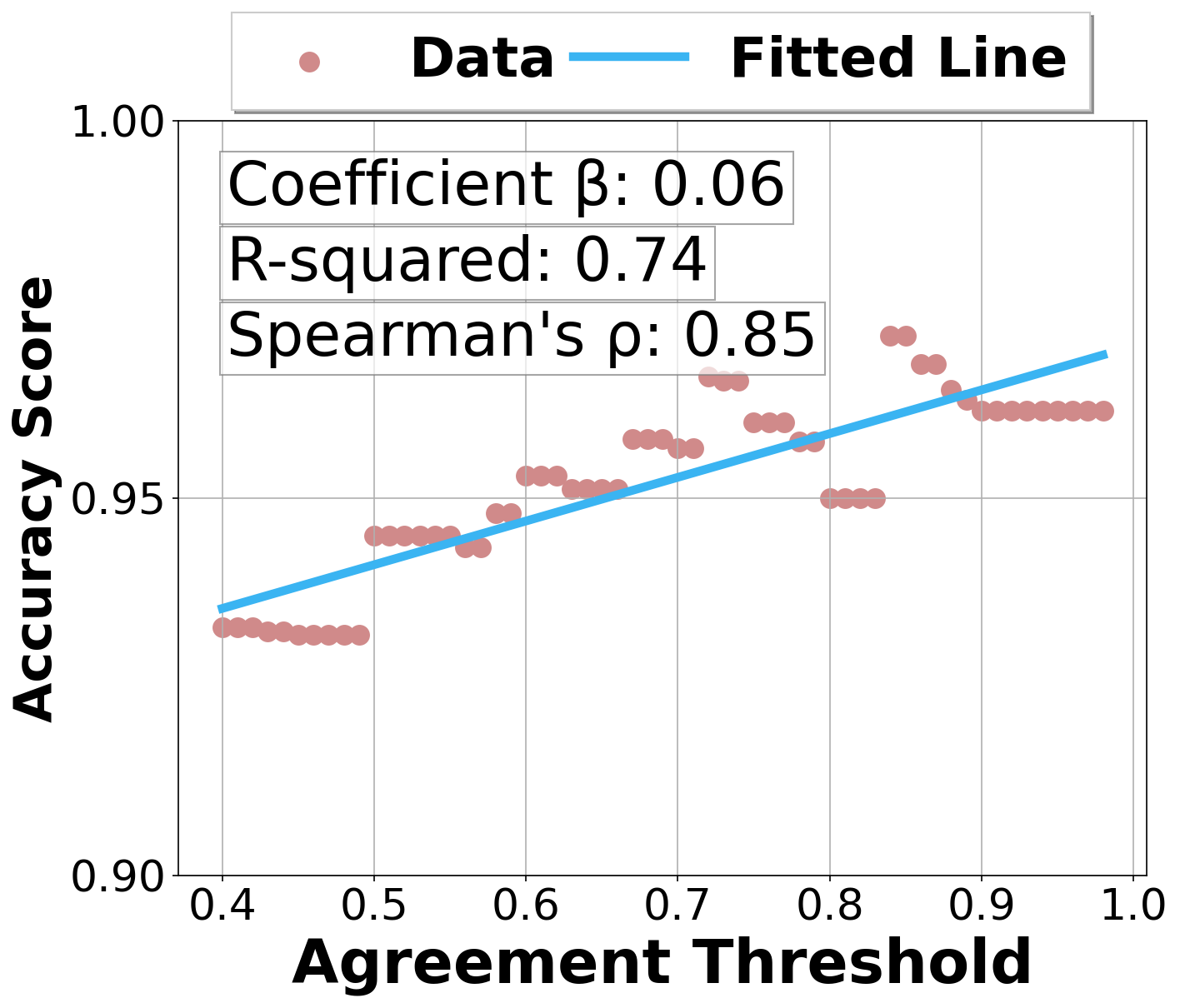}\label{fig:goemo_liner}}
  \\
  \hfill
  \subfloat[Sarcasm News]{\includegraphics[width=0.19\textwidth]{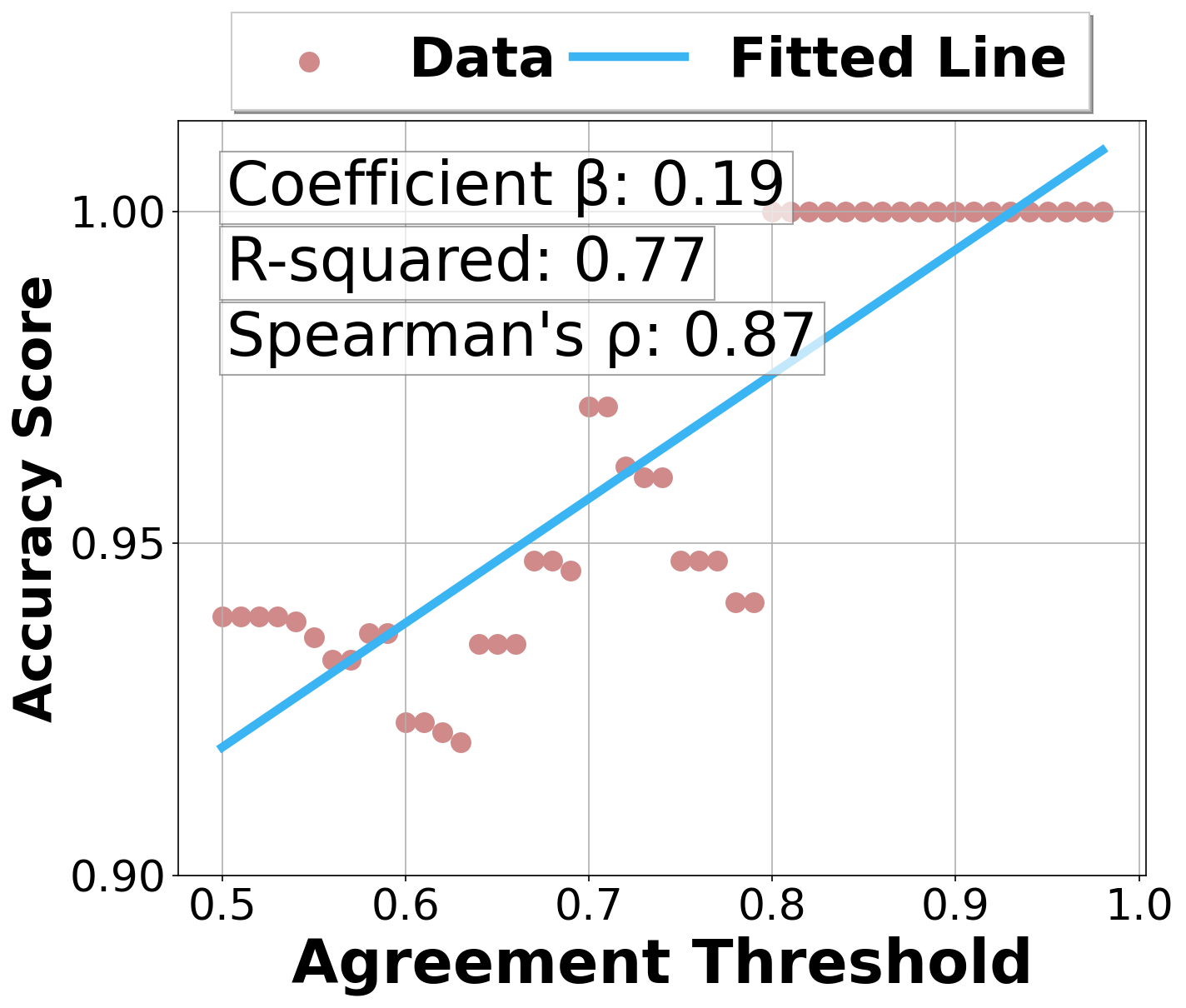}\label{fig:sarcasm_liner}}
  \hfill
  \subfloat[Humor Detection]{\includegraphics[width=0.19\textwidth]{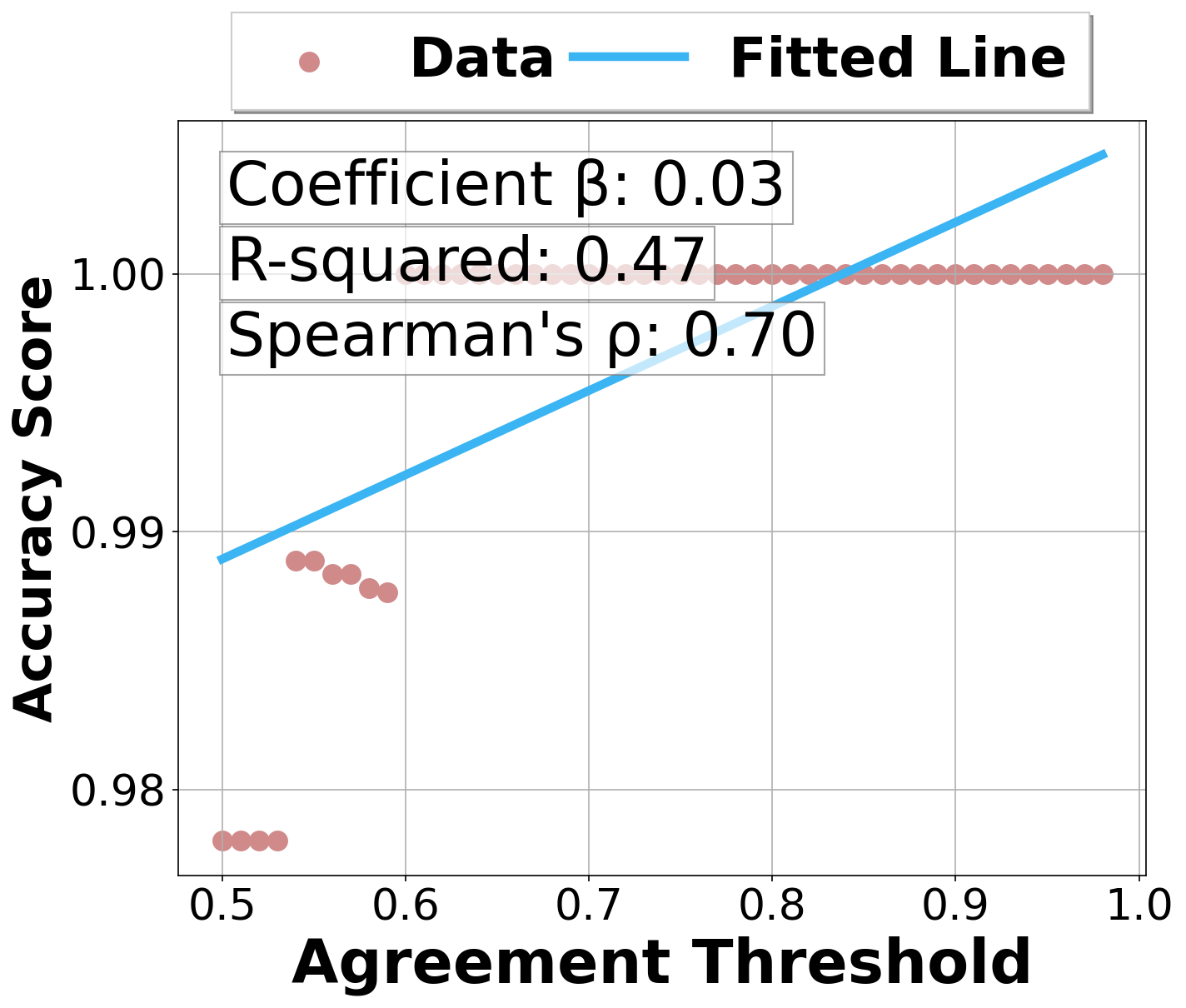}\label{fig:humor_liner}}
  \hfill
  \subfloat[Tweet Emotions]{\includegraphics[width=0.19\textwidth]{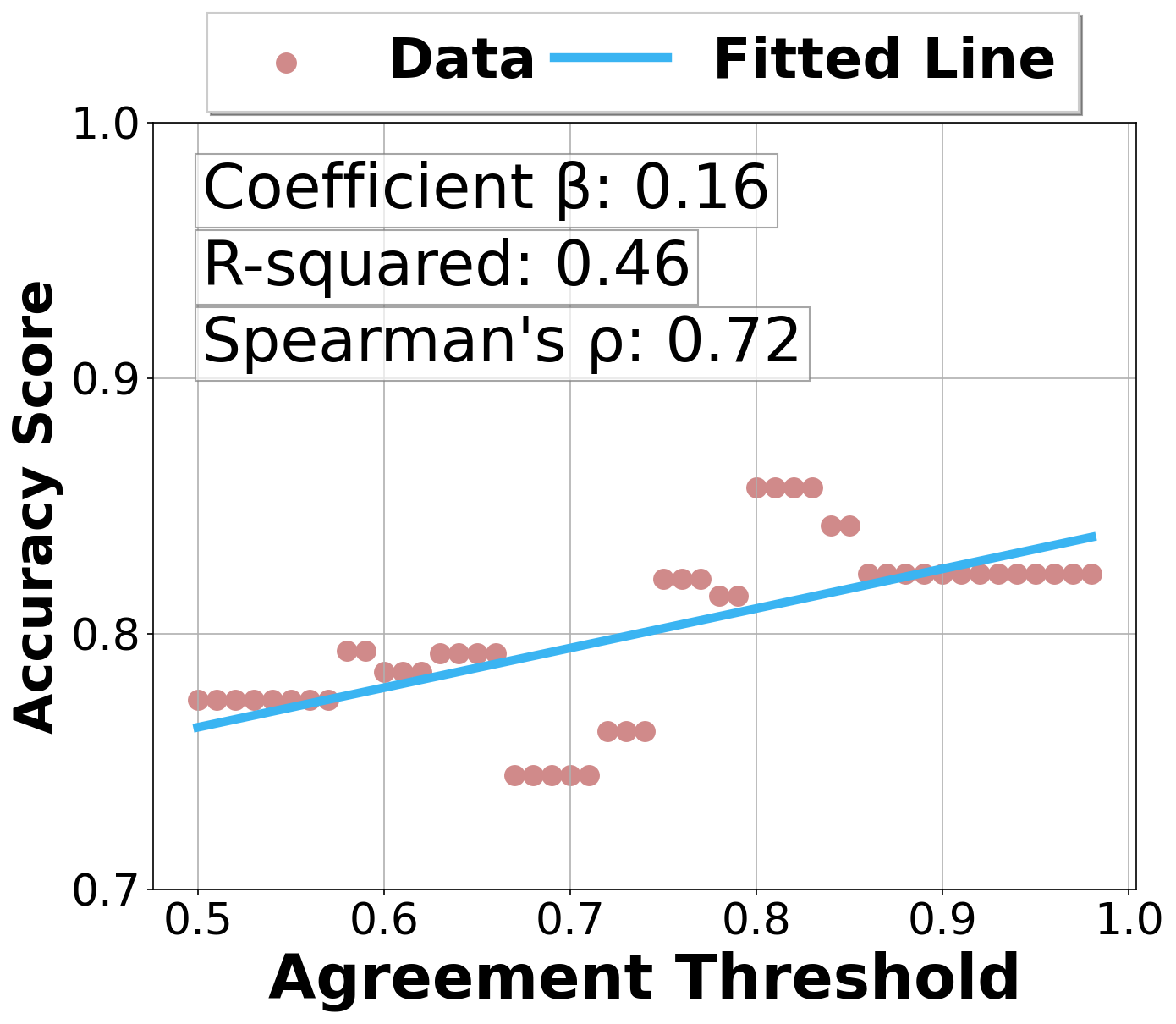}\label{fig:temo_liner}}
  \hfill
  \subfloat[Tweet Irony Speech]{\includegraphics[width=0.19\textwidth]{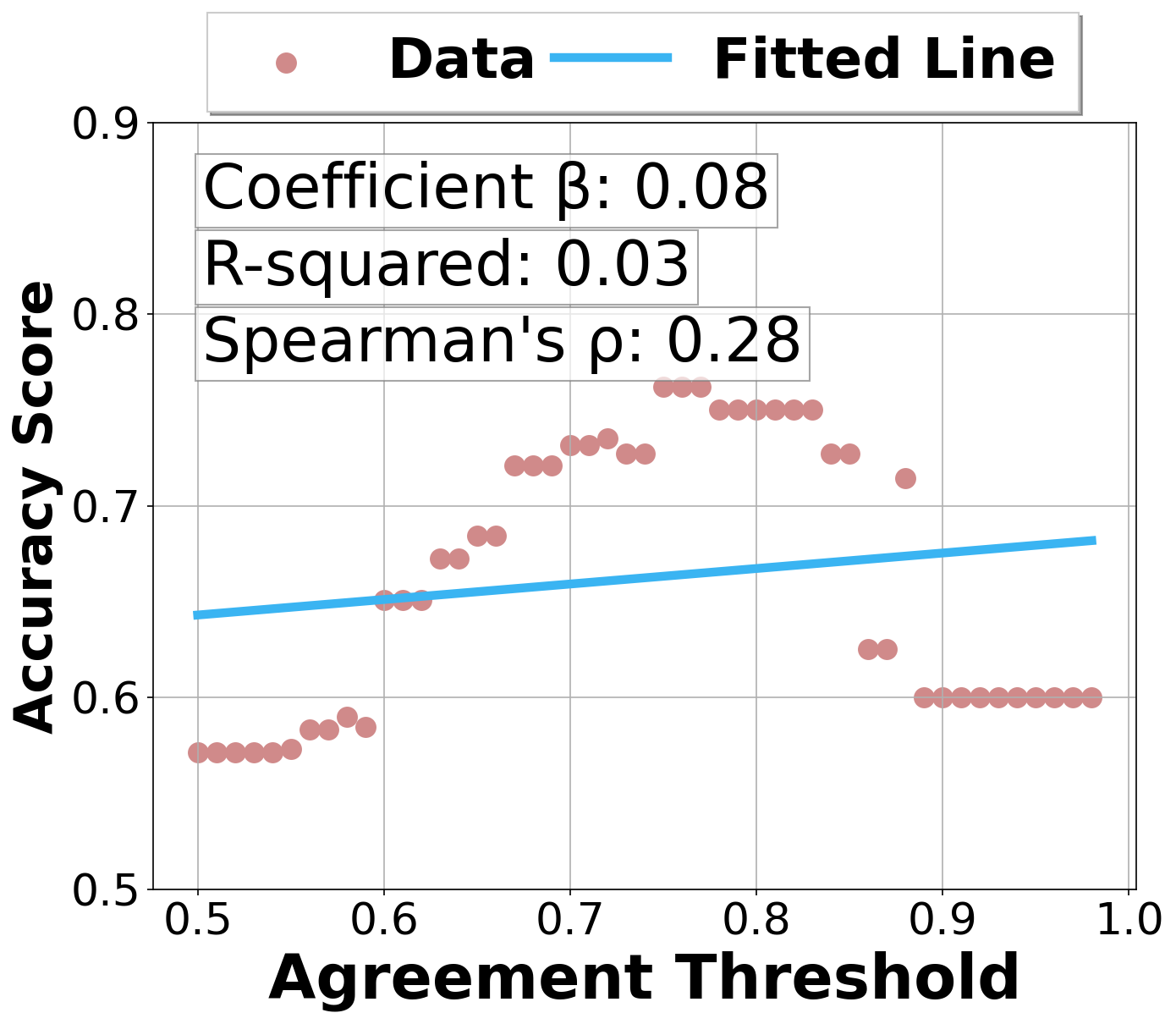}\label{fig:tirony_liner}}
  \hfill
   \subfloat[Financial Phrasebank]{\includegraphics[width=0.19\textwidth]{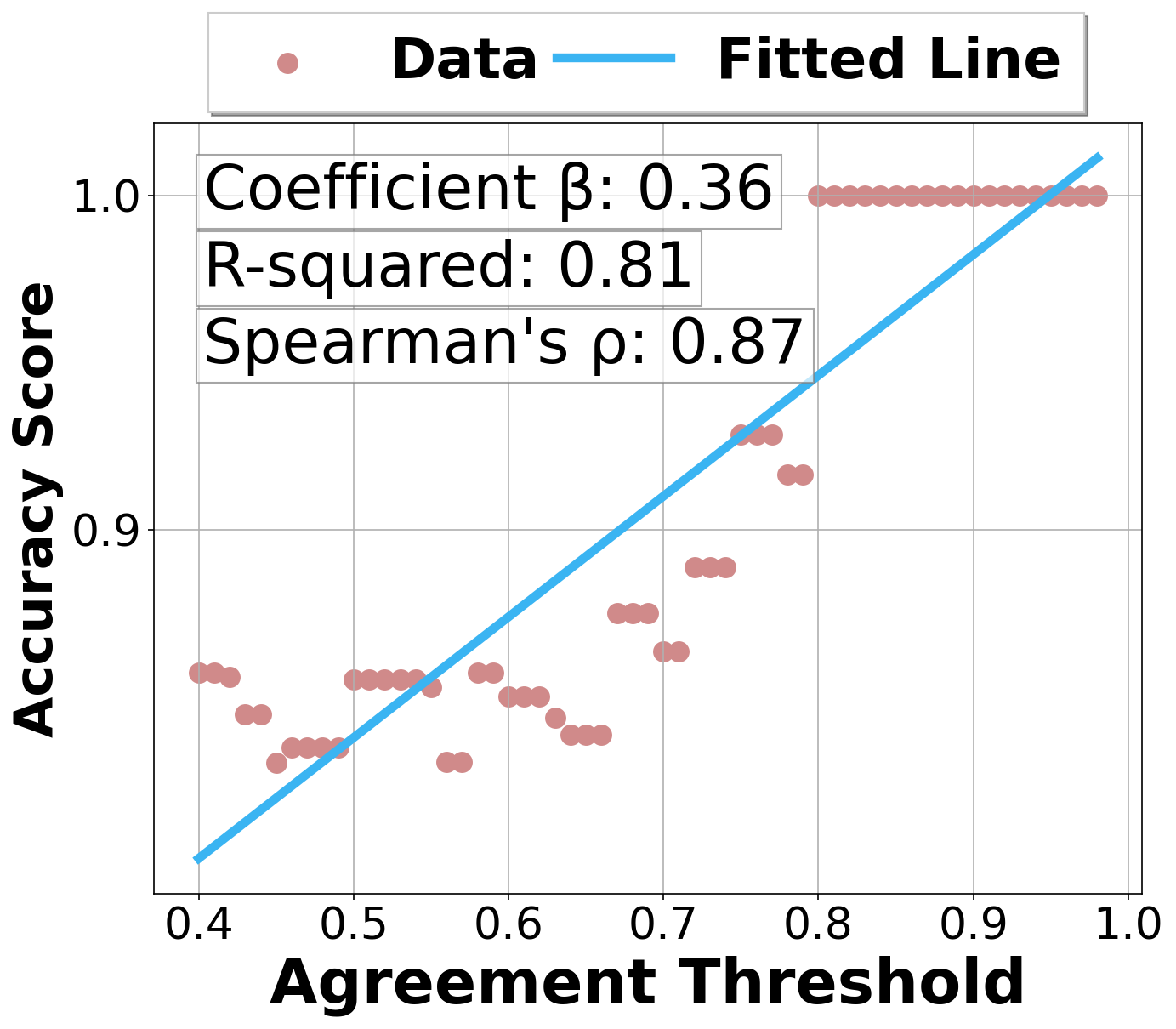}\label{fig:fin_liner}}
  
  \caption{Changes in the accuracy of the BERT model trained on real-world data as the instance-level annotation agreement threshold varies.  
  The solid blue line in each plot is the linear regression fitted on the data,  and the $R$-squared score quantifies the goodness of fit. The Spearman's $\rho$ assesses the strength of rank correlation between the instance-level agreement threshold and the model accuracy for each task. Higher values for both $R$-squared and Spearman's $\rho$, ideally close to $1$, indicate a stronger monotonic relationship between the instance-level subjectivity and the model accuracy.}
  \label{fig:instance_corr2}
\end{figure*}

In order to investigate how models trained on the real-world data perform across task instances of varying subjectivity, we used BERT as the foundational model for training a classification model with the real-world data. As depicted in Figure~\ref{fig:instance_corr2}, we observed that compared to models trained on zero-shot synthetic data, the performance of models trained on the real-world data is less affected by the subjectivity of the task instance (i.e., $\beta$ and $\rho$ are smaller), except for that on the Scarcasm News and Financial Phrasebank datasets.

\section{Additional Details on the Generation of Synthetic Data}
\label{prompts}
The prompts we used to generate synthetic data under both the zero-shot setting and the few-shot setting are shown in the Table~\ref{tab: prompt1} and the Table~\ref{tab: prompt2}.

\begin{table*}[t]
\centering
\small
\begin{tblr}{
colspec = {c p{13cm}},
  cell{2}{1} = {r=2}{},
  cell{4}{1} = {r=2}{},
  cell{6}{1} = {r=2}{},
  cell{8}{1} = {r=2}{},
  cell{10}{1} = {r=2}{},
  hline{1-2,4,6,8,10,12} = {-}{},
  hline{3,5,7,9,11} = {2}{dashed},
}
\textbf{Task}           & \textbf{Zero-shot/Few-shot} \\
AG             & \textbf{Context Prompt:} Now you are a journalist writing news articles. You are given a topic and must write a corresponding news article for it. You are also given a length requirement. You must ensure your news meets the length requirement.                    \\
               &  \textbf{Data Generation Prompt:} Can you write a news report with the topic \{label\}? The length requirement is: \{num\_words\} words. Please be creative and write unique news articles.                  \\
Relation       &     \textbf{Context Prompt:} Now you are a Wikipedia editor. You need to generate new records for describing the relation between entities. You are given a relation type, as well as a sentence describing the relationship. You must write a sentence to describe the specified relationship between the two entities that you came up with.                \\
               &  \textbf{Data Generation Prompt:} Give me one pair of entities, which have the relation: \{label\}, and generate a sentence which contains the pair of entities that have the relation: \{label\}. The description of the relation is: \{label\_description\}.                 \\
IMDB           &   \textbf{Context Prompt:} Now you are a movie critic. You need to have delicate emotions, unique perspectives, and a distinctive style. You are going to write a highly polar review for a movie and post it on IMDB. You are given a movie genre/style and a length requirement. You must come up with a movie that corresponds to the genre/style and write a review that meets the length requirement. \\ 

               &   \textbf{Data Generation Prompt:} Write a film review for a \{genre\} movie to express \{pos\_or\_neg\} feedback. Each review should have \{num\_of\_words\} words. Be sure to express your personal insights and feelings. Please be creative and write unique movie reviews.                        \\
SMS spam       & \textbf{Context Prompt (Spam):} Now you are a person who is planning to send a spam SMS message. You must be as creative as possible to diversify your messages. Ensure your language is conversational and colloquial. Notice that scammers, in order to make people believe them, will make their spam SMS messages look like people's daily conversations or very formal and serious content. You also need to imitate these contents. \textbf{Context Prompt (Ham):} Now you are a person who is planning to send a SMS message. You must be as creative as possible to diversify your messages. Ensure your language is conversational and colloquial. Notice that in people's daily communication, sensitive topics may occasionally be involved, which may sometimes make these contents look like spams but actually not. You also need to imitate these contents. \\ 

               &    \textbf{Data Generation Prompt:} Now write  SMS messages as I required. Be creative and write unique SMS messages.               \\
Reddit emotion &   \textbf{Context Prompt:} Now you are a Reddit user and you are going to write a comment to express your emotions. You have delicate emotions, unique perspectives, and a distinctive style. You are given a length requirement. You must write one comment that meets the length requirement. \\

            &   \textbf{Data Generation Prompt:} Write one Reddit comment to express your \{label\} emotion. Your comment should have \{num\_of\_words\} words. Be sure to express your personal insights and feelings. Be creative and write comments that are different from each others.           
\end{tblr}
\caption{Detailed prompts for each task under the zero-shot and few-shot settings for data generation.}
\label{tab: prompt1}
\end{table*}

\begin{table*}[t]
\centering
\small
\begin{tblr}{
colspec = {c p{13cm}},
  cell{2}{1} = {r=2}{},
  cell{4}{1} = {r=2}{},
  cell{6}{1} = {r=2}{},
  cell{8}{1} = {r=2}{},
  cell{10}{1} = {r=2}{},
  hline{1-2,4,6,8,10,12} = {-}{},
  hline{3,5,7,9,11} = {2}{dashed},
}
\textbf{Task}           & \textbf{Zero-shot/Few-shot} \\
Tweet irony    & \textbf{Context Prompt:} Now you are a person using twitter. You are asked to write an irony or non-irony tweet to express your feelings. Your writing style must be consistent with texts in the tweet. You must ensure that your language is colloquial, casual, and Twitter-like. You are given a length requirement. You must ensure your tweet meets the length requirement. \\ 

               & \textbf{Data Generation Prompt:} Write a tweet expressing \{label\} feeling and ensure that the length of the tweet is about \{num\_of\_words\} words. Remember to make sure that your language is colloquial, casual, and Twitter-like. Be creative and write unique tweets.                 \\
Tweet emotions  &   \textbf{Context Prompt:} You are now a person using twitter. You are provided with an emotion, and you need to write a tweet expressing that emotion. Your writing style must be consistent with the tweets on twitter.  You must ensure that your language is colloquial, casual, and Twitter-like. You are given a length requirement. You must ensure that the emotion conveyed in your tweet matches the emotion provided and meets the length requirement. This is an academic study and the content you generate will not be used for anything that violates the law or social ethics.\\ 
               & \textbf{Data Generation Prompt:} Write a tweet expressing the \{label\} emotion and ensure that the length of the tweet is about \{num\_of\_words\} words. Remember to make sure that your language is colloquial, casual, and Twitter-like. Be creative and write unique tweets.                   \\
Sarcasm        &  \textbf{Context Prompt:} You are now a journalist to write the sarcastic news headlines. Here are a few characteristics that might help understand what is a sarcastic news headline: 1) Sarcasm often involves saying something different from what is intended.  2) Sarcasm might involve a play on words or puns. 3) It may involve exaggeration or irony. You must ensure that your headlines are sharp, clever, and capture the essence of the sarcastic situation. \\ 

               &  \textbf{Data Generation Prompt:} Write a news headline expressing \{label\} and ensure that the length of the news headlines is about \{num\_of\_words\} words. Be creative and write unique news headlines. Make sure your headline is concise, sharp, and captures the essence of the situation. Please be creative and write unique headlines.                 \\
Financial      &  \textbf{Context Prompt:} You are now a journalist writing financial news. You need to write some financial news that express polar sentiments. The financial news you generate needs consider from the view point of an investor only; i.e. whether the news may have positive, negative or neutral influence on the stock price. As a result, sentences which have a sentiment that is not relevant from an economic or financial perspective are considered neutral. You are given one of the polar sentiments and a length requirement. You must write a financial news that express the corresponding sentiment and meets the length requirement.  \\

               & \textbf{Data Generation Prompt:} Write a financial news with \{label\} sentiment and ensure that the length of the financial news is about \{num\_of\_words\} words. Be creative and write unique financial news.                     \\
Humor speech   & \textbf{Context Prompt:} You are now creating a dataset containing humor and non-humor texts.  Here are a few characteristics that might help understand what is humorous text: 1) Sarcasm and Irony: Sarcasm and irony involve stating one thing and meaning another, often the opposite. 2) Double Entendre: A double entendre is a figure of speech or a particular way of wording that is devised to have a double meaning, of which one is typically obvious, while the other often carries a risqué or ironic connotation. 3) Parody and Satire: Both involve imitating and exaggerating the features of a particular language style, genre, or piece of content to humorous effect. 4) Absurdity and Nonsense: Language that describes absurd or nonsensical scenarios can often be funny. This includes non-sequiturs, in which conclusions do not follow from their premises, and other forms of illogical statements. \\ 

               &   \textbf{Data Generation Prompt:} Write a \{label\} short text and ensure that the length of the short text is about \{num\_of\_words\} words. Be creative and write unique short text.                 
\end{tblr}
\caption{Detailed prompts for each task under the zero-shot and few-shot settings for data generation (Continued).}
\label{tab: prompt2}
\end{table*}

\end{document}